\documentclass[3p]{elsarticle}

\usepackage[hidelinks]{hyperref}
\journal{Swarm and Evolutionary Computation}

\usepackage{import}
\usepackage{amsmath, amssymb, amsthm}
\usepackage{mathtools}
\usepackage{pifont}
\usepackage{graphicx}
\usepackage{color}
\usepackage{xcolor}
\usepackage{soul}
\usepackage{algpseudocode}
\usepackage{algorithm}
\usepackage{hyperref}
\usepackage{cleveref}
\usepackage{caption}
\usepackage{titlecaps}
\usepackage{subfig}
\usepackage{placeins}
\usepackage{pgf}
\usepackage{lipsum}
\usepackage{booktabs}
\usepackage[inline]{enumitem}
\usepackage{array}
\usepackage{svg}
\svgsetup{inkscapepath=svgsubdir}

\usepackage{subfiles}
\usepackage{bookmark}

\algnewcommand{\LineComment}[1]{\State{} \(\triangleright\) #1}
\newcommand{\testpnt}[1]{\textbf{#1}_{*}}
\newcommand{\minpnt}[1]{\textbf{#1}_{\textrm{min}}}

\newcommand{\partialDeriv}[2]{\frac{\partial{} #1}{\partial{} #2}}
\newcommand{\partialSecDeriv}[2]{\frac{\partial^{2} #1}{\partial^{2} #2}}
\newcommand{\partialSecDerivDual}[3]{\frac{\partial^{2} #1}{\partial{} #2 \partial{} #3}}

\DeclareMathOperator{\diag}{\textrm{diag}}
\DeclareMathOperator{\minmax}{\textrm{min-max}}
\newcommand{\tr}{\textrm{tr}}
\DeclareMathOperator{\ei}{\textrm{EI}}

\newcommand{\tra}[2][]{
  \ensuremath{#2'^{#1}}
}

\newcommand{\cen}[1]{
  \ensuremath{ #1_{\textrm{cen}} }
}

\newcommand{\rot}[1]{
  \ensuremath{ #1_{\textrm{rot}} }
}

\newcommand{\prevIterIndicator}{
  \ensuremath{t - 1}
}

\newcommand{\prevIterExtra}[2]{
  \ensuremath{ #1_{ #2, \, \prevIterIndicator{}} }
}
\newcommand{\prevIter}[1]{
  \ensuremath{ #1_{\prevIterIndicator{}} }
}

\newcommand{\curIterIndicator}{
  \ensuremath{t}
}

\newcommand{\curIterExtra}[2]{
  \ensuremath{ #1_{ #2, \, \curIterIndicator{}} }
}
\newcommand{\curIter}[1]{
  \ensuremath{ #1_{\curIterIndicator{}} }
}

\newcommand{\nextIterIndicator}{
  \ensuremath{t + 1}
}

\newcommand{\nextIter}[1]{
  \ensuremath{ #1_{\nextIterIndicator{}} }
}

\makeatletter
\newcommand{\subalign}[1]{%
  \vcenter{%
    \Let@ \restore@math@cr \default@tag
    \baselineskip\fontdimen10 \scriptfont\tw@
    \advance\baselineskip\fontdimen12 \scriptfont\tw@
    \lineskip\thr@@\fontdimen8 \scriptfont\thr@@
    \lineskiplimit\lineskip
    \ialign{\hfil$\m@th\scriptstyle##$&$\m@th\scriptstyle{}##$\hfil\crcr
      #1\crcr
    }%
  }%
}
\makeatother

\newcommand{\acr}[1]{\MakeLowercase{#1}}

\newcommand{\ALGPA}{LABCAT}
\newcommand{\ALGPAfull}{\acr{l}ocally \acr{a}daptive Bayesian \acr{o}ptimi\sz{}ation using \acr{p}rincipal-\acr{C}omponent-aligned \acr{t}rust \acr{R}egions}

\newcommand{\TRBO}{trust-region-based BO}
\newcommand{\sz}{z}

\newcommand{\m}{\rho}

\newcommand{\ds}{ds}

\newcommand{\ou}{ou}
\newcommand{\centre}{centre}

\newcommand{\centered}{centred}

\newcommand{\recentered}{recentred}

\newcommand{\recentering}{recentring}

\newcommand{\revision}[1]{#1}  %revision not shown

\usepackage[disable]{todonotes}

\begin{document}

\begin{frontmatter}

  \title{LABCAT:\ Locally \acr{a}daptive Bayesian \acr{o}ptimi\sz{}ation using \revision{\acr{p}rincipal-\acr{c}omponent-aligned} \acr{t}rust \acr{r}egions}
  \date{\today}
  
  \author[adr]{E. Visser\corref{cor1}}
  \address[adr]{Department of Electrical and Electronic Engineering, Stellenbosch University, South Africa}
  \ead{emile.visser@gmail.com}
  \cortext[cor1]{Corresponding author}
  \author[adr]{C.E. van Daalen}
  \ead{cvdaalen@sun.ac.za}
  \author[adr]{J.C. Schoeman}
  \ead{jcschoeman@sun.ac.za}
  
  \begin{abstract}
      Bayesian optimi\sz{}ation (BO) is a popular method for optimi\sz{}ing expensive black-box functions. BO has several well-documented shortcomings, including computational slowdown with longer optimi\sz{}ation runs, poor suitability for non-stationary or ill-conditioned objective functions, and poor convergence characteristics. Several algorithms have been proposed that incorporate local strategies, such as trust regions, into BO to mitigate these limitations; however, none address all of them satisfactorily. To address these shortcomings, we propose the \ALGPA{} algorithm, which extends \revision{trust-region-based} BO by adding a \revision{rotation aligning the trust region with the weighted principal components} and an adaptive rescaling strategy based on the length-scales of a local Gaussian process surrogate model with automatic relevance determination. Through extensive numerical experiments using a set of synthetic test functions and the well-known COCO benchmarking software, we show that the \ALGPA{} algorithm outperforms several state-of-the-art BO and other black-box optimi\sz{}ation algorithms.
  \end{abstract}
  
  
  \end{frontmatter}

%**********************************************
\section{Introduction}\label{sec:intro}
%**********************************************

Optimi\sz{}ation refers to the fundamental aim of finding the specific set of inputs that maximi\sz{}es or minimi\sz{}es a certain objective function, possibly subject to constraints. There is a multitude of optimi\sz{}ation strategies to choose from, each with the goal of converging to a local or global solution as quickly and as accurately as possible. Of specific interest for this paper is the problem of so-called ``black-box'' optimi\sz{}ation: an optimi\sz{}ation problem where only the output of the objective function can be observed for a given input. These functions may arise when the objective function lacks a closed-form expression or gradient information is unavailable, and are often computationally expensive to evaluate.%Examples of these functions can be found in hyperparameter tuning~\cite{BOHyperparameters, BOHypTuning}

One family of optimi\sz{}ation methods that addresses the challenge of optimi\sz{}ing expensive black-box functions in a sample-efficient manner is \acr{S}equential \acr{M}odel-\acr{B}ased \acr{O}ptimi\sz{}ation (SMBO) methods. In contrast to traditional optimi\sz{}ation techniques, SMBO attempts to approximate the objective function with a \textit{surrogate model} (e.g. Gaussian process~\cite{EGO, SeqKriging}, random forest~\cite{SMBO-SMAC} or tree-structured Parzen estimator~\cite{TPE}). Each subsequent objective function evaluation is added to this surrogate model, refining the approximation.\@ \revision{The} next point at which to evaluate the objective function is determined by maximi\sz{}ing an \textit{acquisition function} (e.g.\ expected improvement (EI)~\cite{EIreview} or upper confidence bound (UCB)~\cite{UCB}) that combines exploration of the objective function and exploitation of the best candidate solution. This decision of where to evaluate the objective function according to the acquisition function and refining the surrogate model with this result forms the core loop of SMBO\@. One prominent member of this family of methods is known as Bayesian \acr{o}ptimi\sz{}ation (BO)~\cite{BOReview} with the Gaussian process (GP) surrogate model\footnote{BO using a GP is also known as \acr{E}fficient \acr{G}lobal \acr{O}ptimi\sz{}ation (EGO)~\cite{EGO} and \acr{S}equential Kriging \acr{O}ptimi\sz{}ation (SKO)~\cite{SeqKriging}.}. BO has been extensively studied~\cite{BOReview} and applied to a wide range of problems from hyperparameter optimi\sz{}ation for machine learning models~\cite{BOHyperparameters, BOHypTuning} to materials and chemical design~\cite{BODualPhaseSteel, BOChemicalProducts}.

%DOI LINK FOR REF BO HYPERPARAMETERS SEEMS BUSTED

% SLOWDOWN
BO, however, is no panacea and still has limitations to be aware of. Firstly, it is well-known that BO scales poorly as more observed points are added to the surrogate model (typically in the order $O(n^{3})$~\cite[Ch. 6]{Rasmussen} with $n$ observed points). Practically, this limits BO to lower-dimensional problems since a large number of observed points are necessary to model a high-dimensional objective function, leading to computationally expensive calculations. This also leads to BO slowing down significantly with subsequent algorithm iterations as more observations are added to the surrogate model~\cite{BEA}. Sparse approximations, such as the subset of data (SoD) or subset of regressors (SoR) approaches~\cite{SoD}, may alleviate this somewhat at the cost of surrogate fidelity.

%CONVERGENCE
%Not only does BO slow down computationally during convergence due to the poor scaling in terms of the number of observed points, but the exploration encouraged by the acquisition function also becomes a mixed blessing. In order for BO to converge closer to a solution, the rest of the function space needs to be sufficiently explored. This threshold becomes higher every time a better solution is found. During this final convergence phase the sole focus of BO would ideally be to exploit the small area around the candidate solution, not waste valuable function samples in areas that have already been explored sufficiently.

%KERNEL STATIONARITY
Secondly, the performance of BO when applied to a specific objective function is dependent on the chosen kernel function, which is a function that defines the family of functions that the GP surrogate model is able to represent. Choosing a single, generic kernel (as is done in the case of a black-box function where there is no prior knowledge of the function) reduces the effectiveness of BO in most situations~\cite{Spartan}. This is especially the case where the objective function is either non-stationary, exhibiting different behavi\ou{}r in different regions; or ill-conditioned, being much more sensitive to certain input variables than others~\cite{BO_ill_conditioned}.

Finally, while convergence for BO using EI was established by Vasquez and Bect~\cite{EGO_convergence_established}, explicit convergence rates depend on strong assumptions on the objective function and exist only for certain kernel functions using fixed hyperparameters, such as in the work of Bull~\cite{GP_covergence_asymptotes} or Srinivas et al.~\cite{GP_covergence_asymptotes_bandit}. Bull also showed that for BO using sequentially estimated hyperparameters (a common approach used during BO of black-box functions), BO may not converge at all. Even when these assumptions and criteria for theoretical convergence are met, BO exhibits practical numerical limitations, inhibiting its convergence characteristics. Computational instability in the GP model fitting procedure arises when there is a close proximity between any pair of observed points in the input space, resulting in a near-singular spatial covariance matrix. To address this instability, a common solution is to introduce a small ``nugget'' parameter $\delta$ as diagonal noise~\cite{nugget, DoEBook}. However, this reduces the rate and limit of convergence since an artificial level of noise has been implicitly imposed on the (possibly noiseless) objective function~\cite{BLOSSOM}. %The previously noted computational slowdown of BO may also make convergence to an arbitrary precision impractical as it would require possibly many more samples to reach a target $\epsilon$.

%In order to address these noted shortcomings of standard BO, this paper aims to develop a BO-based algorithm that \begin{enumerate*}[label=(\roman*)]
%  \item is resistant to computational slowdown,
%  \item is adaptable to non-stationary and ill-conditioned functions without kernel engineering, and
%  \item exhibits good convergence characteristics
%\end{enumerate*}. 

To summari\sz{}e, it is clear that standard BO has several shortcomings, namely that it \begin{enumerate*}[label= (\roman*)]
  \item experiences computational slowdown with additional algorithm iterations,
  \item is not well-suited to non-stationary and ill-conditioned functions, and
  \item exhibits poor convergence characteristics
\end{enumerate*}.

\paragraph{Related literature}

A recent avenue of research to mitigate the noted shortcomings of BO is to introduce a form of local focus, relaxing the global perspective of standard BO\@. The aim of this approach is to allow BO to leverage global information about the objective function to guide the search towar\ds{} the optimum and then locally exploit this solution, achieving faster convergence~\cite{Sasena_Local_Focus, Regis_Local_Focus}. The prominent algorithms that follow this approach may be loosely categori\sz{}ed into \revision{a set of} broad classes.

The first class of algorithms consist of \textit{hybrid} BO algorithms that add some mechanism to BO such that a switch is made to another optimi\sz{}ation method with better convergence characteristics at some point during the execution of the algorithm to exploit the best candidate solution. This switch point may be determined according to a metric such as expected gain~\cite{BEA}, estimated regret~\cite{BLOSSOM} or according to some budget-based heuristic~\cite{EGO-CMA}. Unfortunately, determining the optimal switching point is an optimi\sz{}ation problem in itself; switching either too early or too late can easily magnify the noted shortcomings of BO while reducing the sample efficiency gained by using BO in the first place.

Another class of algorithms combines BO with \textit{domain partitioning} of the full input space of the objective function into subdivisions. When these subdivisions are ranked, often based on the value of the acquisition function at the \centre{} of the subdivision~\cite{IMGPO}, promising areas can be exploited and further subdivided while others can be ignored in a branch-and-bound fashion. While these methods may have polynomial~\cite{BaMSOO} or even exponential~\cite{IMGPO} convergence guarantees, they require kernel engineering with prior knowledge of the objective function and scale similarly to standard BO\@.

A different class utili\sz{}e a combined \textit{local and global kernel}. The mechanism underpinning these methods is that a local kernel is used to model local changes in the objective function while a global kernel models the wider global structure. These kernels may be combined similarly to the piecewise-defined kernel of Wabersich and Toussaint~\cite{MGL} or the weighted linear kernel combination of Martinez-Cantin~\cite{Spartan}. While being superior to standard BO when applied to non-stationary problems~\cite{Spartan}, these algorithms still scale similarly to standard BO, converge slowly \revision{and have additional computational overhead due to the doubled kernel hyperparameters.}

\revision{A related class of algorithms are known as \textit{\acr{s}urrogate \acr{a}ssisted \acr{e}volutionary \acr{a}lgorithms} (SAEAs)~\cite{SAEA_review}, a class of methods that have been inspired by the well-established field of \acr{e}volutionary \acr{o}ptimi\sz{}ation (EO).\@ SAEAs combine a surrogate model to approximate the objective function (also known as the fitness function) with the traditional EO selection, crossover~\cite{SAEA_Crossover_Ex} and mutation~\cite{SAEA_Mutation_Ex} procedures to iteratively update a population of candidate solutions. By removing poor candidates from the iteratively updated population, local focus is injected into the algorithm. It is important to note that while SAEAs may not be formally considered as BO methods due to lacking an explicit acquisition function, they may be considered as such in practice. This is due to the similar, iteratively updated surrogate model and the crossover and mutation EO steps being analogous to the acquisition function found in BO.\@ While many SAEAs have proven to be an improvement over their standard EA counterparts~\cite{DTS-CMA-ES,AutoSAEA}, these methods tend to have relatively large population sizes which inhibit performance in lower dimensions, both due to the additional computational overhead and the longer initial selection phases to initiali\sz{}e these populations.}

The last class of methods uses an adaptive \textit{trust region} to encourage standard BO to exploit the local area surrounding the best candidate solution. This trust region acts as a moving window to constrain the acquisition function and, by extension, limit where the next point of the objective function will be evaluated. This window traverses the objective function and is frequently permitted to expand and contract, guided by a progress-dependent heuristic. Specifically, the window may expand during periods of progress and contract when progress stalls~\cite{SRSM, TuRBO, BADS, TRLBO}. Alternatively, similar to classical trust-region optimi\sz{}ation, the window size may be a function of the quality of the surrogate model's approximation of the objective function~\cite{TREGO, TRIKE, HSAGA}. This trust region modification relaxes the global optimi\sz{}ation of standard BO to be more akin to that of robust local optimi\sz{}ation, though in practice this approach is sufficient for most problems~\cite{TuRBO}. To regain a measure of global optimi\sz{}ation performance, these methods can be paired with a complementary mechanism such as alternating between a local and a global approximation~\cite{TREGO}, multistarts with a multi-armed bandit strategy~\cite{TuRBO} or restarts triggered by a minimum acquisition function threshold~\cite{TRIKE}. The TuRBO algorithm \revision{of} Eriksson et al.~\cite{TuRBO} \revision{and the subsequent TRLBO algorithm of Li et al.~\cite{TRLBO} also incorporate} automatic relevance determination (ARD)~\cite{ARD} through an anisotropic kernel to rescale the side length of the local trust region according to the local smoothness determined by the fitted kernel in a volume preserving transformation. This allows the local trust region to emphasi\sz{}e directions in which the objective function is smoother, a benefit when optimi\sz{}ing separable objective functions. However, the ARD \revision{as used in the TuRBO and TRLBO algorithms} is limited to directions defined solely by the coordinate axes. As a result, it may not be able to exploit objective functions that are separable in directions that are not along the coordinate axes. While trust-region-based BO methods scale better than standard BO and are more adaptive to non-stationary functions, these methods tend to have poor convergence characteristics and may even terminate prematurely when applied to ill-conditioned functions.

%Despite these advantages, these methods struggle to converge to an arbitrary precision.

In summary, while current methods may alleviate one or two of the three noted shortcomings of standard BO, none presents a solution that satisfactorily addresses all of them. Among the identified algorithm classes, trust-region-based BO exhibits the most promise for addressing all of the noted limitations. Therefore, we propose a novel algorithm which extends the trust region approach, offering a solution to alleviate all of the observed shortcomings of standard BO\@.

\paragraph{Summary of contributions}

%The novel method proposed in this paper, which we have denominated as Adaptive Local Gaussian Process Approximation (ALGPA) optimi\sz{}ation consists of Bayesian optimi\sz{}ation (BO) applied to an iteratively updated set of points in an adaptive trust region surrounding the best current candidate solution, with the size of this trust region changing in concert with the most likely length-scales of the Gaussian process (GP) model refitted at each algorithm iteration. The observations are also transformed such that the weighted principal components of the data (with more weight given to points with better output values) are aligned with the coordinate axes, allowing the chosen GP kernel function, squared exponential (SE) with automatic relevance determination (ARD), to uncover and exploit local separability. Points that fall outside of the trust region are also greedily discarded to avoid the computational slowdown of standard BO.

%The novel method proposed in this paper, which we have denominated as Adaptive Local Gaussian Process Approximation (ALGPA) optimi\sz{}ation consists of Bayesian optimi\sz{}ation (BO) - trust region - iterative rescaling - pc rotation - approximate hyp opt - observation discarding - algorithm 

This paper presents two novel extensions for trust-region-based BO\@. Firstly, we introduce an adaptive observation rescaling strategy based on the length-scales of the local GP surrogate with an SE kernel and ARD, instead of a heuristic, to allow for improved convergence characteristics. The second extension is a novel rotation of the trust region to align with the weighted principal components of the observed data. This rotation enables the maximum expressive power of the ARD kernel to model non-stationary and ill-conditioned objective functions. These two extensions are combined in a trust region-based BO framework with an iterative, approximate hyperparameter estimation approach and a subset-of-data (SoD) scheme~\cite{SoD} that greedily discards observations to mitigate computational slowdown, yielding the novel method proposed in this paper, which we denominate as the \ALGPAfull{} (\ALGPA{}) algorithm.

Using the well-known \acr{co}mparing \acr{c}ontinuous \acr{o}ptimi\sz{}ers (COCO) benchmarking software~\cite{COCO} with the noiseless \acr{b}lack-\acr{b}ox \acr{o}ptimi\sz{}ation \acr{b}enchmarking (BBOB) test suite~\cite{COCO_2009_doc}, we demonstrate \revision{the performance gains from the novel extensions of the trust region-based BO algorithm} using an ablation study and that the \ALGPA{} algorithm is a leading contender in the domain of expensive black-box function optimi\sz{}ation. For this comprehensive benchmark, \ALGPA{} significantly outperforms standard BO for nearly all tested scenarios and demonstrates superior performance compared to state-of-the-art black-box optimi\sz{}ation methods, particularly in the domain of unimodal and highly conditioned objective functions not typically associated with BO\@.

\paragraph{Structure of the paper} Section~\ref{sec:prelim} gives a brief overview of the requisite theoretical background on Gaussian processes and \revision{trust region-based Bayesian optimi\sz{}ation} that forms the basis of the proposed \ALGPA{} algorithm.\@ \revision{Section~\ref{sec:ALGPA_overview} provides a high-level overview of the \ALGPA{} algorithm and the role of the novel rescaling and rotation strategies.} \revision{Section~\ref{sec:obs_rescale} and~\ref{sec:obs_rotate} present the novel length-scale-based and principal-component-aligned transformations of the observed points. Section~\ref{sec:ALGPA_detail} presents the proposed \ALGPA{} \revision{algorithm} with the combined observation transformation, the iterative estimation of the length-scales used in this transformation, trust region definition, and observation discarding strategy.} Section~\ref{sec:results} applies the proposed algorithm to several well-known synthetic test functions as well as to the BBOB problem suite~\cite{COCO_2009_doc} from the well-known COCO benchmarking software~\cite{COCO}, allowing for a comparison to state-of-the-art black-box optimi\sz{}ation algorithms. Relevant proofs and derivations used in Section~\ref{sec:ALGPA_detail} are included in~\ref{subsec-app:lin_alg_offset_proof} to~\ref{subsec-app:kernel_derivs} \revision{and the full results obtained from the synthetic test functions and the COCO benchmarking software are given in~\ref{sec-app:synth} and~\ref{sec-app:coco}, respectively.}

%**********************************************
\section{Preliminaries}\label{sec:prelim}
%**********************************************

In this section, we provide the requisite theoretical frameworks used in the work presented in this paper. We start this section by reviewing \revision{Gaussian \acr{P}rocess (GP)} regression \revision{with} the squared exponential (SE) kernel function \revision{and} automatic relevance determination (ARD)\revision{, forming the basis of the novel length scale-based rescaling strategy.} \revision{We also} provide an overview of \revision{trust-region-based Bayesian optimi\sz{}ation,} the \revision{foundation of} the proposed \ALGPA{} algorithm.

%**********************************************
\subsection{Gaussian process regression}\label{subsec:gp}
%**********************************************

% \footnote{A naive way to describe a function is to represent it as an infinitely long vector, with each entry specifying the function value $f(x)$. This analogy, while crude, is surprisingly descriptive of the mechanism of a GP.}

A popular choice of surrogate model \revision{to approximate the objective function in BO is to use} a Gaussian process (GP), an extension of a multivariate Gaussian distribution to infinite dimensions~\cite[Ch. 1]{Rasmussen}. This GP model, fitted to $n$ observed $d$-dimensional input points $X = \{ \textbf{x}_{i} \in \mathbb{R}^{d} \,|\, i \in 1, 2, \ldots, n \} $ and observed function values $Y = \{ y_{i} = f(\textbf{x}_{i}) \in \mathbb{R} \,|\, i \in 1, 2, \ldots, n \} $ using a mean function $m(\cdot)$ and kernel function $k(\cdot, \cdot)$, can then be used for regression to estimate \revision{the unknown objective function} $f(\textbf{x})$:

\begin{equation} \label{eq:gp_regression}
    f(\textbf{x}) \sim \mathcal{GP}(m(\cdot), k(\cdot, \cdot) ; X, Y)
\end{equation}

\noindent and infer predictions $y_{*}$ for input points $\testpnt{x}$ using the key assumption of GPs that the posterior distribution for these unobserved points is given by the Gaussian distribution

\begin{equation} \label{eq:gp_pred}
    p(y_{*} \,|\, \testpnt{x}, X, Y) = \mathcal{N}(\mu_{\mathcal{GP}}(\testpnt{x}), \sigma^{2}_{\mathcal{GP}}(\testpnt{x})).
\end{equation}

To fit a GP to the observed data, a Gram matrix~\cite{Gram} $\textbf{K} \in \mathbb{R}^{n \times n}$ is constructed using a valid (symmetric and positive semidefinite~\cite[Ch. 4]{Rasmussen}) kernel function $k$, such that each entry satisfies

%\begin{equation} \label{eq:K}
%    \textbf{K} = \begin{bmatrix}
%    k(\textbf{x}_{1}, \textbf{x}_{1}) & k(\textbf{x}_{1}, \textbf{x}_{2}) & \hdots & k(\textbf{x}_{1}, \textbf{x}_{n})\\
%    k(\textbf{x}_{2}, \textbf{x}_{1}) & k(\textbf{x}_{2}, \textbf{x}_{2}) & \hdots & k(\textbf{x}_{2}, \textbf{x}_{n})\\
%    \vdots & \vdots & \ddots & \vdots \\
%    k(\textbf{x}_{n}, \textbf{x}_{1}) & k(\textbf{x}_{n}, \textbf{x}_{2}) & \hdots & k(\textbf{x}_{n}, \textbf{x}_{n})\\
%    \end{bmatrix}.
%\end{equation}

\begin{equation}
    \textbf{K} = {\bigl[  k(\textbf{x}_{i}, \textbf{x}_{j}) \bigr]}_{1 \leq i,j \leq n}. \label{eq:K}
\end{equation}

Using the matrix $\textbf{K}$ and the column vector $\textbf{y} = {[ y_{i} ]}^{\top}_{1\leq i \leq n}$, the equations for the predicted mean $\mu_{\mathcal{GP}}$ and variance $\sigma^{2}_{\mathcal{GP}}$ of the GP at a given test point $\testpnt{x}$ from (\ref{eq:gp_pred}) are given by~\cite[Ch. 2]{Rasmussen}:

%\begin{equation} \label{eq:gp_mean_pred}
%    \mu_{\mathcal{GP}}(\testpnt{x}) = m(\testpnt{x}) + \textbf{k}^{\top}_{*} \textbf{K}^{-1} (\textbf{y} - \textbf{m}(X))
%\end{equation}

%\noindent and

%\begin{equation} \label{eq:gp_var_pred}
%    \sigma^{2}_{\mathcal{GP}}(\testpnt{x}) = k(\testpnt{x}, \testpnt{x}) - \textbf{k}^{\top}_{*} \textbf{K}^{-1} \textbf{k}_{*},
%\end{equation}

\begin{align}
    \mu_{\mathcal{GP}}(\testpnt{x})        & = m(\testpnt{x}) + \textbf{k}^{\top}_{*} \textbf{K}^{-1} (\textbf{y} - \textbf{m}(X)), \label{eq:gp_mean_pred} \\
    \sigma^{2}_{\mathcal{GP}}(\testpnt{x}) & = k(\testpnt{x}, \testpnt{x}) - \textbf{k}^{\top}_{*} \textbf{K}^{-1} \textbf{k}_{*},\label{eq:gp_var_pred}
\end{align}

\noindent where

\begin{equation}
    \textbf{k}_{*} = \begin{bmatrix} k(\textbf{x}_{1}, \testpnt{x}) & k(\textbf{x}_{2}, \testpnt{x}) & \ldots & k(\textbf{x}_{n}, \testpnt{x})
    \end{bmatrix}^{\top}. \nonumber
\end{equation}

\revision{A common choice of kernel function} is the \acr{s}quared \acr{e}xponential (SE) kernel with \acr{A}utomatic \acr{R}elevance \acr{D}etermination (ARD). This kernel contains a \emph{characteristic length-scale} parameter $\ell$ which is proportional to the smoothness of the fitted GP model, with ARD extending the SE kernel with an independent length-scale for each coordinate direction according to

\begin{gather}\label{eq:sqexp_vector_form}
    k_{\textrm{SE}}(\textbf{x}^{}_{p}, \textbf{x}^{}_{q}) = \sigma_{f}^{2} \cdot \exp (-\frac{1}{2}{(\textbf{x}^{}_{p} - \textbf{x}^{}_{q})}^{\top} \Lambda^{-1} (\textbf{x}_{p} - \textbf{x}_{q})) + \sigma^{2}_{n} \delta^{}_{pq}, \\ \textrm{where} \quad \Lambda^{-1} = \textrm{diag}{(\ell^{2}_{1}, \ell^{2}_{2}, \ldots, \ell^{2}_{d})}^{-1} \nonumber
\end{gather}

\noindent where $\sigma_{f}^{2}$ and $\sigma_{n}^{2}$ are the signal and noise variances, respectively. The hyperparameters $\boldsymbol{\theta}$ of this kernel are often chosen through maximum likelihood estimation (possibly incorporating some hyperparameter prior~$p(\boldsymbol{\theta})$)

\begin{equation}\label{eq:argmax_thetas}
    \boldsymbol{\theta}^{*} = \operatorname*{argmax}_{\boldsymbol{\theta}} (\log p(Y \,|\, X, \boldsymbol{\theta})\, \revision{+} \log p(\boldsymbol{\theta})),
\end{equation}

\noindent where

\begin{equation}\label{eq:gp_log_lik}
    \log p(Y \,|\, X, \boldsymbol{\theta}) = -\frac{1}{2} \textbf{y}^{\top} \textbf{K}^{-1} \textbf{y} -\frac{1}{2} \log |\textbf{K}| - \frac{n}{2} \log 2\pi.
\end{equation}

During the calculation of (\ref{eq:gp_mean_pred}), (\ref{eq:gp_var_pred}) and (\ref{eq:gp_log_lik}), determining the inverse matrix $\textbf{K}^{-1}$ tends to dominate the computation time\footnote{In practical GP implementations, $\textbf{K}^{-1}$ is rarely used directly. Since $\textbf{K}$ is known to be symmetric and positive semidefinite due to the use of a valid kernel function~\cite[Ch. 4]{Rasmussen}, the Cholesky decomposition of the matrix $\textbf{K}$ can be used in conjunction with back substitution \revision{for} complexity and stability benefits.} due to matrix inversion being of the order $O(n^{3})$ for an $n \times n$ matrix~\cite[Ch. 6]{Rasmussen}. This is the principal reason why standard BO typically scales poorly with an increasing number of observations and dimensions. \\

\todo[inline]{``Squared exponential kernel function and hyperparameter estimation'' section combined with previous section.}

\subsection{\revision{Trust-region-based Bayesian optimi\sz{}ation}}\label{subsec:bo}
%**********************************************

Bayesian \acr{o}ptimi\sz{}ation attempts to find the argument $\minpnt{x} \in \Omega$ that minimi\sz{}es\footnote{An arbitrary choice, since any minimi\sz{}ation problem can be converted to a maximi\sz{}ation problem with a simple sign change in the objective function.} a given scalar, bounded, black-box objective function $f(\textbf{x}): \mathbb{R}^{d} \to \mathbb{R}$, where $\Omega \subset \mathbb{R}^{d}$ is the set of all possible arguments subject to the specified constraints. In this paper, we assume that $f$ is observable exactly (i.e., with no noise) and parameteri\sz{}e $\Omega$ using a Cartesian product with bounding values $\Omega^{\min}$ and $\Omega^{\max}$ for each dimension. Formally, we can state this optimi\sz{}ation task as calculating

\begin{gather}\label{eq:black_box_problem}
    \minpnt{x} = \operatorname*{argmin}_{\testpnt{x} \in \Omega} f(\testpnt{x}), \quad \textrm{where} \quad \Omega = \prod^{d}_{i=1}[ \Omega^{\min}_{i}, \Omega^{\max}_{i} ].
\end{gather}

%In essence, BO iteratively builds a GP surrogate model approximation of the objective function $f$ using points selected according to an acquisition function. The GP is chosen as a surrogate model since it is cheap to evaluate compared to the objective function and that the GP can yield both an estimate of the objective function (mean of the GP prediction) and the uncertainty of this estimate (variance of GP prediction). Using the estimate of the objective function (and uncertainty thereof), an acquisition function is constructed which formali\sz{}es this trade-off between exploitation (low $\mu_{\mathcal{GP}}(\testpnt{x})$) and exploration (high $\sigma^{2}_{\mathcal{GP}}(\testpnt{x})$).

% In essence, BO iteratively builds a GP surrogate model approximation of the
% objective function $f$ using points selected according to an acquisition
% function. The GP is chosen as a surrogate model since it is cheap to evaluate
% compared to the objective function and the fact that the GP can yield both an
% estimate of the objective function and the uncertainty of this estimate through
% the mean and variance of the GP prediction. Using the estimate of the objective
% function (and uncertainty thereof), an acquisition function is constructed
% which formali\sz{}es the trade-off between exploitation (low GP mean) and
% exploration (high GP variance) of the objective function.

\revision{In essence, BO iteratively builds a GP surrogate model approximation of the objective function $f$ using points selected according to an acquisition function.} One popular choice of acquisition function is the expected improvement (EI) function~\cite{EIreview}

\begin{gather}\label{eq:EI}
    \alpha_{\textrm{EI}}(\testpnt{x} ;\,\mathcal{GP}) = \sigma_{\mathcal{GP}}(\testpnt{x})[ z\Phi(z) + \phi(z) ], \quad \textrm{where} \quad z = \frac{y_{\min} - \mu_{\mathcal{GP}}(\testpnt{x})}{\sigma_{\mathcal{GP}}(\testpnt{x})} 
\end{gather}

\noindent \revision{with} $\phi(z)$ and $\Phi(z)$ \revision{as} the probability density function and cumulative density function of a univariate standard normal distribution, respectively.

\revision{BO may be initiali\sz{}ed with an initial set of input points before the main loop of BO starts, known as the \acr{D}esign of \acr{E}xperiment (DoE). These points are often selected using random sampling or more informed methods such as Latin hypercube sampling~\cite{DoEBook} based on the objective function constraints set $\Omega$.}

In general, BO does not have a well-defined, general stopping criterion, a property shared with other derivative-free optimi\sz{}ation methods. This is in contrast to gradient-based methods, where the norm of the gradient can be used to ensure first-order stationarity of the solution. Consequently, a maximum objective function \revision{evaluation limit} or minimum decrease condition is often used as a termination condition for BO\@.

\revision{Trust-region-based BO methods extend standard BO by bounding the acquisition function with an iteratively updated trust region $\Omega_{\textrm{TR}}$ to limit where subsequent input points are sampled. At each algorithm iteration, the size of this trust region may be updated by a constant factor~\cite{BADS, TREGO, TRIKE} or for each dimension independently~\cite{TuRBO, SRSM}, according to the new observations. The trust region is also often \centered{} on the current minimum candidate $\minpnt{x} = \{ \textbf{x}_{i} \in X \,|\, y_{i} \leq y_{j}, \forall y_{j} \in Y \}$.} 

The \revision{\TRBO{}} algorithm is summari\sz{}ed in Alg.~\ref{alg:BO_opt}, where the inner loop of selecting an input point that maximi\sz{}es the acquisition function $\alpha$ \revision{bounded by the trust region $\Omega_{\textrm{TR}}$,} evaluating the objective function $f$ at this point, adding the result to the isomorphic sets of observed points $X$ and $Y$, \revision{updating the size of the trust region,} and refitting the model with the augmented observations is given in \revision{lines $5-10$.}

\renewcommand*\Call[2]{\textproc{#1} (#2)}
\begin{algorithm}[h]
    \caption{\revision{Trust-region-based Bayesian optimi\sz{}ation}}\label{alg:BO_opt}
    \begin{algorithmic}[1]
        \Require{} Objective function $f$, Acquisition function $\alpha$, Bounds $\Omega$, Design of experiment DoE

        \vphantom{text}
        \State{} $\revision{X_{0}}$ $\gets$ $\textrm{DoE}(\Omega)$ \Comment{Select initial input points using DoE and bounds}
        \State{} $\revision{Y_{0}}$ $\gets$ $\{ f(\textbf{x}) \,|\, \textbf{x} \in \revision{X_{0}} \}$ \Comment{Evaluate initial input points from DoE}
        \revision{\State{} $\Omega_{\textrm{TR}}$ $\gets$ $\Omega_{\textrm{TR}_{0}}$ \Comment{Initiali\sz{}e trust region}}
        \While{\textbf{not} convergence criterion satisfied}
        \State{} $\mathcal{GP}$ $\gets$ $\mathcal{GP}(m(\cdot), k(\cdot, \cdot) ; X, Y)$ \Comment{Fit GP with $X$ and $Y$}
        \State{} $\textbf{x}_{\alpha}$ $\gets$ $\operatorname*{argmax}_{\subalign{\testpnt{x} &\in \,\Omega_{\textrm{TR}} \\ \testpnt{x} &\in \, \Omega }}$\Call{$\alpha$}{$\testpnt{x} ;\,\mathcal{GP}$} \Comment{Maximi\sz{}e acquisition function}
        \State{} $y_{\alpha}$ $\gets$ $f(\textbf{x}_{\alpha})$ \Comment{Evaluate suggested input point}
        \State{} $X$ $\gets$ $X \cup \{ \textbf{x}_{\alpha} \}$ \Comment{Add observed input point to set}
        \State{} $Y$ $\gets$ $Y \cup \{ y_{\alpha} \}$ \Comment{Add observed output point to set}
        \State{} \revision{$\Omega_{\textrm{TR}}$ $\gets$ $\Omega_{\textrm{TR}}(X, Y)$ \Comment{Update trust region}}

        \EndWhile{}
        \State{} \Return{} $\minpnt{x}, y_{\min}$ \Comment{Return minimum candidate}
    \end{algorithmic}
\end{algorithm}

\revision{This algorithm forms the foundation for the proposed \ALGPA{} algorithm in the next section when combined with the novel length-scale based observation rescaling and weighted principal component rotation strategies of this paper.}
% proposed algorithm's primary change is an adaptive rescaling- and rotation
% strategy based on the surrogate GP model for $X$ and $Y$, as well as an
% iteratively updated trust region to bound the acquisition function and a greedy
% strategy for discarding observations from $X$ and $Y$ that fall \revision{outside} this
% trust region.

%**********************************************
\section{\revision{Overview of the \ALGPA{} algorithm}}\label{sec:ALGPA_overview}
%**********************************************

The novel method proposed in this paper, which we denominate as the \emph{\ALGPAfull{}} (\ALGPA{}) algorithm, \revision{is given in Fig.~\ref{fig:algpa_flowchart}.} LABCAT follows the example of other \revision{\TRBO{}} algorithms by incorporating a local trust region surrounding the current minimum candidate solution to bound the acquisition function maximi\sz{}ation during the determination of subsequent input points~\cite{SRSM, TuRBO, BADS, TRLBO, TREGO, TRIKE, HSAGA}. What distinguishes \ALGPA{} is that the size of this local trust region is selected to be directly proportional to the length-scales of the GP fitted to the observed data instead of according to a progress-based or sufficient decrease heuristic. The trust region is also rotated to align with the weighted principal components of the observed data, allowing the \revision{side lengths} of the trust region to change \revision{independently} along arbitrary directions, not just along the coordinate axes. Additionally, the local trust region is used to greedily discard observed points \revision{outside} the trust region, in contrast to the noted methods that either retain all points or employ a significant subset of the evaluation history.

\newcommand{\flowsize}{0.65} %Standard

\begin{figure}[h]
    \centering
    \includegraphics[width = \flowsize\linewidth]{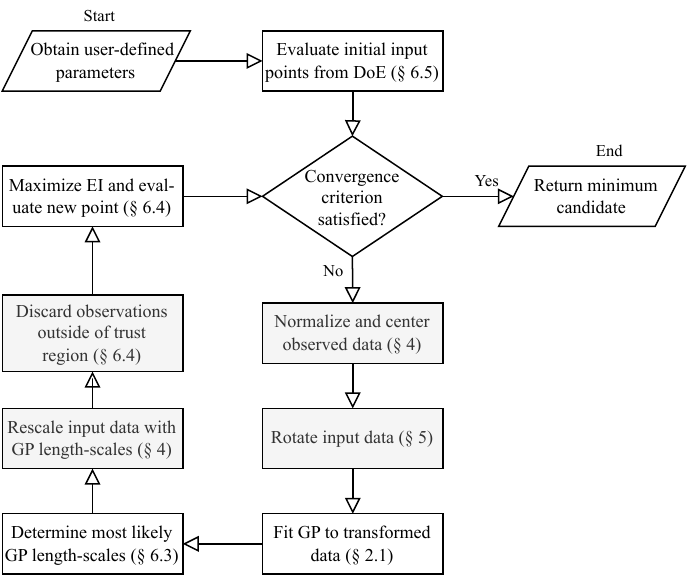}
    \caption{A flowchart of the \ALGPA{} algorithm.\ \revision{The primary added components of the \ALGPA{} algorithm, compared to the standard \revision{\TRBO{} described in Alg.~\ref{alg:BO_opt}}, is given by the shaded areas. A full mathematical description of the \ALGPA{} algorithm in given in Alg.~\ref{alg:algpa}.}}\label{fig:algpa_flowchart}
\end{figure}

Inspecting Fig.~\ref{fig:algpa_flowchart}, the \ALGPA{} algorithm enters a modified version of the standard \revision{\TRBO{}} loop after evaluating the initial set of input points. These modifications consist of transforming the observed data and discarding observations that fall \revision{outside} this trust region. At the start of the modified \revision{\TRBO{}} loop, the current set of observed inputs are \recentered{} on the current minimum candidate and rotated to align the principal components of the observed input points (weighted by the corresponding normali\sz{}ed observed output values) with the coordinate axes. Using this \recentered{}, rotated and normali\sz{}ed observed input and output data, a GP with an SE kernel and ARD is fitted. The MAP estimate for the length-scales of this kernel is used to rescale the observed input data, updating the size of the current trust region. To prevent the unbounded growth in complexity of the GP model, we make use of an approximate model fitted to the local subset of observations \revision{inside} the current trust region. Finally, the EI acquisition function is maximi\sz{}ed, bounded by the trust region, to determine the next sampled point.\@ \revision{The next three sections describe the novel length-scale-based rescaling, weighted principal component rotation and detailed description of the \ALGPA{} algorithm, respectively.}

% The rest of this section describes each of the salient components of the
% \ALGPA{} algorithm: \revision{\begin{enumerate*}[label= (\alph*)]
%     \item the transformations applied to the observed data,
%     \item the calculation of kernel hyperparameters for the fitted GP,
%     \item the definition of the trust region used for the bounding of the EI maximi\sz{}ation and or discarding observations,
%     \item the chosen convergence criteria and 
%     \item choice of initial observed points are presented.
%   \end{enumerate*}. This section concludes with an overview of the
% \ALGPA{} algorithm, a note on the comutational complexity of this method and an illustrative example to provide a visual contrast between the \ALGPA{} algorithm and standard trust region methods as well as providing a qualitative example of the advantages associated with the novel trust region rescaling and -rotation strategies.}

%**********************************************
\section{\revision{Length-scale-based rescaling}}\label{sec:obs_rescale}
%**********************************************

\todo[inline]{Section refactored from ``Transformed representation'' section.}

During the course of a \revision{\TRBO{}} algorithm, the size of the trust region necessarily shrinks as the algorithm converges to a solution. This shrinking trust region may lead to observations being clustered together or ill-conditioned, \revision{causing} the near singularity of the spatial covariance matrix $\textbf{K}$ \revision{noted in Sec.~\ref{sec:intro}.} To \revision{prevent} this, we transform the observations to a transformed space, \revision{using a novel, local length-scale-based rescaling,} in which we perform minimi\sz{}ation actions that work well for a much smaller range of objective function values compared to the original space. In this transformed space, the observations remain well-conditioned and -distributed, even if the corresponding points in the objective function space are not.

To \revision{implement this transformation,} we construct an invertible mapping between the observed data $X, Y$ and a transformed representation $\tra{X}, \tra{Y}$ of the same dimensionality, that is, $X \subset \mathbb{R}^{d} \leftrightarrow \tra{X} \subset \mathbb{R}^{d}$ and $Y \subset \mathbb{R} \leftrightarrow \tra{Y} \subset \mathbb{R}$, according to invariant properties that are preserved at each algorithm iteration. For notational convenience, we indicate the image of a variable under this transformation by using the primed counterpart of the variable and vice versa for the preimage, for example, the variable $x$ and the image of this variable under the transformation $\tra{x}$. Furthermore, in this \revision{paper the} set $\tra{X}$ can be collected into and decomposed from a matrix $\tra{\textbf{X}} \in \mathbb{R}^{d \times n}$, where the $i^{\textrm{th}}$ column of $\tra{\textbf{X}}$ corresponds to $\tra{\textbf{x}_{i}}$ in $\tra{X}$.

For the transformation of the input points from $X$ to $\tra{X}$, we construct the following affine transformation between the elements of these sets

\begin{equation}\label{eq:rescale_transform}
    \revision{\textbf{x}_{i} = \textbf{S} \tra{\textbf{x}_{i}} + \revision{\textbf{c}} \,\,\, \forall i \in \{ 1, 2, \ldots, n \},}
\end{equation}

\noindent using a \revision{(diagonal)} scaling matrix $\textbf{S}$ and offset vector \revision{$\textbf{c}$}. This transformation naturally extends to a relation expressed using the matrices $\textbf{X}$ and $\tra{\textbf{X}}$

\begin{equation}\label{eq:rescale_transform_matrices}
    \revision{\textbf{X} = \textbf{S} \tra{\textbf{X}} + \revision{\textbf{c}} \textbf{1}^{\top}_{n},}
\end{equation}

\noindent with the transformation parameters $\textbf{S}$ and $\revision{\textbf{c}}$ calculated according to three invariant properties. These invariant properties are defined as the following:

\begin{enumerate}[label= (\roman*)]
    \item\label{li:rescale_invariant_1} The transformed \revision{minimum candidate} $\tra{\minpnt{x}}$ is at the origin

          \begin{gather}\label{eq:rescale_invariant_1}
              \tra{\minpnt{x}} =\begin{bmatrix} 0 & \ldots & 0\\\end{bmatrix}^{\top}.
          \end{gather}

    \item\label{li:rescale_invariant_2} $\tra{X}$ is scaled such that the most likely length-scales $\boldsymbol{\ell}^{*}$ for a GP that has been fitted with $\tra{X}$ and $\tra{Y}$ are unity, or

          \begin{equation}\label{eq:rescale_invariant_2}
              \boldsymbol{\ell}^{*} = ( 1, 1, \ldots, 1 ).
          \end{equation}
\end{enumerate}

Similarly to (\ref{eq:rescale_transform}), we construct a relation between the
observed outputs $Y$ and a set of transformed outputs $\tra{Y}$ as

\begin{equation}\label{eq:output_transform}
    y_{i} = a \cdot \tra{y_{i}} + \revision{b}
    \,\,\, \forall i \in \{ 1, 2, \ldots, n \}
\end{equation}

\noindent according to an additional invariant property: \begin{enumerate}[label= (\roman*)]\setcounter{enumi}{2} 
    
    \item\label{li:output_invariant} The current minimum observed output value $y_{\min}$ and the maximum observed value $y_{\max}$ are min-max normali\sz{}ed

          \begin{equation} \label{eq:output_invariant}
              \quad \tra{y_{\min}} = 0 \quad \textrm{and} \quad \tra{y_{\max}} = 1.
          \end{equation}

\end{enumerate}

As can be seen in \revision{the flowchart of} Fig.~\ref{fig:algpa_flowchart} and in the example of Fig.~\ref{fig:rescale_transform_summary}, these invariant properties are preserved through the \recentering{} and the rescaling of the observed input data according to the local length-scales of a GP \revision{with an ARD kernel.}

\newcommand{\tempsize}{0.4} %Review
\captionsetup[subfigure]{justification=centering}
\begin{figure}[ht!]
    \centering
    \subfloat[\label{subfig:length_X_Y}]{\includegraphics[width = \tempsize\linewidth]{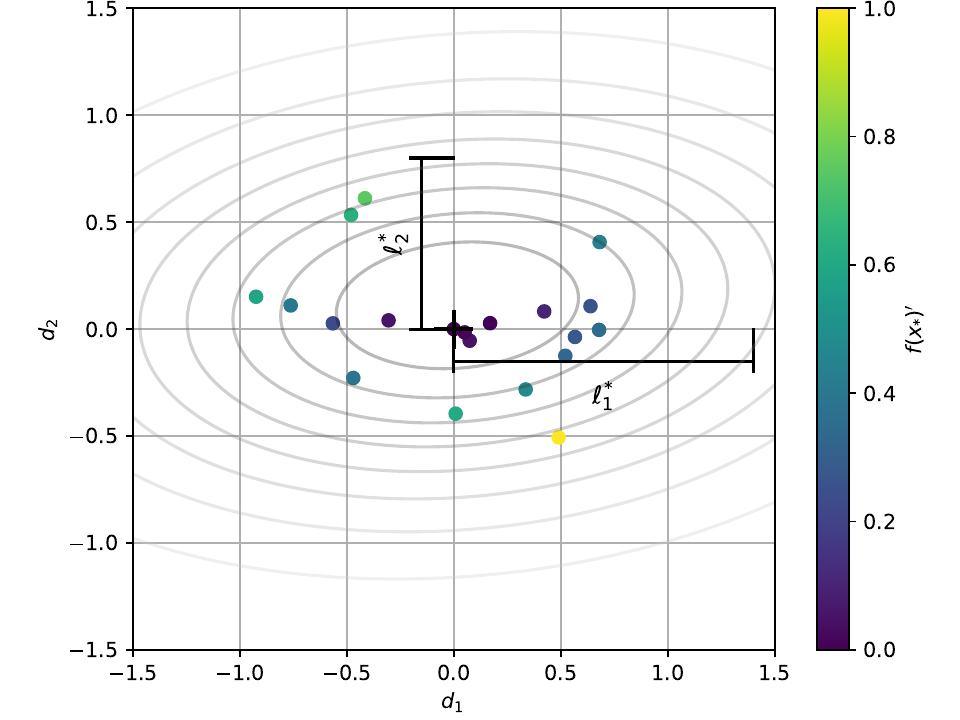}}
    \subfloat[\label{subfig:transformed_X_Y}]{\includegraphics[width = \tempsize\linewidth]{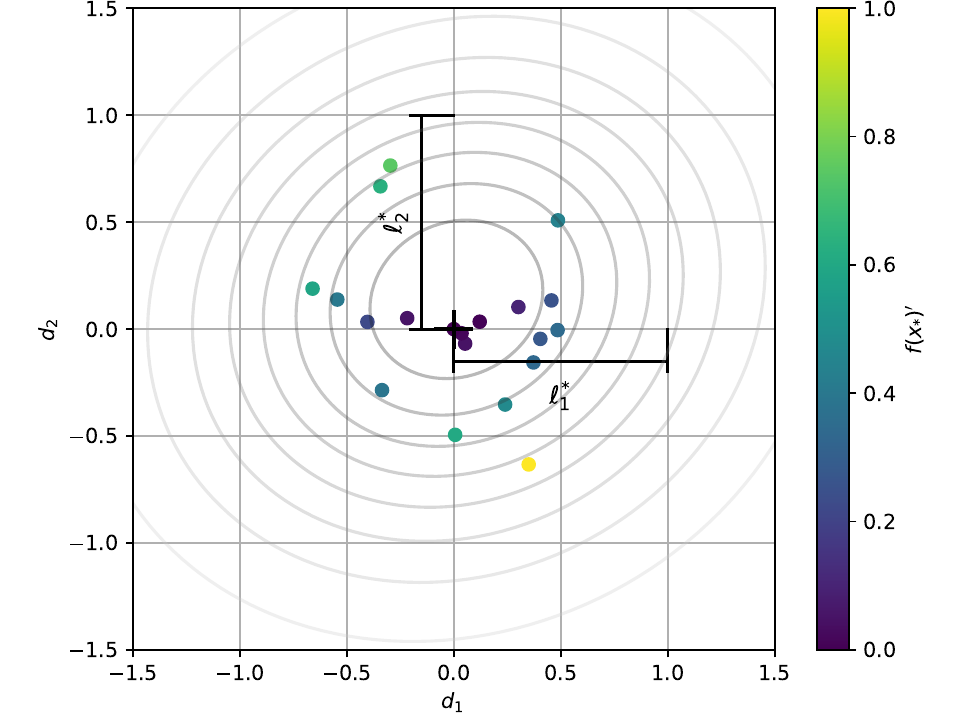}}

    \caption{\revision{A visuali\sz{}ation of enforcing the invariant properties (described by~\ref{li:rescale_invariant_1}--\ref{li:output_invariant}) on a number of observations from an arbitrary function, where the observed output values are represented using a col\ou{}r map. The data is (\hyperref[subfig:length_X_Y]{a}) \centered{} on the minimum candidate (marked with a $+$), the output values are normali\sz{}ed and the most likely length-scales ($\ell^{*}_{1}$, $\ell^{*}_{2}$) for a GP fitted to the data are shown. Using these length-scales, (\hyperref[subfig:transformed_X_Y]{b}) the input data is rescaled such that these length-scales equal unity, with all invariant properties now preserved.}}\label{fig:rescale_transform_summary}
\end{figure}

\revision{As previously stated,} to preserve the invariant properties of the transformed data defined in~\ref{li:rescale_invariant_1},~\ref{li:rescale_invariant_2} and~\ref{li:output_invariant} requires three steps. Firstly, the invariant property of the output data from~\ref{li:output_invariant} is preserved by transforming $\tra{Y}$ using the values of $y_{\min}$ and $y_{\max}$ with min-max normali\sz{}ation~\cite{MinMaxNormalization}

\begin{equation}
    \tra{Y} = \biggl \{ \frac{y_{i} - y_{\min}}{y_{\max} - y_{\min}} \,\big|\, y_{i} \in Y \biggr \}
\end{equation}

\noindent and by setting the output offset and scaling coefficients to

\begin{equation}
    a = y_{\max} - y_{\min} \quad \textrm{and} \quad \revision{b} = y_{\min}.
\end{equation}

For the next step, we enforce invariant property~\ref{li:rescale_invariant_1} through \recentering{} $X$ with the current minimum candidate $\minpnt{x}$, subtracting this value from every element of $X$. This is achieved using the transform

\begin{equation}\label{eq:input_cen_original}
    \cen{\textbf{X}} = \textbf{X} - \minpnt{x}\textbf{1}^{\top}_{n}
\end{equation}

\noindent and by setting the offset vector

\begin{equation}\label{eq:b_update}
    \revision{\textbf{c}} = \minpnt{x}.
\end{equation}

Finally, we fit a GP to $\cen{X}$ and $\tra{Y}$ using an SE kernel with ARD as in (\ref{eq:sqexp_vector_form}) and determine the most likely length-scales $\boldsymbol{\ell}^{*}$ using (\ref{eq:argmax_thetas}). Using $\boldsymbol{\ell}^{*}$, invariant property~\ref{li:rescale_invariant_2} is preserved by scaling the \revision{\centered} input data $\tra{\cen{X}}$ according to

\begin{gather}\label{eq:ell_diag_inv_mult}
    \tra{X} = \textbf{L}^{-1}\cen{\textbf{X}}
\end{gather}

\noindent where the most likely length-scales are collected into a scaling matrix

\begin{equation}\label{eq:ell_diag_inv}
    \textbf{L}^{-1} = {\diag(\ell_{1}^{*}, \ell_{2}^{*}, \ldots, \ell_{n}^{*})}^{-1}.
\end{equation}

\noindent and the scaling matrix $\textbf{S}$ in (\ref{eq:rescale_transform}) is set to

\begin{equation}\label{eq:s_update}
    \textbf{S} = \textbf{L},
\end{equation}

\noindent with no need to recalculate the kernel matrix $\textbf{K}$ for \revision{the GP fitted to $\cen{X}$,} $\tra{Y}$ and $\boldsymbol{\ell}^{*}$, as both the observed inputs and length-scales have been scaled by the same factor. Inspecting~(\ref{eq:sqexp_vector_form}) and factori\sz{}ing the individual length-scales, this rescaled GP is now equivalent to a GP fitted to $\tra{X}$ and $\tra{Y}$ with unit length-scales.

\revision{When used in a \TRBO{} framework and performing the length-scale-based rescaling at each algorithm iteration, this transformation can allow for improved convergence characteristics and better performance on select separable or ill-conditioned functions. However, the full potential of the rescaling is achieved when combined with the trust region rotation of the next section, extending the improved performance to more general objective functions.} 

%**********************************************
\section{\revision{Principal-component-based rotation}}\label{sec:obs_rotate}
%**********************************************

\todo[inline]{Section refactored from ``Transformed representation'' section.}

\revision{As noted in Sec.~\ref{subsec:bo}, several \revision{\TRBO{}} methods allow the side lengths of the trust region to be changed independently for each dimension~\cite{TuRBO, TRLBO, SRSM}. This mechanism allows the trust region to expand in directions for which the objective function may be smoother and vice versa, making these algorithms well suited to separable functions. The main shortcoming of this approach is that the trust region resizing is limited to the directions defined by the coordinate axes. Ideally, the trust region should be allowed to rotate and dynamically align itself with directions of separability in the objective function, for example, local valleys.}

\revision{To implement this rotation, similarly to the transformation defined in~(\ref{eq:rescale_transform}), we construct a \revision{new} affine transformation to determine a transformed representation $\tra{X}$\footnote{\revision{Note that, for notational convenience, we repeat the use of the prime to indicate observed input variables ($X \leftrightarrow \tra{X}$) under the new transformation in~(\ref{eq:rotate_transform_matrices}). This use is also repeated in Sec.~\ref{sec:ALGPA_detail} to refer to the combined transformation in~(\ref{eq:input_transform_matrices}).}} of the observed data $X$ in which the weighted principal components are aligned with the coordinate axes. This new transformation is defined as}

\begin{equation}\label{eq:rotate_transform_matrices}
    \revision{\textbf{X} = \textbf{R} \tra{\textbf{X}} + \revision{\textbf{c}} \textbf{1}^{\top}_{n},}
\end{equation}

\noindent \revision{with the output data normalized according to invariant property~\ref{li:output_invariant}, the input data \centered{} and transformation parameter $\revision{\textbf{c}}$ calculated according to invariant property~\ref{li:rescale_invariant_1}, and the \revision{orthogonal rotation matrix} $\textbf{R}$ calculated according to an additional invariant property}

\begin{enumerate}[label= (\roman*)]\setcounter{enumi}{3} 
    \item\label{li:rotate_invariant} The weighted principal components, described by the orthogonal rotation matrix $\textbf{U}$ of the observed input data $X$ with more weight given to values of $X$ with a lower corresponding value in $Y$, are transformed to be aligned with the coordinate axes of $\tra{X}$ (up to reflection)
    
        \begin{equation}\label{eq:rotate_invariant}
            \tra{\textbf{U}} = \diag(\pm 1, \ldots, \pm 1).
        \end{equation}

    \end{enumerate}

\revision{An example of the process to preserve invariant properties~\ref{li:rescale_invariant_1},~\ref{li:output_invariant} and~\ref{li:rotate_invariant} used to define (\ref{eq:rotate_transform_matrices}) is given in Fig.~\ref{fig:rotate_transform_summary}.}

\newcommand{\tempsizew}{0.4} %Review

\captionsetup[subfigure]{justification=centering}
\begin{figure}[ht!]
    \centering
    % \subfloat[\label{subfig:original_X_Y}]{\includegraphics[width = \tempsizew\linewidth]{figures/Section3/fig/invariant_ex/img_1.pdf}}
    \subfloat[\label{subfig:centered_X_Y}]{\includegraphics[width = \tempsizew\linewidth]{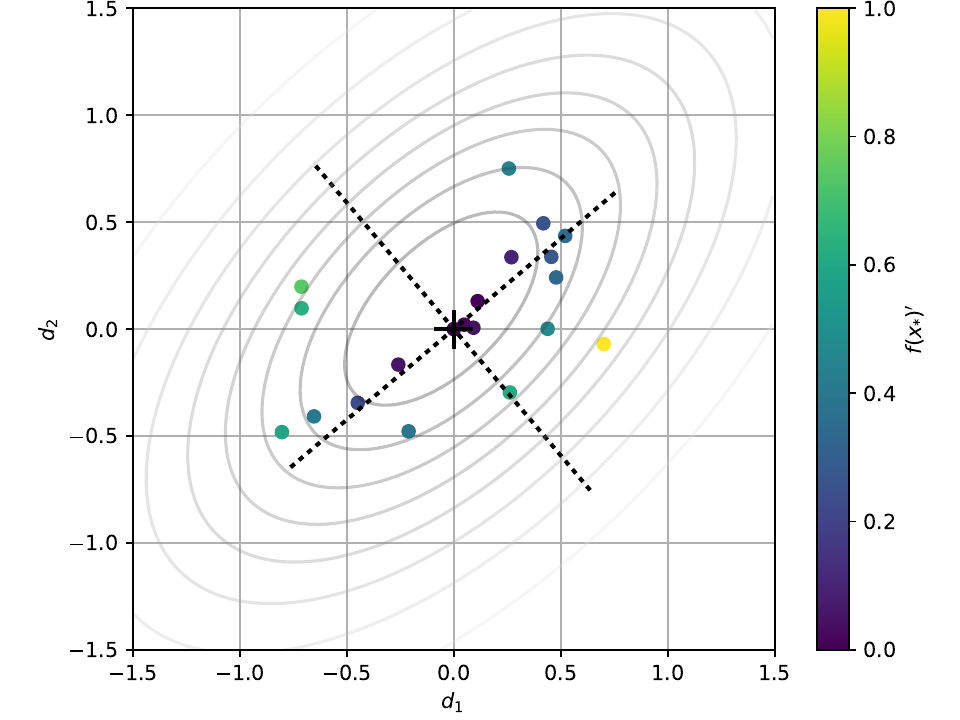}}
    \subfloat[\label{subfig:rotated_X_Y}]{\includegraphics[width = \tempsizew\linewidth]{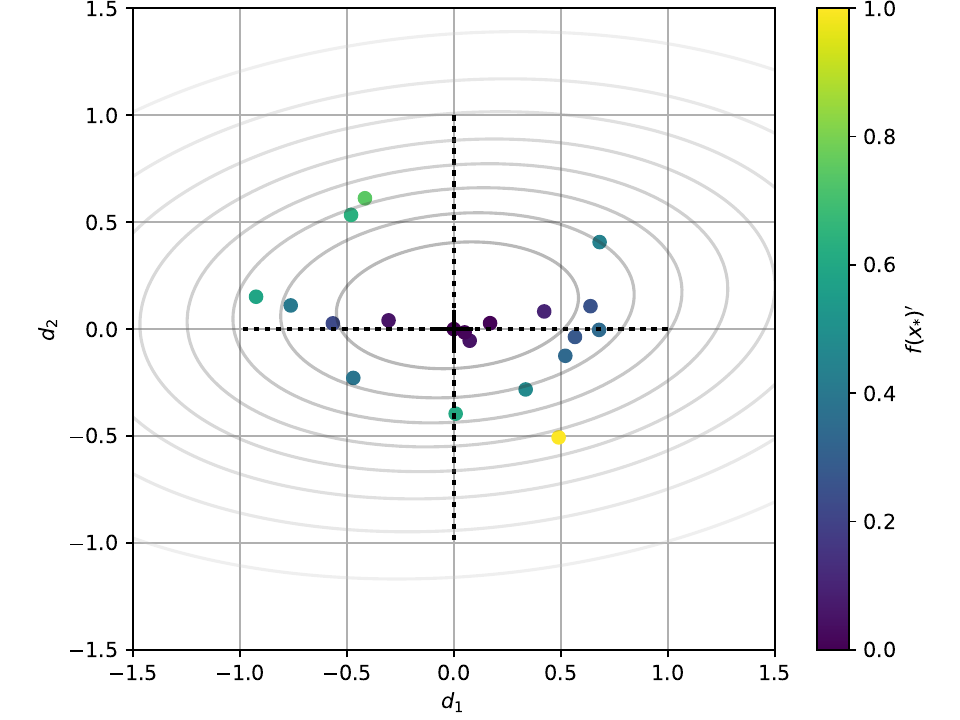}}

    \caption{\revision{A visuali\sz{}ation of enforcing the invariant properties (described by~\ref{li:rescale_invariant_1},~\ref{li:output_invariant} and~\ref{li:rotate_invariant}) on a number of observations from an arbitrary function, where the observed output values are represented using a col\ou{}r map. The data is (\hyperref[subfig:centered_X_Y]{a}) \centered{} on the minimum candidate (marked with a $+$), the output values are normali\sz{}ed and the weighted principal components are shown. Using these principal components, (\hyperref[subfig:rotated_X_Y]{b}) the input data is rotated such that these principal components are aligned with the coordinate axes, with all invariant properties now preserved.}}\label{fig:rotate_transform_summary}
\end{figure}

To enforce invariant property~\ref{li:rotate_invariant}, we use the sample-wise weighted, \centered{} input data $\textbf{X}_{\textrm{cen}} \textbf{W}$ \revision{(\centered{} using~(\ref{eq:input_cen_original}))} to calculate the weighted principal components $\textbf{U}$. We calculate these weighted principal components $\textbf{U}$ using the singular value decomposition (SVD)~\cite{SVD} with the following form

\begin{equation}
    \cen{\textbf{X}} \textbf{W} = \textbf{U} \Sigma \textbf{V}^{\top},
\end{equation}

\noindent with this weight matrix constructed with sample-wise weights chosen, leveraging invariant property~\ref{li:output_invariant} of the output data $\tra{Y}$, to be biased toward lower output values. These weights are determined by subtracting each element of $\tra{Y}$ from $1$ and aggregating into a diagonal matrix

\begin{equation}
    \textbf{W} = \diag (1 - \tra{y_{1}}, 1 - \tra{y_{2}}, \ldots, 1 - \tra{y_{n}}).
\end{equation}

\noindent Note that, since $\cen{\textbf{X}}$ and $\textbf{W}$ are real matrices, the rotation matrix $\textbf{U}$ obtained through the SVD is orthogonal ($\textbf{U}^{-1} = \textbf{U}^{\top}$). Using this fact, multiplying the inverse rotation matrix with $\cen{\textbf{X}}$ aligns the weighted principal components of the product with the coordinate axes, since $\textbf{U}^{\top}\textbf{U}\Sigma \textbf{V}^{\top} = \textbf{I} \Sigma \textbf{V}^{\top}$, given by

\begin{equation}
    \tra{\textbf{X}} = \textbf{U}^{\top} \cen{\textbf{X}}
\end{equation}

\noindent and the rotation matrix $\textbf{R}$ in (\ref{eq:rotate_transform_matrices}) is set to

\begin{equation}\label{eq:u_update}
    \textbf{R} = \textbf{U}.
\end{equation}

\noindent \revision{Constructing a trust region in the transformed space of $\tra{X}$ now effectively induces a rotated trust region in the original space of $X$. Combined with the rescaling performed in the previous section,} this alignment with the coordinate axes of $\tra{X}$ with the weighted principal axes of $X$ assists in uncovering local separability that can be well-modelled by the \revision{length-scales of the} ARD kernel.

% %**********************************************
% \subsection{Illustrative trust region rotation example}\label{subsec:illustr_ex}
% %**********************************************

\todo[inline]{Illustrative example section moved earlier from after detailed algorithm description.}

To illustrate \revision{the advantage of this rotation strategy} compared to other \revision{\TRBO{}} methods, consider the optimi\sz{}ation of the Rosenbrock function~\cite{Rosenbrock}, a well-known test function with a narrow, banana-shaped valley leading towar\ds{} the global optimum. The starting point for the optimi\sz{}ation algorithm is chosen at the end of this valley, typically a very challenging starting point for most optimi\sz{}ation algorithms.\@ \revision{A typical run of this theoretical optimi\sz{}ation scenario is presented in Fig.~\ref{fig:illustrative_example}.}

\newcommand{\subfigSizeEx}{0.35}
\begin{figure}[h]
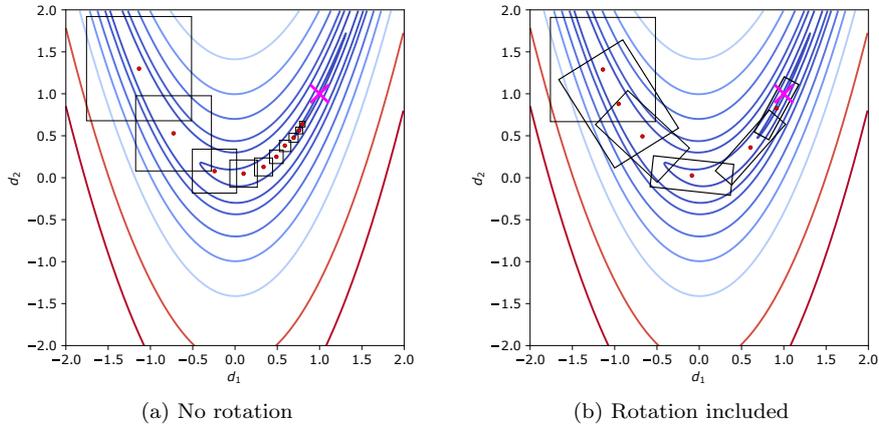

    \centering
    \subfloat[\label{subfig:illus_no_pca}No rotation]{\includesvg[inkscapelatex=false, width=\subfigSizeEx\linewidth]{figures/Section3/fig/rosen_contour_no_pca.svg}}\quad
    \subfloat[\label{subfig:illus_pca}Rotation included]{\includesvg[inkscapelatex=false, width=\subfigSizeEx\linewidth]{figures/Section3/fig/rosen_contour_pca.svg}}
    \caption{Illustrative example \revision{showing a typical run of a hypothetical \TRBO{}} algorithm applied to the 2-D Rosenbrock function \hyperref[subfig:illus_no_pca]{(a)}~without and \hyperref[subfig:illus_pca]{(b)}~with weighted principal component trust region rotation. A subset of trust regions (indicated in black) \centered{} on the respective minimum candidate solutions (indicated in red) are given, the global optimum is indicated by the magenta cross at $(1.0, 1.0)$ and observations other than the minimum candidate are not indicated to maintain visual clarity.}\label{fig:illustrative_example}
\end{figure}

Firstly, consider the \revision{hypothetical \TRBO{}} method described in Alg.~\ref{alg:BO_opt} without the weighted principal component rotation, given in Fig.~\ref{fig:illustrative_example}~\subref{subfig:illus_no_pca}. This algorithm behaves similarly to other trust region algorithms, with the trust region \revision{\recentered{} on new minimum candidates and moving along the valley.} However, the narrowness of the valley forces the trust region to shrink rapidly. Due to the constraint placed on the acquisition function that the objective function may only be evaluated at subsequent points \emph{inside} this trust region, this premature contraction slows progress towar\ds{} the optimum considerably.

Next, consider a \revision{\TRBO{} algorithm} with the weighted principal rotation. Inspecting Fig.~\ref{fig:illustrative_example}~\subref{subfig:illus_pca}, it is clear that this rotation yields a clear improvement. The rotation aligns the major axis of the rectangular trust region along the valley of the Rosenbrock function as the trust region moves through the valley. The trust region can now exploit the local separability of the valley by expanding and contracting along the major and minor axes of the trust region, \revision{which in the LABCAT algorithm is the directions of the length-scales in the ARD kernel from~(\ref{eq:sqexp_vector_form}).} This version of the algorithm finds the optimum \revision{more efficiently} than in Fig.~\ref{fig:illustrative_example}~\subref{subfig:illus_no_pca}, demonstrating the \revision{potential} value of the principal component rotation.

%**********************************************
\section{The \ALGPA{} algorithm}\label{sec:ALGPA_detail}
%**********************************************

\revision{This section presents the components of the proposed LABCAT algorithm, beginning with a combined observation transformation using the transforms defined in Sec.~\ref{sec:obs_rescale} and~\ref{sec:obs_rotate}. Furthermore, the iterative calculation of the parameters of this combined transformation and the estimation of the GP hyperparameters fitted to the transformed data are also given. Additionally, this section describes the specific implementation of the trust region, observation discarding strategy, algorithm initiali\sz{}ation and termination used in the LABCAT algorithm. The section concludes with a detailed mathematical description of the algorithm, synthesi\sz{}ing the work previously discussed.}

%**********************************************
\subsection{\revision{Combined observation transformation}}\label{subsec:combined_transform}
%**********************************************

\revision{Using the length-scale-based observation rescaling of Sec.~\ref{sec:obs_rescale} and weighted principal component trust region rotation of Sec.~\ref{sec:obs_rotate}, we define a combined transformation of the observations from $X$ to $\tra{X}$, leveraging the advantages of both.}

\revision{With the transformations defined in (\ref{eq:rescale_transform_matrices}) and (\ref{eq:rotate_transform_matrices}) and invariant properties~\ref{li:rescale_invariant_1}--\ref{li:rotate_invariant}, we construct the combined, full-rank affine transformation}

\begin{equation}\label{eq:input_transform}
    \textbf{x}_{i} = \textbf{R} \textbf{S} \tra{\textbf{x}_{i}} + \revision{\textbf{c}} \,\,\, \forall i \in \{ 1, 2, \ldots, n \},
\end{equation}

\noindent which also extends to

\begin{equation}\label{eq:input_transform_matrices}
    \textbf{X} = \textbf{R} \textbf{S} \tra{\textbf{X}} + \revision{\textbf{c}} \textbf{1}^{\top}_{n},
\end{equation}

\noindent with the transformation parameters $\textbf{R}$, $\textbf{S}$ and
$\revision{\textbf{c}}$ calculated as defined in the respective sections. The output data is also transformed according to (\ref{eq:output_transform}).

\revision{The process of enforcing the invariant properties~\ref{li:rescale_invariant_1}--\ref{li:rotate_invariant} at each algorithm iteration is the same as in the respective sections. Note that the specific order of enforcing the invariant properties is given in Fig.~\ref{fig:algpa_flowchart}, with the output data first being normali\sz{}ed, input data \recentered{}, rotated and finally rescaled according to the local length-scales.}

During the execution of the \ALGPA{} algorithm, it is not necessary to perform \revision{a} full \revision{recalculation} of $\tra{X}$ and $\tra{Y}$ from $X$ and $Y$ (with the respective transformation parameters defined in~(\ref{eq:input_transform_matrices}) and~(\ref{eq:output_transform})) at each algorithm iteration. Instead, \revision{to calculate the values for the current iteration, indicated by the subscript $\curIterIndicator{}$,} we leverage the transformed representations obtained from the preceding algorithm iteration, denoted as \revision{$\tra{\prevIter{X}}$ and $\tra{\prevIter{Y}}$.} Moreover, we retain the transformation parameters \revision{of the previous algorithm iteration}, also indicated by the subscript $\prevIterIndicator{}$. These transformed representations may incorporate added or removed observations such that the invariant properties~\ref{li:rescale_invariant_1}--\ref{li:rotate_invariant} no longer hold. To efficiently restore the invariant properties for these values from the preceding algorithm iteration, we calculate \revision{$\tra{\curIter{X}}$} and \revision{$\tra{\curIter{Y}}$} with the associated transformation parameters in terms of the values from the previous algorithm iteration. Note that the initial values and transformation parameters for the \ALGPA{} algorithm uses a modified version of this transform based on the bounds of the objective function $\Omega$ and is given in Sec.~\ref{subsec:alg_init_term}.

Firstly, the transformed outputs of the previous iteration \revision{$\tra{\prevIter{Y}}$} are renormali\sz{}ed using the minimum- and maximum values of this set, \revision{$\tra{\prevIterExtra{y}{\min}}$ and $\tra{\prevIterExtra{y}{\max}}$,} as

\revision{\begin{equation}\label{eq:output_min_max}
    \tra{\curIter{Y}} = T_{\minmax}(\tra{\prevIter{Y}}) = \biggl \{ \frac{\tra{\prevIterExtra{y}{i}} - \tra{\prevIterExtra{y}{\min}}}{\tra{\prevIterExtra{y}{\max}} - \tra{\prevIterExtra{y}{\min}}} \,\big|\, \prevIterExtra{y}{i} \in \prevIter{Y} \biggr \}
\end{equation}}

\noindent and the transformation parameters in the output transformation from~(\ref{eq:output_transform}) are calculated as

\revision{\begin{align}
    \revision{\curIter{b}} & = \prevIter{b} + \tra{\prevIterExtra{y}{\min}} \cdot \prevIter{a},                \\
    \curIter{a} & = \prevIter{a} \cdot (\tra{\prevIterExtra{y}{\max}} - \tra{\prevIterExtra{y}{\min}}). \nonumber
\end{align}}

Invariant property~\ref{li:rescale_invariant_1} is preserved by \recentering{} \revision{$\tra{\prevIter{X}}$} on the (possibly new) minimum candidate \revision{$\tra{\prevIterExtra{\textbf{x}}{\min}}$} with

\revision{\begin{equation}\label{eq:input_cen}
     \tra{\curIterExtra{\textbf{X}}{\textrm{cen}}} = \cen{T} (\tra{\prevIter{X}}) = \tra{\prevIter{\textbf{X}}} - \tra{\prevIterExtra{\textbf{x}}{\min}} \textbf{1}^{\top}_{n},
\end{equation}}

\noindent with the offset vector \revision{$\curIter{\textbf{c}}$} updated using the value of the previous iteration and the (possibly new) minimum candidate

\revision{\begin{equation}
    \curIter{\textbf{c}} = \prevIter{\textbf{c}} + \prevIter{\textbf{R}} \prevIter{\textbf{S}} \tra{\prevIterExtra{\textbf{x}}{\min}}.
\end{equation}}

Next, to restore invariant property~\ref{li:rotate_invariant} we inspect the transform defined in (\ref{eq:input_transform_matrices}). We can see that multiplying the transformed input data $\tra{\textbf{X}}$ by the scaling matrix $\textbf{S}$ yields a representation of the data in an intermediate, affine space of the original input space of $X$. After calculating the weighted principal components $\textbf{U}_{\textrm{a}}$ in this intermediate space

\revision{\begin{equation}\label{eq:svd_mod}
    \revision{\prevIter{\textbf{S}^{}}} \tra{\curIterExtra{\textbf{X}}{\textrm{cen}}} \textbf{W} = \textbf{U}^{}_{\textrm{a}} \Sigma^{}_{\textrm{a}} \textbf{V}^{\top}_{\textrm{a}}
\end{equation}}

\noindent we can use this rotation matrix \revision{$\textbf{U}_{\textrm{a}}$,} after a change of basis using \revision{$\prevIter{\textbf{S}}$} from the intermediate space, to rotate \revision{$\tra{\curIterExtra{\textbf{X}}{\textrm{cen}}}$} using the transform

\revision{\begin{equation}\label{eq:input_rot}
    \tra{\curIterExtra{\textbf{X}}{\textrm{rot}}} = \rot{T} (\tra{\curIterExtra{\textbf{X}}{\textrm{cen}}}) = \prevIter{\textbf{S}^{-1}} \textbf{U}^{\top}_{\textrm{a}} \prevIter{\textbf{S}^{}} \tra{\curIterExtra{\textbf{X}}{\textrm{cen}}}
\end{equation}}

\noindent and calculate the transformation parameter \revision{$\curIter{\textbf{R}}$} in terms of the value from the previous iteration

\revision{\begin{equation}
    \curIter{\textbf{R}} = \prevIter{\textbf{R}} \textbf{U}_{\textrm{a}},
\end{equation}}

\noindent essentially updating the rotational component of the transformation defined in~(\ref{eq:input_transform}) without requiring the full transformation of $\tra{X}$ back to $X$.

Finally, similarly to (\ref{eq:ell_diag_inv_mult}), we use the most likely length-scales for a GP fitted to \revision{$\tra{\curIterExtra{\textbf{X}}{\textrm{rot}}}$} and \revision{$\tra{\curIter{Y}}$ (collected into diagonal matrix $\curIter{\textbf{L}}$)} to rescale \revision{$\tra{\curIterExtra{\textbf{X}}{\textrm{rot}}}$} according to

\revision{\begin{equation}\label{eq:input_resc}
    \tra{\curIter{X}} = T_{\textrm{resc}} (\tra{\curIterExtra{\textbf{X}}{\textrm{rot}}}) = \curIter{\textbf{L}}^{-1} \tra{\curIterExtra{\textbf{X}}{\textrm{rot}}}
\end{equation}}

\noindent and recalculate the transformation parameter \revision{$\curIter{\textbf{S}}$} in terms of the value from the previous iteration

\revision{\begin{align}
    \curIter{\textbf{S}} & = \curIter{\textbf{L}} \prevIter{\textbf{S}},
\end{align}}

\noindent also essentially updating the scaling component of the transformation defined in (\ref{eq:input_transform}) without requiring the full transformation of $\tra{X}$ back to $X$.

Proofs for the validity of these updates are given in~\ref{subsec-app:lin_alg_offset_proof} to~\ref{subsec-app:lin_alg_resc_proof}. This cyclical process of \recentering{}, rescaling and rotation ensures that the transformed $\tra{X}$ and $\tra{Y}$ remain well-conditioned, even if the corresponding observations in the objective space $X$ and $Y$ become ill-conditioned or clustered closely together. The rotation performed in (\ref{eq:input_rot}) also ensures that the ARD kernel, as defined in (\ref{eq:sqexp_vector_form}), can effectively leverage local separability within the objective function, aligning the axes of each ARD length-scale with the axes of local separability.

%**********************************************
\subsection{GP hyperparameter estimation}\label{subsec:hyp_est}
%**********************************************

In the previous section describing the transformation from the original observations $X$ and $Y$ to transformed representations $\tra{X}$ and $\tra{Y}$, the length-scale invariant property described in~\ref{li:rescale_invariant_2} \revision{is preserved in (\ref{eq:input_resc})} by calculating the most likely length-scales $\boldsymbol{\ell}^{*}$ for a GP fitted to the \recentered{}, rotated \revision{observed inputs} $\tra{\rot{X}}$ from (\ref{eq:input_cen}) and normali\sz{}ed output values $\tra{Y}$ from (\ref{eq:output_min_max}). Instead of the conventional approach of a full re-estimation of the kernel hyperparameters $\boldsymbol{\theta}^{*}$ using the maximi\sz{}ation of the MAP estimate \revision{from} (\ref{eq:argmax_thetas}) at each algorithm iteration or once \revision{per a fixed number of algorithm} iterations, we adopt an approximative scheme. In this approach, the approximate hyperparameters for the local GP model are calculated at each algorithm iteration using a \revision{fixed, small} number of optimi\sz{}ation steps, tending towar\ds{} the exact hyperparameters with subsequent \emph{algorithm iterations}, not with additional \emph{optimi\sz{}ation steps} per algorithm iteration. Using a fixed number of optimi\sz{}ation steps per algorithm iteration, instead of executing as many optimi\sz{}ation steps as necessary for convergence of the hyperparameters, results in computational advantages by reducing the number of operations with a complexity of $O(n^{3})$ (recalculating the $\textbf{K}^{-1}$ matrix from Sec.~\ref{subsec:gp}) performed during each algorithm iteration.

We set the mean function $m(\cdot)$ of the GP to the mean of the transformed output data $\tra{Y}$ and choose to set the noise variance to $\sigma_{n} = 10^{-6}$ in our chosen kernel function from (\ref{eq:sqexp_vector_form}) to function as a small ``nugget'' term for increased numerical stability~\cite{nugget}. This prevents the kernel matrix $\textbf{K}$ in (\ref{eq:K}) from becoming singular if the observed points become very correlated by slightly inflating the uncertainty of the observed output values, in effect, adding a small offset to the diagonal entries of the kernel matrix. While the standard approach is to optimi\sz{}e $\sigma_{f}$ directly, we set this parameter to a fixed value of the standard deviation of $\tra{Y}$ \revision{(similar to the choice of prior in the BADS~\cite{BADS} algorithm)} at each algorithm iteration. Due to the continuous rescaling of the outputs described in (\ref{eq:output_min_max}), the algorithm is not very sensitive to this choice.

With these design choices, the hyperparameters to be optimi\sz{}ed are reduced to only the length-scales of the kernel $\boldsymbol{\ell}$ for the GP fitted in the transformed space of $\tra{\rot{X}}$ and $\tra{Y}$. The derivatives of the log-likelihood surface from (\ref{eq:gp_log_lik}) with respect to the length-scales $\boldsymbol{\ell}$~\cite{LogLikHess}, with the transformation such that the derivatives are with respect to the length-scales in logarithmic space to ensure that parameters are strictly positive, are given by the Jacobian defined as

%\begin{gather} %DOUBLE COL
%    \nabla\log p(\tra{Y} \,|\, \tra{\rot{X}}, \boldsymbol{\theta}) = \textbf{J} := \nonumber\\
%    \partialDeriv{\log p(\tra{Y} \,|\, \tra{\rot{X}}, \boldsymbol{\theta})}{\ln\ell_{i}} =
%    \frac{1}{2}\tra{\textbf{y}}^{\top} \textbf{K}^{-1} \partialDeriv{\textbf{K}}{\ln\ell_{i}} \textbf{K}^{-1} \tra{\textbf{y}} \nonumber \\
%    -\frac{1}{2}\tr \biggl( \textbf{K}^{-1} \partialDeriv{\textbf{K}}{\ln\ell_{i}} \biggr) \,\,\, \forall i \in \{1, 2, ..., d\} \label{eq:log_lik_Jac} 
%\end{gather}

\begin{equation} %SINGLE COL
    \nabla\log p(\tra{Y} \,|\, \tra{\rot{X}}, \boldsymbol{\theta}) = \textbf{J} :=
    {\biggl[ \partialDeriv{\log p(\tra{Y} \,|\, \tra{\rot{X}}, \boldsymbol{\theta})}{\ln\ell_{i}} \biggr]}_{1 \leq i \leq d} \in \mathbb{R}^{d \times 1} \label{eq:log_lik_Jac}
\end{equation}

\noindent and Hessian

%\begin{gather} %DOUBLE COL
%    \nonumber \nabla^{2}\log p(\tra{Y} \,|\, \tra{\rot{X}}, \boldsymbol{\theta}) = \textbf{H} :=  \\
%    \partialSecDerivDual{\log p(\tra{Y} \,|\, \tra{\rot{X}}, \boldsymbol{\theta})}{\ln\ell_{i}}{\ln\ell_{j}} =
%    \frac{1}{2}\tr \biggl( \textbf{K}^{-1} \partialSecDerivDual{\textbf{K}}{\ln\ell_{i}}{\ln\ell_{j}} \biggr) \nonumber\\ 
%    \nonumber -\frac{1}{2}\tr \biggl( \textbf{K}^{-1} \partialDeriv{\textbf{K}}{\ln\ell_{j}} \textbf{K}^{-1} \partialDeriv{\textbf{K}}{\ln\ell_{i}} \biggr) \\
%    + \tra{\textbf{y}}^{\top} \textbf{K}^{-1} \partialDeriv{\textbf{K}}{\ln\ell_{j}} \textbf{K}^{-1}
%    \partialDeriv{\textbf{K}}{\ln\ell_{i}} \textbf{K}^{-1} \tra{\textbf{y}}  \nonumber \\
%    - \frac{1}{2}\tra{\textbf{y}}^{\top} \textbf{K}^{-1} \partialSecDerivDual{\textbf{K}}{\ln\ell_{i}}{\ln\ell_{j}} \textbf{K}^{-1} \tra{\textbf{y}} \,\,\, \forall i, j \in \{1, 2, ..., d\} \label{eq:log_lik_Hess}
%\end{gather}

\begin{equation} %SINGLE COL
    \nabla^{2}\log p(\tra{Y} \,|\, \tra{\rot{X}}, \boldsymbol{\theta}) = \textbf{H} :=
    {\biggl[ \partialSecDerivDual{\log p(\tra{Y} \,|\, \tra{\rot{X}}, \boldsymbol{\theta})}{\ln\ell_{i}}{\ln\ell_{j}} \biggr]}_{1 \leq i,j \leq d} \in \mathbb{R}^{d \times d}\label{eq:log_lik_Hess}
\end{equation}

\noindent where the partial derivatives $\partialDeriv{\log p(\tra{Y} \,|\, \tra{\rot{X}}, \boldsymbol{\theta})}{\ln\ell_{i}}$ and $\partialSecDerivDual{\log p(\tra{Y} \,|\, \tra{\rot{X}}, \boldsymbol{\theta})}{\ln\ell_{i}}{\ln\ell_{j}}$ for the SE kernel with ARD from (\ref{eq:sqexp_vector_form}) are given in~\ref{subsec-app:kernel_derivs}.

\revision{It is reasonable to assume,} with respect to BO with a local trust region and the rescaling performed in the previous section, that the most likely length-scales for a GP fitted to a trust region typically exhibit mostly gradual changes as the trust region shifts. Thus, if the most likely length-scales are calculated for a GP fitted to a locally constrained window and a new potential minimum is determined, causing a slight shift in the window's location, the hyperparameters for the new window are expected to be similar to the previous values. To incorporate this assumption, similarly to the technique in~\cite{log_normal_hyperparam_prior}, we place a Gaussian prior \centered{} on $1$ ($0$ in $\log$-space) over the length-scales. This augmentation effectively restrains the GP from making abrupt changes in hyperparameters, ensuring the stability of the algorithm. For the standard deviation of this prior $\sigma_{\textrm{prior}}$ \revision{we suggest a default value} of approximately $0.1$, such that, by the three-sigma rule of thumb, the side lengths of the local trust region is unlikely to change by more than $30\%$ per algorithm iteration. Consequently, the log-likelihood formula from (\ref{eq:gp_log_lik}) is augmented with this new term (ignoring normali\sz{}ation constants) and is given by

\begin{equation} \label{eq:length_scale_prior}
    \log p(\tra{Y} | \tra{\rot{X}}, \boldsymbol{\theta}, \sigma_{\textrm{prior}}) = \log p(\tra{Y} | \tra{\rot{X}}, \boldsymbol{\theta}) - \sum^{d}_{i=1} \frac{\ln \ell^{2}_{i}}{2 \sigma^{2}_{\textrm{prior}}}, %- \frac{(\ln\sigma_{f} - \ln\sigma_{f \textrm{old}})^2}{2 \sigma^{2}_{\textrm{prior}}},
\end{equation}

\noindent with the Jacobian defined in (\ref{eq:log_lik_Jac}) augmented as

\begin{equation}
    \nabla\log p(\tra{Y} | \tra{\rot{X}}, \boldsymbol{\theta}, \sigma_{\textrm{prior}}) = \textbf{J} - \frac{1}{\sigma^{2}_{\textrm{prior}}}\ln \boldsymbol{\ell}
\end{equation}

\noindent and the Hessian defined in (\ref{eq:log_lik_Hess}) augmented as

\begin{equation}\label{eq:log_lik_Hess_aug}
    \nabla^{2}\log p(\tra{Y} | \tra{\rot{X}}, \boldsymbol{\theta}, \sigma_{\textrm{prior}}) = \textbf{H} - \frac{1}{\sigma^{2}_{\textrm{prior}}} \textbf{I}.
\end{equation}

Upon the calculation of these Jacobian and Hessian matrices, if $\textbf{H}$ is negative definite (all of the eigenvalues of $\textbf{H}$ are negative) it can be concluded that the current hyperparameters are in a convex-down region of the log-likelihood space. Consequently, a second-order Newton step can be effectively employed. Conversely, if $\textbf{H}$ is not negative definite, a gradient ascent step is used. Both of these steps are combined with a backtracking line search~\cite{BacktrackingLinesearch} to determine the optimal step length.

%This fact is evident from the examination of the ablation study in Sec.~\ref{subsubsec:coco_ablation}, where using a single Newton step exhibits virtually identical performance metrics to using ten.

Considering the rescaling of the input data performed in (\ref{eq:input_resc}), the calculation of the most likely length-scales during each algorithm iteration can be interpreted as an indication to expand or contract each of the dimensions of the transformed input data $\tra{X}$. If $\tra{X}$ is never \recentered{}, additional observations allow the GP to construct a more accurate model of the objective function and the length-scales of this better GP model will tend to unity with subsequent observed points.

%\begin{equation}\label{eq:log_lik_2nd_newton_step}
%    \boldsymbol{\theta}_{k+1} = \boldsymbol{\theta}_{k} - \gamma\textbf{H}^{-1}\textbf{J}
%\end{equation}

%with $\gamma$ selected with a backtracking line search~\cite{BacktrackingLinesearch}. 
%This Newton step has a guarantee of quadratic convergence speed. 

%\begin{equation}\label{eq:log_lik_2nd_newton_step}
%    \boldsymbol{\theta}_{k+1} = \boldsymbol{\theta}_{k} - \gamma\textbf{H}^{-1}\textbf{J}
%\end{equation}

%with $\gamma$ selected with a backtracking line search~\cite{BacktrackingLinesearch}. 
%This Newton step has a guarantee of quadratic convergence speed. 

%If it is not satisfied (i.e. the hyperparameters are not in a convex-down region of the log-likelihood) a simple gradient ascent step is used (with $\gamma$ also selected with backtracking line search):

%\begin{equation}\label{eq:log_lik_grad_step}
%    \boldsymbol{\theta}_{k+1} = \boldsymbol{\theta}_{k} + \gamma\textbf{J}.
%\end{equation}

%After new hyperparameters have been determined, the centered, rotated input data $\tra{\rot{\textbf{X}}}$ from (Sec.~\ref{subsec:internal_repr}) of the observed points is rescaled according to the new length-scales in the set $\boldsymbol{\ell}$ using (\ref{eq:input_resc}) after which $\boldsymbol{\ell}$ is homogenised and the GP model is refitted. This rescaling of the observed data according to the new length-scales forms the mechanism that allows the search domain defined in the next section to expand and contract based on the local objective function landscape.

%**********************************************
\subsection{Trust region definition and observation discarding}\label{subsec:search_dom}
%**********************************************

A key mechanism of \revision{\TRBO{}} is limiting the region of the acquisition function around the best candidate solution in which the next point is observed by means of a trust region. In the \ALGPA{} algorithm, after the set of observations $X$ is transformed to $\tra{X}$ according to the invariant properties~\ref{li:rescale_invariant_1} and~\ref{li:rescale_invariant_2}, that is, transformed such that the most likely length-scales of the kernel are unity ($\boldsymbol{\ell}^{*} = \{\ell_{i} = 1 \,|\,i\in 1, 2, \ldots, n\}$) and that the observed inputs are \centered{} on the current minimum candidate ($\tra{\textbf{x}_{\min}} = \begin{bmatrix} 0 & \ldots & 0 \end{bmatrix}^{\top} \in \tra{X}$), a trust region $\Omega_{\textrm{TR}}$ is constructed in the space of $\tra{X}$ as a closed, compact $d$-cube with a side length of $2\beta$. Using the Cartesian product, this trust region is defined as

\begin{equation}\label{eq:trust_region_def}
    \Omega_{\textrm{TR}} = {[ -\beta, \beta ]}^{d},
\end{equation}

\noindent where $\beta$ is a tunable parameter that captures the trade-off between the exploration of the region surrounding and the exploitation of the current minimum value. Small values for $\beta$ strongly encourage local exploitation, but may lead to small step sizes. In the case where $\beta$ tends to infinity, the algorithm will search for the next point in an unconstrained manner and revert to the global optimi\sz{}ation of standard Bayesian \acr{o}ptimi\sz{}ation. For this parameter, we recommend values in the interval $0.1 \leq \beta \leq 1$ according to the rough heuristic $\beta \approx \frac{1}{d}$. In our experience, \revision{this} range of values \revision{provides} a good trade-off, preventing the trust region from growing infinitely and encouraging convergence to a local optimum.

It is important to note that there is slight a difference between the trust region used by \ALGPA{} and those used by other, more classical methods. Instead of the size of the trust region being directly modified inside the fixed space of the original observations $X$, in the \ALGPA{} algorithm the size of the trust region is specified by (\ref{eq:trust_region_def}) in the transformed space of observations $\tra{X}$ and it is this space that is scaled. In effect, this induces a trust region in the original objective function space of $X$ without requiring the transformation of $\tra{X}$ back to $X$.

The acquisition function, chosen as EI in this paper (Sec.~\ref{subsec:bo})\footnote{Note that many alternative acquisition functions have been proposed~\cite{UCB, KG, ES} and while we have chosen EI for simplicity, other acquisition functions could potentially be substituted for EI in our proposed algorithm.}, is maximi\sz{}ed using \revision{$10d$ random samples uniformly distributed across the trust region $\Omega_{\textrm{TR}}$, similarly to the technique used by the TRLBO algorithm~\cite{TRLBO} and with the similar justification that the expected optimi\sz{}ation error from using random search will become smaller and smaller as the trust region shrinks.} The results obtained from this maximi\sz{}ation are also validated against the original constraints of the objective function $\Omega$ using rejection sampling after transforming the points back into the objective function space using the inverse of the transform defined in (\ref{eq:input_transform}), in effect, maximi\sz{}ing over the intersection of $\Omega$ and $\Omega_{\textrm{TR}}$, or

\begin{equation}
    \revision{\tra{\textbf{x}_{\ei}}} = \operatorname*{argmax}_{\subalign{\tra{\testpnt{x}} &\in \,\Omega_{\textrm{TR}} \\ \testpnt{x} &\in \, \Omega }} \revision{\alpha_{\ei}} (\tra{\testpnt{x}} ;\,\mathcal{GP}).
\end{equation}

Apart from limiting the region in which the next observation should be chosen, the trust region $\Omega_{\textrm{TR}}$ is also used to determine which observed points from $\tra{X}$ and $\tra{Y}$ should be preserved at each algorithm iteration. We assume that observations outside this trust region no longer contribute significant information, therefore, we discard these observations. This keeps the number of observations in the model and, by extension, the computation time per algorithm iteration relatively constant across the runtime of the algorithm, alleviating the noted computational slowdown of standard BO (Sec.~\ref{subsec:gp})\footnote{It is also not necessary to recalculate the kernel matrix, and the inverse/Cholesky decomposition thereof, for the reduced set of observations. The corresponding rows and columns of the observations in $\tra{X_{\textrm{rem}}}$ can be removed from $\textbf{K}$ while efficient updates can be applied to $\textbf{K}^{-1}$ in terms of the Schur complement~\cite{SchurComplement} or rank-1 downdates of the Cholesky decomposition of $\textbf{K}$, both of the order of $O(n^{2})$.}.

A minimum number of observations are preserved at each algorithm iteration, even if some may fall outside the trust region, as there may be cases where discarding enough observations cause the observation set to become rank-deficient, occurrences that may lead the fitted GP to make \revision{explosive} changes to the length-scales. We denote this \revision{cache size factor} $\revision{\m{}}$, such that the minimum number of preserved observations is $\revision{\m{}}$ \revision{multiplied by} the objective function dimensionality $d$. The operation applied to $\tra{X}$, removing the observations in the set $\tra{X_{\textrm{rem}}}$ if the size of $\tra{X}$ is larger than $\revision{\m{}} d$, is defined as

\begin{gather}\label{eq:input_discard}
    T_{\textrm{discard}}(\tra{X}, \tra{Y}) = \begin{cases}(\tra{X} \revision{, } \tra{Y}) & | \tra{X} | \leq \revision{\m{}} d \\ (\tra{X} \setminus \tra{X_{\textrm{rem}}}, \,\, \tra{Y} \setminus \{ \tra{y_{i}} \,|\, \tra{\textbf{x}_{i}} \in \tra{X_{\textrm{rem}}} \}) & | \tra{X} | > \revision{\m{}} d
    \end{cases}
\end{gather}

\noindent with the input observations to be discarded chosen, prioriti\sz{}ing older observations, to be those that fall outside the current trust region ($\forall \tra{\textbf{x}} \in \tra{X_{\textrm{rem}}},\, \tra{\textbf{x}} \notin \Omega_{\textrm{TR}}$) until the size of $\tra{X}$ reaches the $\revision{\m{}}d$ threshold. Note that in this operation, the current minimum candidate solution $\tra{\minpnt{x}}$ is guaranteed to be preserved due to invariant property~\ref{li:rescale_invariant_1}, which guarantees that this candidate is moved to the origin, an element of $\Omega_{\textrm{TR}}$ by definition. The corresponding elements from $\tra{Y}$ are also removed to ensure that $\tra{X}$ and $\tra{Y}$ retain a one-to-one correspondence.

This cache size factor $\revision{\m{}}$ is a user-specified parameter and a poor choice thereof may lead to suboptimal performance characteristics. For instance, if $\revision{\m{}}$ is set too low, the model GP fitted to $\tra{X}$ and $\tra{Y}$ may have too few observations to model the objective function. Conversely, if $\revision{\m{}}$ is set too high, the algorithm may become sluggish as it struggles to discard old, non-informative observations quickly enough to keep up with the moving trust region. Bearing these remarks in mind, we recommend a value in the interval $5 \leq \revision{\m{}} \leq 10$.

%A more informed method of forgetting points would involve scoring each point by its influence over the GP model approximating the search space. One way to do this involves calculating the squared difference of the predicted means of the full model and the model with the point removed using Bayesian quadrature~\cite{LossyCompression}. During the development of the algorithm, however, this was found to not be necessary and similar performance was achieved using the naive method.

%**********************************************
%\subsection{Multistart} \label{subsec:conv}
%**********************************************

%alg in current form is a local optimizer
%need multistart
%from TuRBO
%proof from TRIKE

%**********************************************
\subsection{Algorithm initiali\sz{}ation and termination}\label{subsec:alg_init_term}
%**********************************************

During the initiali\sz{}ation of the \ALGPA{} algorithm, similarly to standard BO, a set of initial points are chosen to be evaluated before the main loop begins, known as the design of experiment (DoE). Using the provided input domain $\Omega$, the DoE for the initial GP surrogate model $\revision{X_{0}}$ is distributed according to a Latin hypercube design~\cite{LHS} with $2d+1$ points to ensure full rank. Latin hypercube sampling (LHS) was chosen for the initial points to avoid clustering of the initial points, which might occur with random sampling, and to capture as much projected variance along the objective function's coordinate axes as possible.

After observing these initial points $\revision{X_{0}}$ and $\revision{Y_{0}}$ from the objective function, the upper- and lower bounds placed on the objective function used to define $\Omega$ \revision{in (\ref{eq:black_box_problem})} are used to intiali\sz{}e the respective transformed representations $\revision{\tra{X}_{0}}$ and $\revision{\tra{Y}_{0}}$. To do this, we adapt the mechanisms used to enforce the invariant properties~\ref{li:rescale_invariant_1},~\ref{li:rescale_invariant_2} and~\ref{li:output_invariant}, with the modification that $\revision{\tra{X}_{0}}$ is \centered{} on the midpoint of the bounds, no rotation is performed, and the bounds are rescaled to lie on the hypercube ${[-1, 1]}^{d}$ (in other words, unit length from the origin in each dimension). This modification ensures that initial observations from the DoE are well-conditioned, but also that the invariant properties~\ref{li:rescale_invariant_1}--\ref{li:rotate_invariant} temporarily no longer hold until the first algorithm iteration completes. The modified transformation to \revision{calculate the initial} \revision{$\tra{X}_{0}$} and \revision{$\tra{Y}_{0}$} is given by

\begin{equation} \label{eq:input_resc_bounds}
    (\revision{\tra{X}_{0}},\,\revision{\tra{Y}_{0}}) = T_{\Omega}(\revision{X^{}_{0}}, \revision{Y^{}_{0}}, \Omega) = \biggl( \revision{\textbf{S}_{0}^{-1}} (\textbf{X} - \revision{\textbf{c}^{}_{0}}\textbf{1}^{\top}), \, \biggl\{ \frac{y_{i} - y_{\min}}{y_{\max} - y_{\min}} \,\Big|\, y_{i} \in Y \biggr\} \biggr)
\end{equation}

\noindent where the input transformation parameters $\textbf{S}$, $\textbf{R}$ and $\revision{\textbf{c}^{}_{0}}$ are initiali\sz{}ed using $\Omega$ according to

\begin{align}
    \revision{\textbf{S}_{0}^{-1}}               & = \diag{\biggl(\frac{| \Omega_{1}^{\max} - \Omega_{1}^{\min} |}{2}, \frac{| \Omega_{2}^{\max} - \Omega_{2}^{\min} |}{2}, \ldots, \frac{| \Omega_{d}^{\max} - \Omega_{d}^{\min} |}{2} \biggr)}^{-1}, \nonumber \\
    \revision{\textbf{R}_{0}} &= \textbf{I}, \nonumber \\
    \textrm{and} \quad \revision{\textbf{c}^{}_{0}} & = {\biggr[ \frac{\Omega_{1}^{\max} + \Omega_{1}^{\min}}{2} \quad \frac{\Omega_{2}^{\max} + \Omega_{2}^{\min}}{2} \quad \ldots \quad \frac{\Omega_{d}^{\max} + \Omega_{d}^{\min}}{2} \biggl]}^{\top} \label{eq:input_init}
\end{align}

%\noindent $\tra{Y}$ is also initiali\sz{}ed by rescaling the initial observed outputs $Y$ using the min-max normali\sz{}ation of (\ref{eq:output_min_max}) and setting the transformation parameters $k$ and $\revision{b}$ as

\noindent with the output transformation parameters initiali\sz{}ed as

\begin{equation} \label{eq:output_init}
    \revision{a_{0}} = y_{\max} - y_{\min} \quad \textrm{and} \quad \revision{b_{0}} = y_{\min}.
\end{equation}

As with standard BO, the \ALGPA{} algorithm has no specific convergence criterion (Sec.~\ref{subsec:bo}) and may terminate as specified by the user if either \begin{enumerate*}[label= (\roman*)]
    \item the current candidate minimum output $y_{\min}$ is less than some target value,
    \item if the range of output values in $Y$ (captured by the variable $a$) falls below some tolerance or
    \item the maximum objective function evaluation budget is reached
\end{enumerate*}.

%**********************************************
\subsection{\revision{Algorithm pseudocode and discussion}}\label{subsec:alg_detail}
%**********************************************

Synthesi\sz{}ing the detailed descriptions given in this section of the salient components of the \ALGPA{} algorithm, as seen in Fig.~\ref{fig:algpa_flowchart}, the \ALGPA{} algorithm is given in Alg.~\ref{alg:algpa}.

\renewcommand*\Call[2]{\textproc{#1} (#2)}
\begin{algorithm*}[h]
\caption{\ALGPA{}}\label{alg:algpa}
\begin{algorithmic}[1] %[1] adds line numbers
\Require{} Objective function $f$, Bounds $\Omega$, Trust region size factor $\beta$, Observation cache size factor \revision{$\m{}$}, Length-scale prior standard deviation $\sigma_{\textrm{prior}}$

\vphantom{text}
\State{} $\revision{X_{0}}$ $\gets$ \Call{LatinHypercubeSampling}{$\Omega$} \Comment{Select initial DoE with Latin hypercube over bounds.}
\State{} $\revision{Y_{0}}$ $\gets$ $\{ f(\textbf{x}) \,|\, \textbf{x} \in \revision{X_{0}} \}$ \Comment{Evaluate objective function at initial input points.}
\State{} ($\revision{\tra{X}_{0}},\,\revision{\tra{Y}_{0}}$) $\gets$ $T_{\Omega}(\revision{X_{0}}, \revision{Y_{0}}, \Omega)$ \Comment{Initiali\sz{}e transformed representation of observed data (see~(\ref{eq:input_resc_bounds})).}
\State{} $\Omega_{\textrm{TR}}$ $\gets$ ${[ -\beta, \beta ]}^{d}$ \Comment{Construct trust region $d$-cube (see Sec.~\ref{subsec:search_dom}).}
%\State Fit GP with $\tra{X}$ and $\tra{Y}$ \Comment{Sec.~\ref{subsec:gp}}
\While{\textbf{not} converged}
    \State{} $\revision{\curIter{\tra{Y}}}$ $\gets$ $T_{\minmax}(\revision{\curIter{\tra{Y}}})$ \Comment{Normali\sz{}e observed output values (see~(\ref{eq:output_min_max})).}
    \State{} $\revision{\tra{\curIterExtra{\textbf{X}}{\textrm{rot}}}}$ $\gets$ $\rot{T}(\cen{T} (\revision{\tra{\curIter{X}}}))$ \Comment{\parbox[t]{0.585\linewidth}{\titlecap{\centre{}} observed inputs using current minimum candidate and rotate with weighted principal components (see~(\ref{eq:input_cen},~\ref{eq:input_rot})).}}
    \State{} $\mathcal{GP}$ $\gets$ $\mathcal{GP}(m(\cdot), k_{\textrm{SE}}(\cdot, \cdot) ; \revision{\tra{\curIterExtra{\textbf{X}}{\textrm{rot}}}}, \revision{\tra{\curIter{Y}}})$ \Comment{\parbox[t]{0.34\linewidth}{Fit GP with an SE kernel with ARD to $\tra{\rot{X}}$ and $\tra{Y}$ (see Sec.~\ref{subsec:gp}).}}
    %\State $\tra{X}$ $\gets$ \Call{forgetObs}{$\tra{X}$, $\beta$} \Comment{Sec.~\ref{subsec:search_dom}}
    \State{} $\boldsymbol{\ell}^{*} \gets$ $\operatorname*{argmax}_{\boldsymbol{\ell}} (\log p(\revision{\tra{\curIter{Y}}} \,|\, \revision{\tra{\curIterExtra{\textbf{X}}{\textrm{rot}}}}, \boldsymbol{\ell}) -\log p(\boldsymbol{\ell} \,|\, \sigma_{\textrm{prior}}))$ \Comment{Find most likely length-scales (see Sec.~\ref{subsec:hyp_est}).}
    \State{} $\revision{\tra{\curIter{X}}}$ $\gets$ $T_{\textrm{resc}}(\revision{\tra{\curIterExtra{\textbf{X}}{\textrm{rot}}}})$ \Comment{Rescale obs.\ inputs with most likely length-scales (see~(\ref{eq:input_resc})).}
    \State{} $(\revision{\tra{\curIter{X}}}, \revision{\tra{\curIter{Y}}})$ $\gets$ $T_{\textrm{discard}}(\revision{\tra{\curIter{X}}}, \revision{\tra{\curIter{Y}}})$  \Comment{\parbox[t]{0.5\linewidth}{Discard observations if over $\revision{\m{}}d$ threshold (see~(\ref{eq:input_discard})).}}
    \State{} $\revision{\tra{\textbf{x}_{\ei}}}$ $\gets$ $\operatorname*{argmax}_{\subalign{\tra{\testpnt{x}} &\in \,\Omega_{\textrm{TR}} \\ \testpnt{x} &\in \, \Omega }}$\Call{$\revision{\alpha_{\ei}}$}{$\tra{\testpnt{x}} ;\,\mathcal{GP}$} \Comment{\parbox[t]{0.37\linewidth}{Maximi\sz{}e EI acquisition function over trust region and bounds (see~(\ref{eq:EI})).}}
    \State{} $\revision{y_{\ei}}$ $\gets$ $f(\revision{\textbf{x}_{\ei}})$ \Comment{\parbox[t]{0.43\linewidth}{Evaluate suggested input point from objective function, transformed according to (\ref{eq:input_transform}).}}
    \State{} $\revision{\tra{\nextIter{X}}}$ $\gets$ $\revision{\tra{\curIter{X}}}$ $\cup$ $ \{ \revision{\tra{\textbf{x}_{\ei}}} \}$ 
    \Comment{Append suggested input point.}
    \State{} $\revision{\tra{\nextIter{Y}}}$ $\gets$ $\revision{\tra{\curIter{Y}}}$ $\cup$ $ \{ \revision{\tra{y_{\ei}}} \}$ \Comment{\parbox[t]{0.6\linewidth}{Append evaluated output value, transformed according to (\ref{eq:output_transform}).}}
\EndWhile{}
\State{} \Return{} $\textbf{x}_{\min}$, $y_{\min}$ \Comment{Return current minimum candidate solution.}
\end{algorithmic}
\end{algorithm*}

The modified \revision{\TRBO{}} loop is clearly visible (lines $5-16$), with the additions of the transformation of the observed data (lines $3$, $6-7$ and $10$), determining the optimal length-scales (line $9$) \revision{and the} discarding of observations (line $11$).

\revision{Performing an analysis of the asymptotic computational complexity of the LABCAT algorithm, given in~\ref{subsec-app:comp_complex}, we can conclude that the complexity would be roughly linear in terms of the number of algorithm iterations. This is due to the dominant operation of each iteration, the Cholesky decomposition used to calculate $\textbf{K}^{-1}$ during the fitting of the GP, being $O (N n^{3})$ for $N$ iterations. However, the observation discarding strategy defined in Sec.~\ref{subsec:search_dom} creates an approximate upper limit of $\m{}d$ for the number of stored observations $n$ and, since $\m{}$ and $d$ are constant for a single instance of the LABCAT algorithm, we can conclude $O(N n^{3}) \approx O(N \m{}^{3}d^{3}) \approx O(N)$.}

We can now identify the mechanisms though which the objectives of this paper are addressed, these objectives being to develop a \revision{\TRBO{}} method that (i) is resistant to computational slowdown, (ii) is adaptable to non-stationary and ill-conditioned functions without kernel engineering, and (iii) exhibits good convergence characteristics. Firstly, using the greedy data discarding strategy defined in Sec.~\ref{subsec:search_dom}, the number of observations used to construct the local GP surrogate is kept relatively constant at each algorithm iteration. This avoids the computational slowdown of standard BO with more observations noted in Sec.~\ref{subsec:gp}. Secondly, the use of a local trust region based on the local length-scales of GP surrogate allows the algorithm to adapt to the local behavi\ou{}r of a non-stationary objective function. The rotation of the trust region using the weighted principal components also allows the trust region to adapt to ill-conditioning of the objective function in arbitrary directions. Lastly, the use of a transformed representation of the observed data $\tra{X}$ and $\tra{Y}$ that is forced to be well-conditioned allows the \ALGPA{} algorithm to converge much closer to a solution before encountering the numerical issues \revision{found in} standard BO\@.

%**********************************************
\section{Experimental Results}\label{sec:results}
%**********************************************

With the proposed \ALGPA{} algorithm, we present the results of a numerical performance analysis using computational experiments consisting of two sections. The first experiment compares the proposed \ALGPA{} algorithm with similar \revision{\TRBO{}} algorithms by applying them to several well-known, synthetic optimi\sz{}ation test functions. The second experiment applies these algorithms, as well as algorithms from the wider field of derivative-free optimi\sz{}ation, to the BBOB test suite using the COCO benchmarking software~\cite{COCO}. This extensive test suite is designed to be a representative sample of the more difficult problem distribution that can be expected in practical continuous-domain optimi\sz{}ation. An ablation study is also performed on the \ALGPA{} algorithm using this benchmark to determine the contribution of each element of the algorithm to overall performance. Using the results from these benchmarks, we comment on the relative performance of \ALGPA{} when compared to other algorithms and when applied to certain function groups with shared characteristics. All results in this report were obtained on \revision{an} 11th generation Intel i7--11700 CPU @ 2.5 GHz, with the exception of those extracted from the COCO archive.

%**********************************************
\subsection{Synthetic test function benchmarks}\label{subsec:synth_results}
%**********************************************

In this section, we perform a comparison of the proposed \ALGPA{} algorithm with standard BO and other \revision{\TRBO{}} algorithms on selected, well-known 2-D test objective functions: the sphere~\cite{Sphere}, \revision{quartic~\cite{Quartic}, Booth~\cite{Booth}}, Branin-Hoo~\cite{Branin}, Rosenbrock~\cite{Rosenbrock} and Levy~\cite{Levy} functions\footnote{The sphere, Rosenbrock and quartic functions are also known as the first, second and fourth De Jong functions~\cite{Sphere}.}. These synthetic functions are each designed with certain properties and have been selected to cover a large range of these problem characteristics.\@ \revision{The sphere, quartic, Booth and Rosenbrock functions are unimodal, with the Booth and Rosenbrock functions being badly conditioned and the optimum of the Rosenbrock function in a curved valley making convergence difficult.} The Branin-Hoo and Levy functions are multimodal, with the Branin-Hoo function having several global minima and the Levy function having several local minima. The results in this section are, therefore, included for a comparison of the convergence behavi\ou{}r between the proposed- and selected algorithms on a range of objective function characteristics.

\ALGPA{} has been implemented using Rust and is available in the \texttt{labcat} library\footnote{\revision{\url{https://github.com/esl-sun/LABCAT}}}. This library also provides a Python interface. For the other algorithms, we use the SRSM implementation from the Python \texttt{bayesian-optimization}\footnote{\url{https://github.com/bayesian-optimization/BayesianOptimization}} library as well as the Python interfaces provided by the authors of the TuRBO, BADS and TREGO algorithms from the \texttt{turbo}\footnote{\url{https://github.com/uber-research/TuRBO}}, \texttt{pybads}\footnote{\url{https://github.com/acerbilab/pybads}} and \texttt{trieste}\footnote{\url{https://github.com/secondmind-labs/trieste}} libraries, respectively. A purely random search as well as standard BO, also from the \texttt{bayesian-optimization} package, is also included to serve as a performance baseline.

We \revision{set the parameters} of the \ALGPA{} algorithm within the recommended ranges ($\beta = 0.5, m = 7$ and $\sigma_{\textrm{prior}} = 0.1$) with the recommended DoE budget ($2d + 1$) as described in Sec.~\ref{subsec:alg_init_term}. Each of the algorithms that we compared against is initiali\sz{}ed with default parameters and a DoE budget of $2d + 1$. The TuRBO algorithm is also used in \revision{the two configurations of the original paper:} a single trust region (``TuRBO-1'') and five parallel trust regions (``TuRBO-5''). The results of this comparison is given in Fig.~\ref{fig:synth_benches} and~\revision{\ref{subsec-app:synth_full}}.\@ \revision{Note that while the LABCAT algorithm exhibits the best wall-clock times (aside from the random algorithm), this should be interpreted as an indication of relative performance as these times do not account for differing implementations and levels of optimization.}

\renewcommand{\tempsize}{0.32}
\begin{figure*}[h]
    \centering
    \subfloat[Sphere]{\includegraphics[width=\tempsize\textwidth]{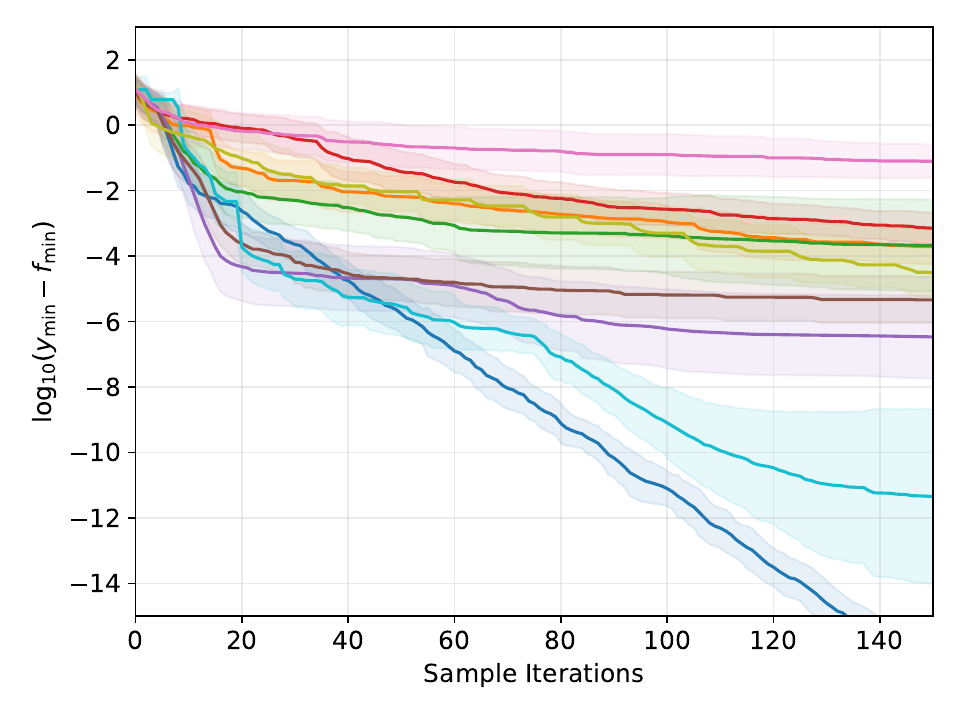}}
    \subfloat[Quartic]{\includegraphics[width=\tempsize\textwidth]{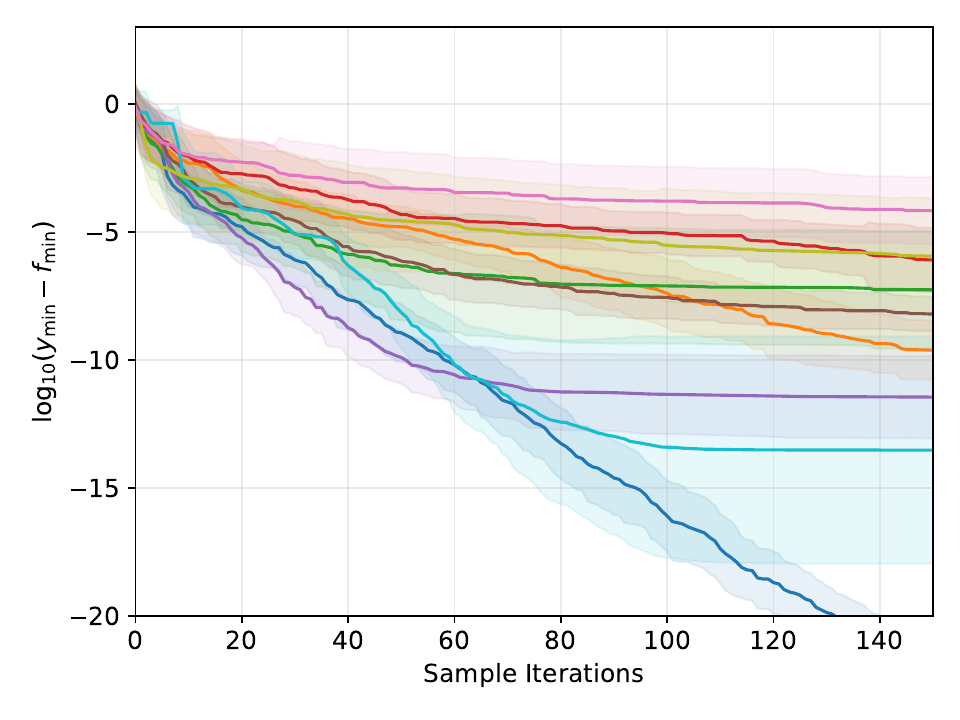}}
    \subfloat[Booth]{\includegraphics[width=\tempsize\textwidth]{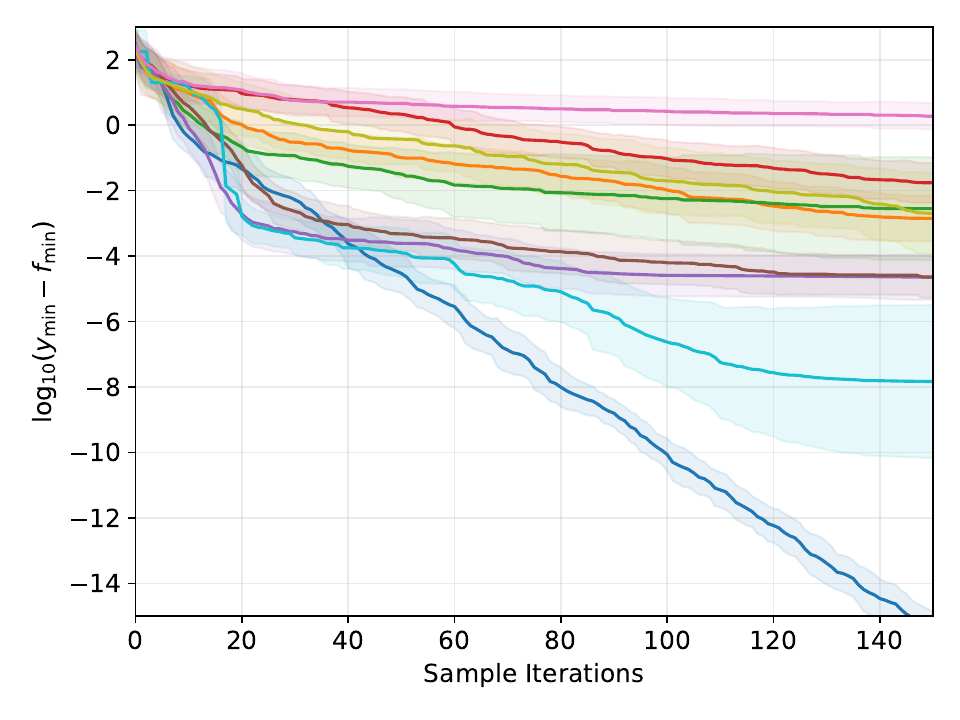}} \\
    \subfloat[Rosenbrock]{\includegraphics[width=\tempsize\textwidth]{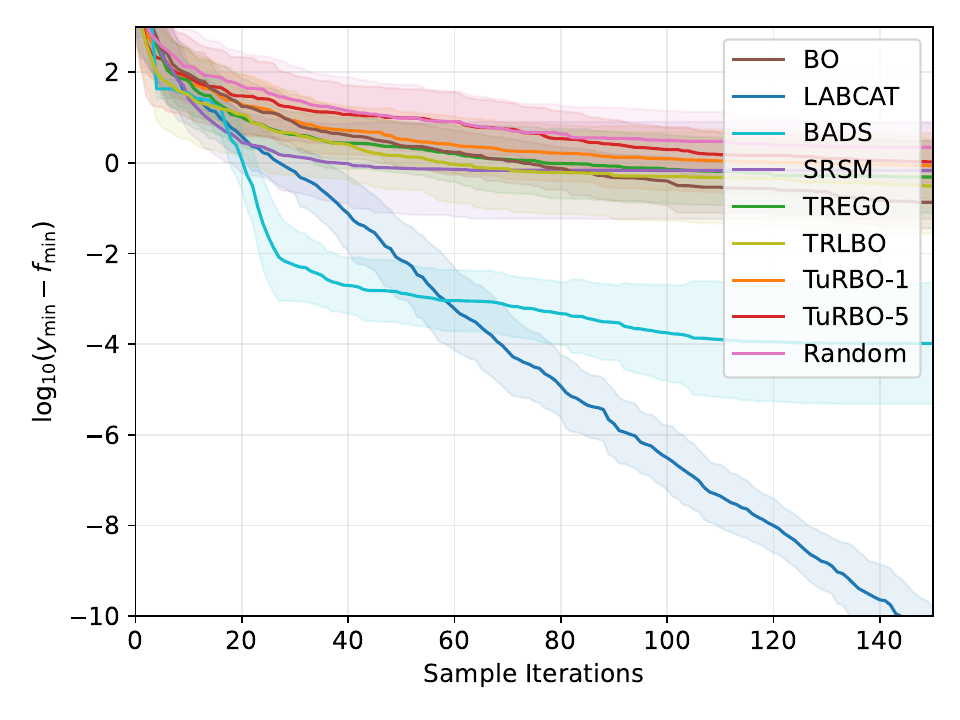}}
    \subfloat[Branin-Hoo]{\includegraphics[width=\tempsize\textwidth]{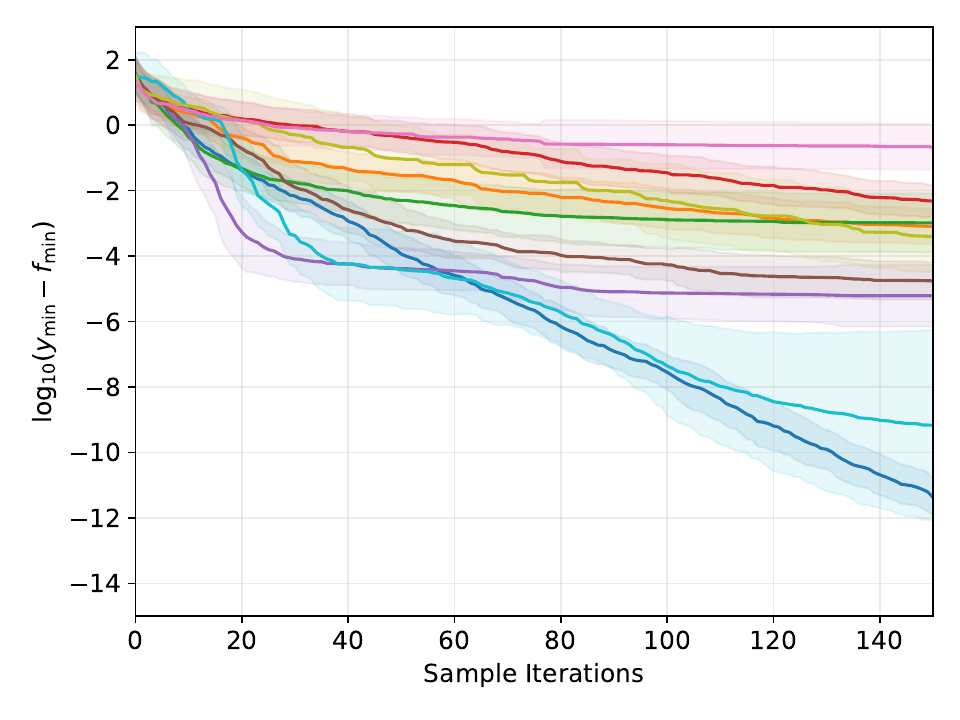}}
    \subfloat[Levy]{\includegraphics[width=\tempsize\textwidth]{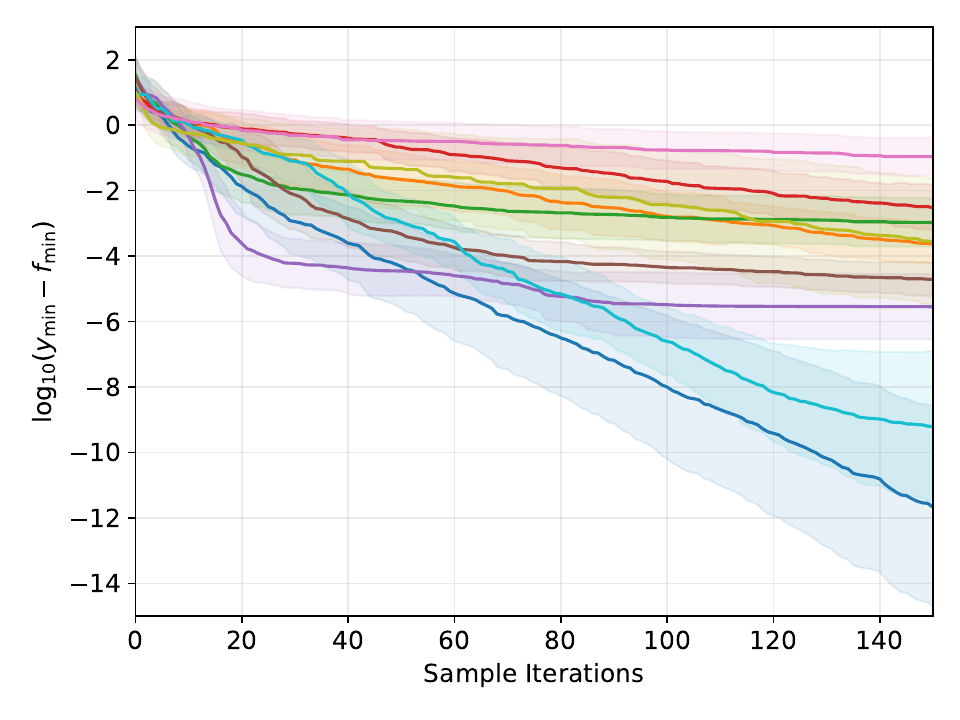}}
    \caption{Performance of selected algorithms applied to synthetic 2-D test functions. We conduct 50 independent \revision{optimi\sz{}ation} runs per algorithm with a sampling budget of 150 objective function evaluations. The mean and standard deviation, indicated by the shaded regions, of the logarithmic global regret, which is the log-difference between the best candidate solution $y_{\min}$ at each sampling iteration of the objective function and the global \revision{minimum $f_{\min}$,} are reported.\ \revision{The definitions and domains of each objective function is given in~\ref{subsec-app:synth_bench_funcs}.}}\label{fig:synth_benches}
\end{figure*}

The noted shortcoming of standard BO, namely that BO struggles to converge to an arbitrary precision, is clearly visible. This deficiency is also inherited by the SRSM, TuRBO \revision{and TRLBO} algorithms, clearly making very slow progress \revision{after} 150 objective function samples. The BADS algorithm exhibits better convergence characteristics for the Branin-Hoo and Levy functions, possibly due to the deterministic mesh adaptive direct search (MADS) fallback step incorporated in this algorithm. It is clear that the \ALGPA{} algorithm is not only capable of consistent convergence to a much higher level of precision for a wide range of objective function characteristics, but also may do so faster than comparable algorithms without needing to switch away from BO\@.

%**********************************************
\subsection{COCO black-box optimi\sz{}ation benchmark}\label{subsec:coco_results}
%**********************************************

For the second experiment, we base our analysis on the \acr{co}mparing \acr{c}ontinuous \acr{o}ptimi\sz{}ers (COCO) benchmarking software~\cite{COCO}. In this paper, we make use of the noiseless \acr{b}lack-\acr{b}ox \acr{o}ptimi\sz{}ation \acr{b}enchmarking (BBOB) test suite~\cite{COCO_2009_doc} that comprises 24 black-box objective functions to optimi\sz{}e, \revision{given in~\ref{subsec-app:bbob_2009}.} The functions in this test suite are collected into 5 groups, each with the following shared characteristics:
\begin{enumerate*}[label= (\roman*)]
    \item separable,\label{li:separable}
    \item unimodal with moderate conditioning,\label{li:unimod_low_cond}
    \item unimodal with high conditioning,\label{li:unimod_high_cond}
    \item multimodal with adequate global structure, and\label{li:multimod_adeq_struct}
    \item multimodal with weak global structure.\label{li:multimod_weak_struct}
\end{enumerate*}

To evaluate the performance of a single optimi\sz{}ation algorithm on this suite, 15 \emph{instances} of each function are generated, with each instance corresponding to a randomi\sz{}ed modification of the function by a random translation of the optimum and a random rotation of the coordinate system. For each instance, an array of \emph{problems} \revision{is} also generated. Each of these problems are defined as a tuple comprising a function instance and a \emph{target} precision to reach. To set these targets and give a good performance reference, COCO defines a composite algorithm known as best2009, an algorithm composed of the best performing optimi\sz{}ation algorithm for each function from the BBOB-2009 workshop~\cite{best_2009}\footnote{Note that, due to being composed of the best performing algorithms for each function, outperforming best2009 is quite a difficult task.}. Similar to the experimental setup of Diouane et al.~\cite{TREGO} \revision{and Bajer et al.~\cite{DTS-CMA-ES},} we set the targets for each instance as the values reached by best2009 after a certain number of objective function evaluations. Specifically, we set this number of function evaluations to a set of 50 values $[ 0.5, \ldots, 100] \times d$, uniformly distributed in log-space. For the analysis in this paper, all of the algorithms tested are provided with a total sampling budget of $200d$ objective function evaluations per function instance to reach these targets. If an algorithm terminates before exhausting the sampling budget, unless otherwise specified, an independent restart is performed with the remaining sampling budget. With the results of an optimi\sz{}ation algorithm applied to these problems, empirical cumulative distribution functions of runtimes (runtime ECDFs) are compiled, which gives the proportion of problems solved by the algorithm for a given budget of objective function evaluations. Note that, similarly to the benchmarks in the previous section, a purely random search is also included in the analysis to serve as lower bound for performance. The results in this paper from the COCO software represent several weeks of combined CPU time on the previously mentioned hardware used in this paper.

%**********************************************
\subsubsection{\ALGPA{} ablation study with the COCO benchmark}\label{subsubsec:coco_ablation}
%**********************************************

In this section, an ablation study is performed on the \ALGPA{} algorithm to assess the contribution and significance of each component of the \ALGPA{} algorithm on overall performance. The unablated algorithm is set \revision{as} the \ALGPA{} algorithm with a set of parameters within the recommended ranges ($\beta = \frac{1}{d}, \revision{\m{}} = 7$ and $\sigma_{\textrm{prior}} = 0.1$), denoted as ``\ALGPA{}'', and compared against instances of \ALGPA{} with \begin{enumerate*}[label= (\alph*)]
    \item no principal component rotation (``\ALGPA{} noPC''),
    \item more passive discarding of observations by doubling the maximum recommended value for $\revision{\m{}}$ to $20$ (``\ALGPA{} \revision{p20}''),
    \item a uniform length-scale prior distribution (``\ALGPA{} ULSP''), and
    \item an increased number of \revision{Newton} or gradient steps during hyperparameter optimi\sz{}ation
          (``\ALGPA{} n10'').
\end{enumerate*} The results obtained by applying each of the ablated versions of \ALGPA{} on the BBOB test suite is summari\sz{}ed in Fig.~\ref{tab:ertd_summary_param_sweep} and the full results are provided in~\ref{subsec-app:ertd_coco_pca_comp}.

\captionsetup[table]{name=Figure}
\newcommand{\subfigSize}{0.28}
\newcolumntype{M}[1]{>{\centering\arraybackslash}m{#1}}
\setcounter{table}{\thefigure}
\begin{table*}[!ht]
    \centering
    \begin{tabular}{cM{\subfigSize\textwidth}M{\subfigSize\textwidth}M{\subfigSize\textwidth}}
        \toprule
    & $d=2$ & $d=5$ &$d=10$ \\
        \midrule
        \rotatebox[origin=c]{90}{All Functions} & \includesvg[width=\subfigSize\textwidth]{figures/Section4/coco_param_sweep/pprldmany_02D_noiselessall.svg} & \includesvg[width=\subfigSize\textwidth]{figures/Section4/coco_param_sweep/pprldmany_05D_noiselessall.svg} & \includesvg[width=\subfigSize\textwidth]{figures/Section4/coco_param_sweep/pprldmany_10D_noiselessall.svg} \\
        \bottomrule
    \end{tabular}
    \caption{Selected empirical cumulative distribution functions (ECDFs) of runtimes table for ablation study with the COCO dataset over all functions in dimensions $2, 5$ and $10$. The values on the $y$-axis represent the proportion of runtime-based optimi\sz{}ation targets achieved for a certain number of objective function samples. Algorithms that achieve these targets faster have a larger area under the curve and are therefore considered to have superior performance.}\label{tab:ertd_summary_param_sweep}
\end{table*}

It is clear that the principal component rotation contributes significantly to the overall \ALGPA{} algorithm performance, especially in lower dimensions and for unimodal functions in groups~\ref{li:unimod_low_cond} and~\ref{li:unimod_high_cond}. This aligns with the behavi\ou{}r observed in Fig.~\ref{fig:illustrative_example}, as several of the functions in these groups are characteri\sz{}ed by valleys with changes in direction similar to the Rosenbrock function. A similar contribution can be seen by the removal of observations, with a more pronounced difference in $5$ and $10$ dimensions. Severe performance degradation is observed when the Gaussian prior placed over the kernel length-scales in (\ref{eq:length_scale_prior}) is removed. While not indicated by the COCO-generated runtime ECDF graphs, a significant increase in the number of restarts was observed, indicating instability in the ablated \ALGPA{} algorithm. As opposed to the previously mentioned modifications that may yield performance gains for certain function groups, removing the length-scale prior is a strict downgrade, with no performance gains in any function group. Additional or gradient steps during the optimi\sz{}ation of the length scales yield little to no significant performance increases, therefore a single or gradient step seems to be sufficient, with all of the associated computational savings.

Inspecting~\ref{subsec-app:ertd_coco_pca_comp}, it is interesting to note that removal of the principal component rotation and slower observation removal yields modest improvements when applied to the multimodal function groups~\ref{li:multimod_adeq_struct} and~\ref{li:multimod_weak_struct}. This may be due to the slower convergence of these modified versions of the \ALGPA{} algorithm leading to more exploration of the objective function space, finding slightly better solutions for these functions. In practice, if prior information regarding the objective function is available that indicates a multimodal structure and the additional computational cost can be spared, the observation cache multiplier $\revision{\m{}}$ could be increased for better performance.

In summary, each of the constituent components of \ALGPA{} contribute significantly to overall performance and the assumptions made in Sec.~\ref{subsec:hyp_est} and~\ref{subsec:search_dom} are shown to be well-founded.

%**********************************************
\subsubsection{Comparison with state-of-the-art optimi\sz{}ation algorithms using the COCO benchmark}\label{subsubsec:coco_full}
%**********************************************

To compare the proposed \ALGPA{} algorithm to similar algorithms from the wider field of derivative-free optimi\sz{}ation, we expand the set of algorithms included in the comparison from Sec.~\ref{subsec:synth_results} with the state-of-the-art DTS-CMA-ES, AutoSAEA, MCS, NEWUOA and SMAC algorithms. The DTS-CMA-ES~\cite{DTS-CMA-ES} \revision{and AutoSAEA~\cite{AutoSAEA} algorithms use a surrogate-assisted evolutionary strategy based on a combination of the evolutionary algorithm} and GP surrogates, known to be well-suited to multimodal problems. SMAC~\cite{SMBO-SMAC} is a variation of standard BO using an isotropic GP kernel and a locally biased stochastic search to optimi\sz{}e the EI\@. Multilevel coordinate search (MCS)~\cite{MCS} balances a global search based on the DIRECT~\cite{DIRECT} algorithm and a local search using a local quadratic interpolation. NEWUOA~\cite{NEWUOA} also uses a quadratic interpolation, but combines it with a classical trust region approach.

Results for DTS-CMA-ES, MCS, NEWUOA and SMAC were obtained from the COCO database (see the respective publications~\cite{DTS_CMA_ES_COCO, MCS_COCO, NEWUOA_COCO, SMAC_COCO}). The results obtained by applying each of the functions on the BBOB test suite are shown on Fig.~\ref{tab:ertd_summary}, with results in 2, 3 and 5 dimensions across all functions and from the selected function groups~\ref{li:unimod_high_cond} and~\ref{li:multimod_weak_struct}. Complete results that include the other function groups can be found in~\ref{subsec-app:ertd_coco_full}.

%\begin{figure}[H]
%\centering
%\subfloat[All functions]{\includesvg[width=0.31\textwidth]{figures/Section4/coco_results/pprldmany_02D_noiselessall.svg}}\quad
%\subfloat[Unimodal, High Cond.]{\includesvg[width=0.31\textwidth]{figures/Section4/coco_results/pprldmany_02D_hcond.svg}}\quad
%\subfloat[Multimodal, Weak Struct.]{\includesvg[width=0.31\textwidth]{figures/Section4/coco_results/pprldmany_02D_mult2.svg}}\quad
%\subfloat[All functions.]{\includesvg[width=0.31\textwidth]{figures/Section4/coco_results/pprldmany_03D_noiselessall.svg}}\quad
%\subfloat[Unimodal, High Cond.]{\includesvg[width=0.31\textwidth]{figures/Section4/coco_results/pprldmany_03D_hcond.svg}}\quad
%\subfloat[Multimodal, Weak Struct.]{\includesvg[width=0.31\textwidth]{figures/Section4/coco_results/pprldmany_03D_mult2.svg}}\qquad
%\subfloat[All functions]{\includesvg[width=0.3\textwidth]{figures/Section4/coco_results/pprldmany_05D_noiselessall.svg}}\quad
%\subfloat[Unimodal, High Cond.]{\includesvg[width=0.31\textwidth]{figures/Section4/coco_results/pprldmany_05D_hcond.svg}}\quad
%\subfloat[Multimodal, Weak Struct.]{\includesvg[width=0.31\textwidth]{figures/Section4/coco_results/pprldmany_05D_mult2.svg}}\quad
%\caption{Three sub-floats.}
%\label{fig:a}
%\end{figure}

\captionsetup[table]{name=Figure}
%\newcommand{\subfigSize}{0.28}
%\newcommand{\dummyfigure}{#1}{\includesvg[width=\subfigSize\textwidth]{#1}}
%\newcolumntype{M}[1]{>{\centering\arraybackslash}m{#1}}
%\setcounter{table}{\thefigure}
\begin{table*}[ht!]
    \centering
    \begin{tabular}{cM{\subfigSize\textwidth}M{\subfigSize\textwidth}M{\subfigSize\textwidth}}
        \toprule
        Dim. & All Functions &\ref{li:unimod_high_cond} Unimodal, High~Conditioning &\ref{li:multimod_weak_struct} Multimodal, Weak~Structure \\
        \midrule
        \rotatebox[origin=c]{90}{$d=2$}  & \includesvg[width=\subfigSize\textwidth]{figures/Section4/coco_results/pprldmany_02D_noiselessall.svg} & \includesvg[width=\subfigSize\textwidth]{figures/Section4/coco_results/pprldmany_02D_hcond.svg} & \includesvg[width=\subfigSize\textwidth]{figures/Section4/coco_results/pprldmany_02D_mult2.svg} \\
        \rotatebox[origin=c]{90}{$d=5$}  & \includesvg[width=\subfigSize\textwidth]{figures/Section4/coco_results/pprldmany_05D_noiselessall.svg} & \includesvg[width=\subfigSize\textwidth]{figures/Section4/coco_results/pprldmany_05D_hcond.svg} & \includesvg[width=\subfigSize\textwidth]{figures/Section4/coco_results/pprldmany_05D_mult2.svg} \\
        \rotatebox[origin=c]{90}{$d=10$} & \includesvg[width=\subfigSize\textwidth]{figures/Section4/coco_results/pprldmany_10D_noiselessall.svg} & \includesvg[width=\subfigSize\textwidth]{figures/Section4/coco_results/pprldmany_10D_hcond.svg} & \includesvg[width=\subfigSize\textwidth]{figures/Section4/coco_results/pprldmany_10D_mult2.svg} \\
        \bottomrule
    \end{tabular}
    \caption{Selected empirical cumulative distribution functions (ECDFs) of runtimes table from the COCO dataset for comparison of the \ALGPA{} algorithm with various state-of-the-art optimi\sz{}ation algorithms in dimensions $2, 5$ and $10$. The first column reports the results of these algorithms applied to all 24 functions of the BBOB-2009 test suite (\ref{subsec-app:bbob_2009}) for a given dimension while the second- and third columns report the results for the two most difficult function groups:~\ref{li:unimod_high_cond} unimodal functions with high conditioning \revision{($f_{10}$-$f_{14}$)} and~\ref{li:multimod_weak_struct} multimodal functions with weak global structure \revision{($f_{20}$-$f_{24}$)}.}\label{tab:ertd_summary}
\end{table*}

The results obtained from applying the selected algorithms to the COCO dataset reveal several important findings. Firstly, the \ALGPA{} algorithm emerges as the top performer when considering the aggregate of all tested functions, having, from inspection, the largest area under the curve for any single algorithm in all dimensions (excluding the composite best2009 algorithm). Additionally, the algorithm excels particularly well when applied to unimodal functions with high conditioning from group~\ref{li:unimod_high_cond}, even surpassing best2009 for a function group that is not traditionally considered to be well-suited for BO\@. Furthermore, the algorithm proves to be highly proficient for 2-D and 5-D, where it achieves the best performance of all of the BO-based algorithms, with a smaller performance gap in 10-D between the \ALGPA{} and BADS algorithms. Inspecting the results in~\ref{subsec-app:ertd_coco_full}, the performance of the \ALGPA{} algorithm is also slightly lower than those of other \revision{\TRBO{}} algorithms (BADS, SRSM, TuRBO, TREGO) when applied to function group~\ref{li:multimod_adeq_struct}, probably due to \ALGPA{} heavily fav\ou{}ring local exploration of the objective function and, by extension, being unable to model the underlying global structure of these functions as well as the other algorithms.

The closest competitor from the \revision{compared \TRBO{}} algorithms, when considering all functions, is the BADS algorithm. The BADS algorithm is, however, not very resistant to highly conditioned functions from group~\ref{li:unimod_high_cond}, being consistently outperformed in this function group by the \ALGPA{} algorithm. It is unclear how much of the performance of the BADS algorithm can be ascribed to its deterministic MADS fallback step, although some information may be gleaned by comparing this performance to another \revision{\TRBO{}} method without this fallback step, such as TuRBO\@. In this comparison, similar performance is observed for multimodal functions, while BADS performs better for separable and unimodal functions. This may imply that the MADS algorithm incorporated into BADS allows for increased local exploitation when compared to other, ``purer'' \revision{\TRBO{}} algorithms.

The only algorithm that approaches the performance of the \ALGPA{} algorithm, and slightly outperforms it in 10-D, for all functions is the DTS-CMA-ES algorithm. This algorithm ends essentially tied with \ALGPA{} for 2-D and slightly ahead in 10-D once the sampling budgets have been exhausted, although the \ALGPA{} algorithm still has a larger area under the curve in both of these cases. Although the DTS-CMA-ES algorithm exhibits a somewhat more sluggish start, it makes significant progress in subsequent iterations. From~\ref{subsec-app:ertd_coco_full}, DTS-CMA-ES also performs noticeably better than the rest of the algorithms for multimodal functions with adequate global structure from group~\ref{li:multimod_adeq_struct}, with this performance gap growing with dimension. This implies that DTS-CMA-ES is the algorithm that can leverage the underlying structure the most to ignore local minima.

Other observations of note include that SMAC seems to have an advantage for a very limited number of objective function evaluations before being overtaken, presumably due to SMAC being able to start optimi\sz{}ing before other algorithms have finished sampling their respective initial DoEs. The MCS algorithm performs notably well for separable functions, possibly due to the deterministic nature of the search that is aligned with the separability axes.\@ \revision{The AutoSAEA algorithm also performs notably well after a very sluggish start, similar to DTS-CMA-ES, due to the need for sampling the larger initial population required for the evolutionary framework.}

In summary, the \ALGPA{} algorithm is shown to be a leading contender in the field of expensive black-box function optimi\sz{}ation, performing better, when considering all of the BBOB functions, than all of the considered BO-based methods, with the exception of being tied in 10-D with the BADS algorithm.

%In summary, the addition of a trust region to BO is a clearly worthwhile addition, outperforming standard BO for nearly all tested functions. This may be due to mitigating the tendency of standard BO to over-explore the search space and allowing for the more efficient exploitation of the current candidate solution. The particular version of trust region BO presented in this paper, the \ALGPA{} method, further emphasi\sz{}es this local focus, allowing for highly competitive performance on functions not normally associated with standard BO, namely unimodal and highly conditioned functions.

\section{Conclusions}\label{sec:conclusion}

Standard Bayesian \acr{o}ptimi\sz{}ation (BO) has several notable shortcomings, namely experiencing computational slowdown with additional algorithm iterations, not being well-suited to non-stationary and ill-conditioned functions, and exhibiting poor convergence characteristics. Trust-region-based BO algorithms partially address these shortcomings by constraining the selection of the next point to be sampled from the objective function and incorporated into the Gaussian \acr{p}rocess (GP) surrogate model using an iteratively updated trust region. In this paper, we have constructed the \ALGPA{} algorithm using two novel extensions of \revision{\TRBO{}}. The first extension is an adaptive trust region- and objective function observation rescaling strategy, based on the length-scales of the local GP surrogate with an SE kernel and ARD, to allow for improved convergence characteristics. Secondly, this trust region is also rotated such that its axes are aligned with the weighted principal components of the observed data to allow the SE kernel with ARD to model non-stationary and ill-conditioned functions. Along with these extensions, the length-scales of the ARD kernel are approximated iteratively and observed data \revision{outside} this trust region is greedily discarded to alleviate computational slowdown. An ablation study, using the extensive benchmark suite provided by the COCO software, is performed on the \ALGPA{} algorithm which shows that each of the components of the \ALGPA{} algorithm contribute significantly to the overall performance of the algorithm.

Using a set of diverse synthetic test functions, a comparison of the proposed \ALGPA{} algorithm with standard BO and a variety of \revision{\TRBO{}} algorithms shows that the \ALGPA{} algorithm is capable of convergence to a much higher level of precision without encountering numerical issues or instability. A second comparison with a range of state-of-the-art black-box optimi\sz{}ation methods from the wider field of black-box optimi\sz{}ation, also performed using the aforementioned COCO benchmarking software, shows that the \ALGPA{} algorithm is a leading contender in the domain of expensive black-box function optimi\sz{}ation, significantly outperforming standard BO for nearly all tested scenarios and demonstrating exceptional performance compared to state-of-the-art black-box optimi\sz{}ation methods, particularly in the domain of unimodal- and highly conditioned objective functions not typically associated with BO.\@ This is coupled with a slight reduction in its effectiveness compared to other \revision{\TRBO{}} methods when dealing with multimodal functions due to the increased emphasis of the \ALGPA{} algorithm on local exploitation and a lack of a global surrogate model.

An important avenue for future work may include extending the \ALGPA{} algorithm for use with a more general class of objective functions, such as modifying \ALGPA{} to incorporate noisy output observations or categorical- and integer-valued input values (commonly encountered in the hyperparameters of machine learning models~\cite{ScikitLearn}) similarly to the kernel modification technique proposed by Garrido-Merch\'{a}n and Hern\'{a}ndez-Lobato~\cite{GP_integer_categorical_variables}. The local GP model used in the \ALGPA{} could also be augmented with gradient observations~\cite[Ch. 9]{Rasmussen} to improve the speed of the algorithm, possibly allowing for competitive performance when applied to non-black-box optimi\sz{}ation problems.\\

\section*{CRediT authorship contribution statement}
\markright{}
\addcontentsline{toc}{section}{CRediT authorship contribution statement}

\textbf{E. Visser:} Conceptualization, Methodology, Software, Formal analysis, Writing - Original Draft, Writing - Review \& Editing.\ \textbf{J.C. Schoeman:} Conceptualization, Writing - Review \& Editing, Supervision.\ \textbf{C.E. van Daalen:} Conceptualization, Writing - Review \& Editing, Supervision.

\section*{Declaration of competing interest}
\markright{}
\addcontentsline{toc}{section}{Declaration of competing interest}

The authors declare that they have no known competing financial interests or
personal relationships that could have appeared to influence the work reported
in this paper.

\section*{Acknowledgements}
\markright{}
\addcontentsline{toc}{section}{Acknowledgements}

This work was supported by grants from the Wilhelm Frank Bursary Fund as
administered by Stellenbosch University from 2022 to 2024.\ \revision{The authors
would also like to thank anonymous reviewers for their valuable suggestions and
comments, which improved the final version of the paper.}

\appendix

%**********************************************
\section{Offset proof}\label{subsec-app:lin_alg_offset_proof}
%**********************************************

In the \ALGPA{} algorithm, a transformed representation $\tra{X}$ of the observed
input set $X$ is constructed according to the relation defined in
(\ref{eq:input_transform_matrices})

\begin{equation} \label{app-eq:matrix_mapping_1}
    \textbf{X} = \textbf{R} \textbf{S} \tra{\textbf{X}} + \revision{\textbf{c}}\textbf{1}^{\top}_{n}
\end{equation}

During one of these transformations of $\tra{X}$, the current minimum of the
\centered{} input data $\tra{\textbf{x}_{\min}}$ is used to update the values of
$\tra{X}$ using the transform defined in (\ref{eq:input_cen})

\begin{gather}
    \tra{\textbf{X}_{\textrm{new}}} = \tra{\textbf{X}} - \tra{\minpnt{x}}\textbf{1}^{\top}_{n}
\end{gather}

\noindent with the transformation parameter $\revision{\textbf{c}}$ updated according to (\ref{eq:b_update})

\begin{equation}
    \revision{\textbf{c}_{\textrm{new}}} = \revision{\textbf{c}} + \textbf{R} \textbf{S} \tra{\textbf{x}_{\min}}.
\end{equation}

\noindent Substituting these new values into (\ref{app-eq:matrix_mapping_1}), using the fact that matrix multiplication is left- and right distributive, and simplifying yields

\begin{gather}
    \textbf{X} = \textbf{R} \textbf{S} \tra{\textbf{X}_{\textrm{new}}} + \revision{\textbf{c}_{\textrm{new}}}\textbf{1}^{\top}_{n} \\
    \textbf{X} = \textbf{R} \textbf{S} (\tra{\textbf{X}} - \tra{\minpnt{x}}\textbf{1}^{\top}_{n}) + (\revision{\textbf{c}} + \textbf{R} \textbf{S} \tra{\textbf{x}_{\min}})\textbf{1}^{\top}_{n} \nonumber \\
    \textbf{X} = \textbf{R} \textbf{S} \tra{\textbf{X}} + \revision{\textbf{c}}\textbf{1}^{\top}_{n} - \textbf{R} \textbf{S} \tra{\minpnt{x}}\textbf{1}^{\top}_{n} + \textbf{R} \textbf{S} \tra{\textbf{x}_{\min}}\textbf{1}^{\top}_{n} \nonumber \\
    \textbf{X} = \textbf{R} \textbf{S} \tra{\textbf{X}} + \revision{\textbf{c}}\textbf{1}^{\top}_{n}. \nonumber
\end{gather}

\noindent Clearly, this implies that this is the same underlying mapping between $X$ and $\tra{X}$ from (\ref{app-eq:matrix_mapping_1}) expressed using the updated values of $\tra{X}$ and $\revision{\textbf{c}}$, concluding the proof.

%**********************************************
\section{Weighted principal component alignment proof}\label{subsec-app:lin_alg_pc_proof}
%**********************************************

In the \ALGPA{} algorithm, a transformed representation $\tra{X}$ of the observed
input set $X$ is constructed according to the relation defined in
(\ref{eq:input_transform_matrices})

\begin{equation} \label{app-eq:matrix_mapping_2}
    \textbf{X} = \textbf{R} \textbf{S} \tra{\textbf{X}} + \revision{\textbf{c}}\textbf{1}^{\top}_{n}
\end{equation}

During one of these transformations of $\tra{X}$, the weighted principal components
of the \centered{} input data $\textbf{U}_{\textrm{a}}$ (\ref{eq:svd_mod}) is
used to update the values of $\tra{X}$ using the transform defined in
(\ref{eq:input_rot})

\begin{gather}
    \tra{\textbf{X}_{\textrm{new}}} = \textbf{S}^{-1} \textbf{U}^{\top}_{\textrm{a}} \textbf{S} \tra{\textbf{X}}
\end{gather}

\noindent with the transformation parameter $\textbf{R}$ updated according to (\ref{eq:u_update})

\begin{equation}
    \textbf{R}_{\textrm{new}} = \textbf{R} \textbf{U}.
\end{equation}

\noindent Substituting these new values into (\ref{app-eq:matrix_mapping_2}), recalling that the matrix $\textbf{U}$ obtained from the SVD is orthogonal ($\textbf{U}^{-1} = \textbf{U}^{\top}$), and simplifying yields

\begin{gather}
    \textbf{X} = \textbf{R}^{}_{\textrm{new}} \textbf{S} \tra{\textbf{X}_{\textrm{new}}} + \revision{\textbf{c}}\textbf{1}^{\top}_{n} \\
    \textbf{X} = \textbf{R} \textbf{U}^{}_{\textrm{a}} \textbf{S} \textbf{S}^{-1} \textbf{U}^{\top}_{\textrm{a}} \textbf{S} \tra{\textbf{X}} + \revision{\textbf{c}}\textbf{1}^{\top}_{n} \nonumber\\
    \textbf{X} = \textbf{R} \textbf{U}^{}_{\textrm{a}} \textbf{U}^{\top}_{\textrm{a}} \textbf{S} \tra{\textbf{X}} + \revision{\textbf{c}}\textbf{1}^{\top}_{n} \nonumber \\
    \textbf{X} = \textbf{R} \textbf{S} \tra{\textbf{X}} + \revision{\textbf{c}}\textbf{1}^{\top}_{n} \nonumber
\end{gather}

\noindent Clearly, this implies that this is the same underlying mapping between $X$ and $\tra{X}$ from (\ref{app-eq:matrix_mapping_2}) expressed using the updated values of $\tra{X}$ and $\textbf{R}$, concluding the proof.

%**********************************************
\section{Rescaling proof}\label{subsec-app:lin_alg_resc_proof}
%**********************************************

In the \ALGPA{} algorithm, a transformed representation $\tra{X}$ of the observed
input set $X$ is constructed according to the relation defined in
(\ref{eq:input_transform_matrices})

\begin{equation} \label{app-eq:matrix_mapping_3}
    \textbf{X} = \textbf{R} \textbf{S} \tra{\textbf{X}} + \revision{\textbf{c}}\textbf{1}^{\top}_{n}
\end{equation}

During one of these transformations of $\tra{X}$, the most likely length-scales
$\boldsymbol{\ell}^{*}$, collected into a diagonal matrix $\textbf{L}^{-1}$
from (\ref{eq:ell_diag_inv}), is used to update the values of $\tra{X}$ using the
transform defined in (\ref{eq:input_resc})

\begin{gather}
    \tra{\textbf{X}_{\textrm{new}}} = \tra{\textbf{L}^{-1}\textbf{X}}
\end{gather}

\noindent with the transformation parameter $\textbf{S}$ updated according to (\ref{eq:s_update})

\begin{equation}
    \textbf{S}_{\textrm{new}}  = \textbf{L} \textbf{S}
\end{equation}

Both $\textbf{S}$ and $\textbf{L}^{-1}$ are diagonal scaling matrices.\@ \revision{Given that} diagonal matrices are commutative\revision{, simplifying yields}

\begin{gather}
    \textbf{X} = \textbf{R} \textbf{S}_{\textrm{new}} \tra{\textbf{X}_{\textrm{new}}} + \revision{\textbf{c}}\textbf{1}^{\top}_{n} \\
    \textbf{X} = \textbf{R} \textbf{L} \textbf{S} \textbf{L}^{-1} \tra{\textbf{X}} + \revision{\textbf{c}}\textbf{1}^{\top}_{n} \nonumber \\
    \textbf{X} = \textbf{R} \textbf{S} \textbf{L} \textbf{L}^{-1} \tra{\textbf{X}} + \revision{\textbf{c}}\textbf{1}^{\top}_{n} \nonumber \\
    \textbf{X} = \textbf{R} \textbf{S} \tra{\textbf{X}} + \revision{\textbf{c}}\textbf{1}^{\top}_{n} \nonumber
\end{gather}

\noindent Clearly, this implies that this is the same underlying mapping between $X$ and $\tra{X}$ from (\ref{app-eq:matrix_mapping_3}) expressed using the updated values of $\tra{X}$ and $\textbf{S}$, concluding the proof.

\section{Kernel matrix length-scale derivatives for squared exponential kernel with automatic relevance determination}\label{subsec-app:kernel_derivs}

During the optimi\sz{}ation of the length-scales in Sec.~\ref{subsec:hyp_est},
first- and second derivatives of the log-likelihood for GP fitted to the
observations $\tra{X}$ and $\tra{Y}$ (\ref{eq:gp_log_lik}) with respect to the
length-scales the length-scales of the squared exponential kernel function with
automatic relevance determination (\ref{eq:sqexp_vector_form}) are calculated.
These are defined by the Jacobian

\begin{equation} %SINGLE COL
    \nabla\log p(\tra{Y} \,|\, \tra{X}, \boldsymbol{\theta}) = \textbf{J} :=
    {\biggl[ \partialDeriv{\log p(\tra{Y} \,|\, \tra{X}, \boldsymbol{\theta})}{\ln\ell_{i}} \biggr]}_{1 \leq i \leq d} \label{app-eq:log_lik_Jac}
\end{equation}

\noindent and the Hessian

\begin{equation} %SINGLE COL
    \nonumber \nabla^{2}\log p(\tra{Y} \,|\, \tra{X}, \boldsymbol{\theta}) = \textbf{H} :=
    {\biggl[ \partialSecDerivDual{\log p(\tra{Y} \,|\, \tra{X}, \boldsymbol{\theta})}{\ln\ell_{i}}{\ln\ell_{j}} \biggr]}_{1 \leq i,j \leq d}.  \label{app-eq:log_lik_Hess}
\end{equation}

\noindent Using the results derived in~\cite{LogLikHess}\footnote{As stated in~\cite{LogLikHess}, directly calculating the matrix-matrix products in (\ref{eq:log_lik_Jac}) and (\ref{eq:log_lik_Hess}) should be avoided as far as possible. Instead, the products $\textbf{K}^{-1} \partialDeriv{\textbf{K}}{\ell_{i}}$ should be cached, matrix-vector products should be prioritised and only the products needed for the trace terms should be calculated.}, the derivatives that define the entries of these matrices are given for the Jacobian

\begin{equation}
    \partialDeriv{\log p(\tra{Y} \,|\, \tra{X}, \boldsymbol{\theta})}{\ln\ell_{i}} = \frac{1}{2}\tra[\top]{\textbf{y}} \textbf{K}^{-1} \partialDeriv{\textbf{K}}{\ln\ell_{i}} \textbf{K}^{-1} \tra{\textbf{y}}
    -\frac{1}{2}\tr \biggl( \textbf{K}^{-1} \partialDeriv{\textbf{K}}{\ln\ell_{i}} \biggr) \,\,\, \forall i \in \{1, 2, \ldots, d\}
\end{equation}

\noindent and Hessian

\begin{align}
    \partialSecDerivDual{\log p(\tra{Y} \,|\, \tra{X}, \boldsymbol{\theta})}{\ln\ell_{i}}{\ln\ell_{j}} & = \frac{1}{2}\tr \biggl( \textbf{K}^{-1} \partialSecDerivDual{\textbf{K}}{\ln\ell_{i}}{\ln\ell_{j}} \biggr)
    \nonumber -\frac{1}{2}\tr \biggl( \textbf{K}^{-1} \partialDeriv{\textbf{K}}{\ln\ell_{j}} \textbf{K}^{-1} \partialDeriv{\textbf{K}}{\ln\ell_{i}} \biggr)\nonumber \\
    & \quad + \tra[\top]{\textbf{y}} \textbf{K}^{-1} \partialDeriv{\textbf{K}}{\ln\ell_{j}} \textbf{K}^{-1}
    \partialDeriv{\textbf{K}}{\ln\ell_{i}} \textbf{K}^{-1} \tra{\textbf{y}} \nonumber  \\
    & \quad - \frac{1}{2} \tra[\top]{\textbf{y}} \textbf{K}^{-1} \partialSecDerivDual{\textbf{K}}{\ln\ell_{i}}{\ln\ell_{j}} \textbf{K}^{-1} \tra{\textbf{y}} \,\,\, \forall i, j \in \{1, 2, \ldots, d\}.
\end{align}

These derivatives are specifically calculated with respect to the logarithm of
the length-scales and achieved by transforming the derivatives of the kernel
matrix $\textbf{K}$ (\ref{eq:K}) according to

\begin{equation}
    \partialDeriv{\textbf{K}}{\ln\ell_{i}} = \partialDeriv{\textbf{K}}{\ell_{i}} \partialDeriv{\ell_{i}}{\ln\ell_{i}}
\end{equation}

\noindent to ensure that the length-scales are strictly positive. This transform is accomplished by multiplying the derivative of the length-scale with respect to its logarithm

\begin{gather}
    \partialDeriv{\ln\ell_{i}}{\ell_{i}} = \frac{1}{\ell_{i}} \nonumber \\
    \partialDeriv{\ell_{i}}{\ln\ell_{i}} = \ell_{i}.
\end{gather}

%\begin{equation}
%    \partialDeriv{\textbf{K}}{\ell_{i}} = \begin{bmatrix}
%    \partialDeriv{k(\textbf{x}_{1}, \textbf{x}_{1})}{\ell_{i}} & \partialDeriv{k(\textbf{x}_{1}, \textbf{x}_{2})}{\ell_{i}} & \hdots & \partialDeriv{k(\textbf{x}_{1}, \textbf{x}_{n})}{\ell_{i}} \\
%    \partialDeriv{k(\textbf{x}_{2}, \textbf{x}_{1})}{\ell_{i}} & \partialDeriv{k(\textbf{x}_{2}, \textbf{x}_{1})}{\ell_{i}} & \hdots & \partialDeriv{k(\textbf{x}_{2}, \textbf{x}_{n})}{\ell_{i}} \\
%    \vdots & \vdots & \ddots & \vdots \\
%    \partialDeriv{k(\textbf{x}_{n}, \textbf{x}_{1})}{\ell_{i}} & \partialDeriv{k(\textbf{x}_{n}, \textbf{x}_{2})}{\ell_{i}} & \hdots & \partialDeriv{k(\textbf{x}_{n}, \textbf{x}_{n})}{\ell_{i}} \\
%    \end{bmatrix}
%\end{equation}

%\begin{equation}
%    \partialDeriv{\textbf{K}}{\ell_{i}} = \biggl( \partialDeriv{\textbf{K}}{\ell_{i}} \biggr) ^{\top}
%\end{equation}

\noindent The kernel matrix derivative now becomes

\begin{equation} \label{app-eq:kernel_1st_deriv_matrix}
    \partialDeriv{\textbf{K}}{\ln \ell_{i}} = \begin{bmatrix}
        \partialDeriv{k(\textbf{x}_{1}, \textbf{x}_{1})}{\ln \ell_{i}} & \partialDeriv{k(\textbf{x}_{1}, \textbf{x}_{2})}{\ln \ell_{i}} & \hdots & \partialDeriv{k(\textbf{x}_{1}, \textbf{x}_{n})}{\ln \ell_{i}} \\
        \partialDeriv{k(\textbf{x}_{2}, \textbf{x}_{1})}{\ln \ell_{i}} & \partialDeriv{k(\textbf{x}_{2}, \textbf{x}_{1})}{\ln \ell_{i}} & \hdots & \partialDeriv{k(\textbf{x}_{2}, \textbf{x}_{n})}{\ln \ell_{i}} \\
        \vdots                                                         & \vdots                                                         & \ddots & \vdots                                                         \\
        \partialDeriv{k(\textbf{x}_{n}, \textbf{x}_{1})}{\ln \ell_{i}} & \partialDeriv{k(\textbf{x}_{n}, \textbf{x}_{2})}{\ln \ell_{i}} & \hdots & \partialDeriv{k(\textbf{x}_{n}, \textbf{x}_{n})}{\ln \ell_{i}} \\
    \end{bmatrix}.
\end{equation}

\noindent Note that the squared exponential kernel is symmetric (i.e. $k(\textbf{x}_{p}, \textbf{x}_{q}) = k(\textbf{x}_{q}, \textbf{x}_{p})$), therefore the derivative of the kernel matrix is also symmetric

\begin{equation}
    \partialDeriv{\textbf{K}}{\ln \ell_{i}} = {\biggl( \partialDeriv{\textbf{K}}{\ln \ell_{i}} \biggr)}^{\top}.
\end{equation}

\noindent In practical implementations, this symmetry allows the kernel matrix derivatives to be fully described by only calculating the upper- or lower triangular portions of the matrix.

Calculating the entries of (\ref{app-eq:kernel_1st_deriv_matrix}) simply
involve taking the derivative of (\ref{eq:sqexp_vector_form}) with respect to
$\ell_{i}$

\begin{equation}
    \partialDeriv{k(\textbf{x}_{p}, \textbf{x}_{q})}{\ell_{i}} = k(\textbf{x}_{p}, \textbf{x}_{q}) \cdot \frac{{(x_{p_{i}} -x_{q_{i}})}^{2}}{\ell^{3}_{i}}
\end{equation}

\noindent and multiplying with $\ell_{i}$ to obtain the derivative with respect to the logarithm of $\ell_{i}$

\begin{equation}
    \partialDeriv{k(\textbf{x}_{p}, \textbf{x}_{q})}{\ln \ell_{i}} = \partialDeriv{k(\textbf{x}_{p}, \textbf{x}_{q})}{\ell_{i}} \partialDeriv{\ell_{i}}{\ln\ell_{i}} = k(\textbf{x}_{p}, \textbf{x}_{q}) \cdot \frac{{(x_{p_{i}} -x_{q_{i}})}^{2}}{\ell^{2}_{i}}.
\end{equation}

Using these results, the second-derivative matrix with respect to the
length-scales $\partialSecDerivDual{\textbf{K}}{\ln \ell_{i}}{\ln \ell_{j}}$
can be constructed. The entries of these matrices, in the case that $i \neq j$,
is given as

\begin{align}
    \partialSecDerivDual{k(\textbf{x}_{p}, \textbf{x}_{q})}{\ln \ell_{j}}{\ln \ell_{i}}
     & = \partialDeriv{k(\textbf{x}_{p}, \textbf{x}_{q})}{\ln \ell_{j}} \cdot \frac{{(x_{p_{i}} -x_{q_{i}})}^{2}}{\ell^{2}_{i}}.
    %&= k(\textbf{x}_{p}, \textbf{x}_{q}) \cdot \frac{(x_{p_{j}} -x_{q_{j}})^{2}}{\ell^{2}_{j}} \cdot \frac{(x_{p}_{i} -x_{q}_{i})^{2}}{\ell^{2}_{i}} \nonumber \\
    %&= \partialDeriv{k(\textbf{x}_{p}, \textbf{x}_{q})}{\ln \ell_{i}} \cdot \frac{(x_{p_{j}} -x_{q_{j}})^{2}}{\ell^{2}_{j}}  \\ &= \partialSecDerivDual{k(\textbf{x}_{p}, \textbf{x}_{q})}{\ln \ell_{i}}{\ln \ell_{j}} \nonumber
\end{align}

\noindent Expanding and rearranging terms in this formula yields another symmetry

\begin{align}
    \partialSecDerivDual{k(\textbf{x}_{p}, \textbf{x}_{q})}{\ln \ell_{j}}{\ln \ell_{i}}
     & = \partialDeriv{k(\textbf{x}_{p}, \textbf{x}_{q})}{\ln \ell_{j}} \cdot \frac{{(x_{p_{i}} -x_{q_{i}})}^{2}}{\ell^{2}_{i}} \nonumber                          \\
     & = k(\textbf{x}_{p}, \textbf{x}_{q}) \cdot \frac{{(x_{p_{j}} -x_{q_{j}})}^{2}}{\ell^{2}_{j}} \cdot \frac{{(x_{p_{i}} -x_{q_{i}})}^{2}}{\ell^{2}_{i}} \nonumber \\
     & = k(\textbf{x}_{p}, \textbf{x}_{q}) \cdot \frac{{(x_{p_{i}} -x_{q_{i}})}^{2}}{\ell^{2}_{i}} \cdot \frac{{(x_{p_{j}} -x_{q_{j}})}^{2}}{\ell^{2}_{j}} \nonumber \\
     & = \partialDeriv{k(\textbf{x}_{p}, \textbf{x}_{q})}{\ln \ell_{i}} \cdot \frac{{(x_{p_{j}} -x_{q_{j}})}^{2}}{\ell^{2}_{j}} \nonumber                          \\
     & = \partialSecDerivDual{k(\textbf{x}_{p}, \textbf{x}_{q})}{\ln \ell_{i}}{\ln \ell_{j}}.
\end{align}

\noindent This symmetry also implies the following symmetry in the log-likelihood Hessian matrix $\textbf{H}$, reducing the calculations required to determine this matrix,

\begin{equation}
    \partialSecDerivDual{\textbf{K}}{\ln \ell_{i}}{\ln \ell_{j}} = {\biggl( \partialSecDerivDual{\textbf{K}}{\ln \ell_{i}}{\ln \ell_{j}} \biggr)}^{\top}.
\end{equation}

Lastly, for the case where $i=j$, i.e.\ the main diagonal of $\textbf{H}$
(\ref{eq:log_lik_Hess}), the derivative is determined to be

%\begin{equation}
%    \partialSecDeriv{k(\textbf{x}_{p}, \textbf{x}_{q})}{\ell_{i}} = \partialDeriv{k(\textbf{x}_{p}, \textbf{x}_{q})}{\ell_{i}} \biggl( \frac{(x_{p_{i}} -x_{q_{i}})^{2}}{\ell^{3}_{i}} - \frac{3}{\ell_{i}} \biggr)
%\end{equation}

\begin{equation}
    \partialSecDeriv{k(\textbf{x}_{p}, \textbf{x}_{q})}{\ln \ell_{i}} = \partialDeriv{k(\textbf{x}_{p}, \textbf{x}_{q})}{\ln \ell_{i}} \biggl( \frac{{(x_{p_{i}} -x_{q_{i}})}^{2}}{\ell^{2}_{i}} - 2 \biggr).
\end{equation}

%\begin{align}
%    \partialSecDerivDual{k(\textbf{x}_{p}, \textbf{x}_{q})}{\ell_{i}}{\ell_{j}} 
%    &= \partialDeriv{k(\textbf{x}_{p}, \textbf{x}_{q})}{\ell_{i}} \cdot \frac{(x_{p}_{j} -x_{q}_{j})^{2}}{\ell^{3}_{j}} \nonumber \\ 
%    &= k(\textbf{x}_{p}, \textbf{x}_{q}) \cdot \frac{(x_{p_{i}} -x_{q_{i}})^{2}}{\ell^{3}_{i}} \cdot \frac{(x_{p}_{j} -x_{q}_{j})^{2}}{\ell^{3}_{j}} \nonumber \\
%    &= \partialDeriv{k(\textbf{x}_{p}, \textbf{x}_{q})}{\ell_{j}} \cdot \frac{(x_{p_{i}} -x_{q_{i}})^{2}}{\ell^{3}_{i}}  \\ &= \partialSecDerivDual{k(\textbf{x}_{p}, \textbf{x}_{q})}{\ell_{j}}{\ell_{i}} \nonumber
%\end{align}

\pagebreak

\section{LABCAT computational complexity}\label{subsec-app:comp_complex}

\todo[inline]{New appendix section.}

To analy\sz{}e computational complexity of the \ALGPA{} algorithm, we compile the operations used in a single iteration with their respective complexities in Tab.~\ref{tab:operations_complexity} in terms of the number of observations $n$ and dimensionality of the objective function $d$. From this table, we conclude that the asymptotic complexity of the \ALGPA{} algorithm, performed for a total of $N$ iterations, to be $O(N n^{3})$. At each algorithm iteration, due to the observation cache factor $\m{}$ limiting the number of stored observations, we expect the number of observations at each algorithm iteration to remain relatively constant $n \approx \m{} d$. Therefore, the complexity of a single instance of the LABCAT algorithm would be approximately linear in terms of the number of iterations $O(N n^{3}) \approx O(N \m{}^{3}d^{3}) \approx O(N)$, since the values of $\m{}$ and $d$ are fixed at the start of the algorithm instance.

\begin{table}[h]
    \centering
    \begin{tabular}{@{}lcc@{}}
        Operation                          & Example Eq.                  & Computational complexity \\ \midrule
        Cholesky decomposition             & (\ref{eq:K})                 & $O(n^{3})$               \\
        SVD                                & (\ref{eq:svd_mod})           & $O(n^{2}d)$              \\
        Eigenvalue decomposition           & (\ref{eq:log_lik_Hess_aug})  & $O(d^{3})$               \\
        Rotation matrix multiplication     & (\ref{eq:input_rot})         & $O(nd^{2})$              \\
        Scale matrix multiplication        & (\ref{eq:input_resc})        & $O(nd)$                  \\
        Matrix-vector addition/subtraction & (\ref{eq:input_resc_bounds}) & $O(nd)$
    \end{tabular}
    \caption{Computational complexity of operations that predominantly influence 
    the complexity of the \ALGPA{} algorithm with example equations where the respective 
    operations are applied. Operations are ordered by noting that, in almost all cases, $n \gg d$}.\label{tab:operations_complexity}
\end{table}

\section{Synthetic test function benchmark results}\label{sec-app:synth}

\todo[inline]{New appendix section.}

\subsection{Synthetic test function definitions}\label{subsec-app:synth_bench_funcs}

This section contains the definitions of the synthetic test functions used in Sec.~\ref{subsec:synth_results}. Note that while some test functions are defined for $d$-dimensions, we use the two-dimensional instances of these functions for our analysis.

  \newcolumntype{M}[1]{>{\arraybackslash}m{#1}}
  \newcolumntype{F}[1]{>{\small}p{#1}}
  \begin{table}[H]
    \centering
  \resizebox{\textwidth}{!}{%
    \begin{tabular}{@{}p{0.18\textwidth}F{0.45\textwidth}F{0.25\textwidth}F{0.23\textwidth}@{}}
    \toprule
    Function   & Definition & Domain & $f_{\min}$ \\ \midrule
    Sphere~\cite{Sphere} & $f(\textbf{x}) = \sum_{i=1}^{d} x_{i}^{2}$ & $x_{i} \in \lbrack -5.12, 5.12 \rbrack$ & $f(\testpnt{x}) = 0, \newline \testpnt{x} = (0, \ldots, 0)$ \\
    Quartic~\cite{Quartic} & $f(\textbf{x}) = \sum_{i=1}^{d} i x_{i}^{4}$ & $x_{i} \in \lbrack -1.28, 1.28 \rbrack$ & $f(\testpnt{x}) = 0, \newline \testpnt{x} = (0, \ldots, 0)$ \\
    Booth~\cite{Booth} & $f(\textbf{x}) = {(x_{1} + 2 x_{2} - 7)}^{2} \newline \phantom{mmmn}+ {(2 x_{1} + x_{2} - 5)}^2$ & $x_{1}, x_{2} \in \lbrack -10, 10 \rbrack$ & $f(\testpnt{x}) = 0, \newline \testpnt{x} = (1, 3)$ \\
    Rosenbrock~\cite{Rosenbrock} & $f(\textbf{x}) = \sum_{i=1}^{d-1} [ 100 {(x_{i+1} - x_{i}^{2})}^{2} \newline \phantom{mmmn}+ {(x_{i} - 1)}^{2} ] $ & $x_{i} \in \lbrack -5, 10 \rbrack$ & $f(\testpnt{x}) = 0, \newline \testpnt{x} = (1, \ldots, 1)$ \\
    Branin-Hoo~\cite{Branin} & $f(\textbf{x}) = {(x_{2} - \frac{5.1}{4 \pi^{2}} x_{1}^{2} + \frac{5}{\pi} x_{1} - 6)}^{2} \newline \phantom{mmmn}+ 10(1 - \frac{1}{8 \pi}) \cos(x_{1}) + 10$ & $x_{1} \in \lbrack -5, 10 \rbrack, \newline x_{2} \in \lbrack 0, 15 \rbrack$ & $f(\testpnt{x}) = 0.397887, \newline \testpnt{x} = (-\pi, 12.275), \newline (\pi, 2.275), \,\textrm{and} \newline (9.42478, 2.475)$ \\
    Levy~\cite{Levy} & $f(\textbf{x}) = \sum_{i=1}^{d-1} {(w_{i} - 1)}^{2} [1 \newline \phantom{mmmn}+ 10 \sin^{2}(\pi w_{i} +1)] \newline \phantom{mmmn}+ \sin^{2}(\pi w_{1}) \newline \phantom{mmmn}+ {(w_{d} - 1)}^{2} [1 + \sin^{2}(2 \pi w_{d})], \newline \textrm{where} \quad w_{i} = 1 + \frac{x_{i} - 1}{4} \quad \forall i = 1,\ldots,d$ & $x_{i} \in \lbrack -10, 10 \rbrack$ & $f(\testpnt{x}) = 0, \newline \testpnt{x} = (1, \ldots, 1)$ \\ \bottomrule
    \end{tabular}
  }
    \caption{Synthetic benchmark functions}\label{app-tab:synth_bench_funcs}
    \end{table}

    \pagebreak
    \subsection{Full synthetic test function benchmark results}\label{subsec-app:synth_full}
    
    The full results of the comparative algorithm study from
    Sec.~\ref{subsec:synth_results} is provided in this section. Each of the
    compared algorithms chosen in Sec.~\ref{subsec:synth_results} are applied to
    each test function, defined in~\ref{subsec-app:synth_bench_funcs}, for 50
    independent runs. In Tab.~\ref{tab:synth_var_1} and~\ref{tab:synth_var_2}, the
    means and standard deviations of the minimum global regret ($y_{\min} -
        f_{\min}$) reached and the wall-clock times needed for each run are reported.
    Similarly to the method used in~\cite{AutoSAEA,GL_SADE}, for each test
    function, we have also performed a Wilcoxon rank-sum test~\cite{Wilcoxon} (also
    known as the Mann-Whitney U test) on the minimum global regret for each
    algorithm compared to the \ALGPA{} algorithm. We indicate the results of this
    test using ``$+$'', ``$\approx$'' and ``$-$'', if the LABCAT algorithm performs
    statistically significantly better than, comparable to, or worse than the
    compared algorithm. We consider the result to be statistically significant if
    the $p$-value is less than 0.05, which, after Bonferroni
    correction~\cite{Bonferroni} by the number of compared algorithms against
    \ALGPA{} (8), yields a new value of $\frac{0.05}{8} = 0.00625$.
    
    %“+,” “≈,” and “−” indicate that GL-SADE
    %is signiﬁcantly better than, comparable to, or signiﬁcantly out-
    %performed by the compared algorithm, respectively.
    
    %\hl{WILCOXON}~\cite{Wilcoxon} ALSO USED BY ~\cite{AutoSAEA}
    
    \newcommand{\expnum}[2]{{#1}\mathrm{e}{#2}}
    \newcommand{\meanvarexp}[4]{$\expnum{#1}{#2}\pm(\expnum{#3}{#4})$}
    \newcommand{\meanvar}[2]{{$#1\pm(#2)$}}
    \newcommand{\best}[1]{\boldmath{#1}}
    
    \newcommand{\subfigSizeL}{0.15}
    \newcommand{\subfigSizeM}{0.08}
    \newcommand{\subfigSizeR}{0.15}
    \newcolumntype{M}[1]{>{\centering\arraybackslash}m{#1}}
    \begin{table}[H]
        \centering
        \resizebox{\textwidth}{!}{%
            \begin{tabular}{@{}|l|M{\subfigSizeL\textwidth}M{\subfigSizeM\textwidth}|M{\subfigSizeR\textwidth}|M{\subfigSizeL\textwidth}M{\subfigSizeM\textwidth}|M{\subfigSizeR\textwidth}|M{\subfigSizeL\textwidth}M{\subfigSizeM\textwidth}|M{\subfigSizeR\textwidth}|@{}}
                \toprule
                          & \multicolumn{3}{c}{Sphere}        & \multicolumn{3}{c}{Quartic} & \multicolumn{3}{c}{Booth}                                                                                                                                                                              \\ \midrule
                Algorithm & $\mu\pm(\sigma)$                  & $+$/$\approx$/$-$           & Time                      & $\mu\pm(\sigma)$                  & $+$/$\approx$/$-$             & Time                     & $\mu\pm(\sigma)$                  & $+$/$\approx$/$-$            & Time                     \\ \midrule
                LABCAT    & \meanvarexp{5.68}{-17}{7.44}{-17} & N/A & \meanvar{0.055}{0.011}    & \meanvarexp{2.79}{-22}{6.40}{-22} & N/A                   & \meanvar{0.055}{0.007}   & \meanvarexp{9.98}{-16}{1.28}{-15} & N/A                  & \meanvar{0.055}{0.012}   \\ \midrule
                BADS      & \meanvarexp{3.80}{-09}{1.20}{-08} & $+$ & \meanvar{5.448}{0.704}    & \meanvarexp{1.20}{-08}{5.29}{-08} & $+$ & \meanvar{4.890}{0.758}   & \meanvarexp{4.88}{-06}{2.25}{-05} & $+$ & \meanvar{6.868}{0.801}   \\ \midrule
                BO        & \meanvarexp{1.36}{-05}{1.93}{-05} & $+$ & \meanvar{21.507}{0.779}   & \meanvarexp{1.52}{-08}{1.81}{-08} & $+$ & \meanvar{24.213}{0.933}  & \meanvarexp{4.87}{-05}{5.48}{-05} & $+$ & \meanvar{24.271}{1.119}  \\ \midrule
                Random    & \meanvarexp{1.32}{-01}{1.18}{-01} & $+$ & \meanvar{0.010}{0.005}    & \meanvarexp{5.38}{-04}{1.15}{-03} & $+$ & \meanvar{0.009}{0.002}   & \meanvarexp{2.62}{+00}{1.92}{+00} & $+$ & \meanvar{0.010}{0.005}   \\ \midrule
                SRSM      & \meanvarexp{2.14}{-02}{1.50}{-01} & $+$ & \meanvar{34.696}{2.249}   & \meanvarexp{9.17}{-10}{5.90}{-09} & $+$ & \meanvar{29.548}{0.916}  & \meanvarexp{5.77}{-05}{7.77}{-05} & $+$ & \meanvar{37.755}{1.628}  \\ \midrule
                TREGO     & \meanvarexp{1.73}{-03}{2.80}{-03} & $+$ & \meanvar{111.496}{2.323}  & \meanvarexp{3.82}{-04}{1.91}{-03} & $+$  & \meanvar{111.667}{8.665} & \meanvarexp{6.23}{-02}{1.30}{-01} & $+$ & \meanvar{109.807}{1.846} \\ \midrule
                TRLBO     & \meanvarexp{1.34}{-04}{3.37}{-04} & $+$ & \meanvar{1.408}{0.067}    & \meanvarexp{1.90}{-04}{5.24}{-04} & $+$ & \meanvar{1.427}{0.110}   & \meanvarexp{2.64}{-02}{5.75}{-02} & $+$ & \meanvar{1.326}{0.073}   \\ \midrule
                TuRBO-1   & \meanvarexp{3.36}{-04}{3.18}{-04} & $+$ & \meanvar{1.407}{0.124}    & \meanvarexp{8.94}{-09}{5.47}{-08} & $+$ & \meanvar{1.474}{0.102}   & \meanvarexp{4.67}{-03}{1.13}{-02} & $+$ & \meanvar{1.446}{0.134}   \\ \midrule
                TuRBO-5   & \meanvarexp{1.22}{-03}{1.27}{-03} & $+$ & \meanvar{4.019}{0.442}    & \meanvarexp{4.40}{-06}{6.09}{-06} & $+$ & \meanvar{4.138}{0.339}   & \meanvarexp{3.36}{-02}{3.63}{-02} & $+$ & \meanvar{4.005}{0.392}   \\ \bottomrule
            \end{tabular}%
        }
        \caption{Average and standard deviation of the minimum global regret, their statistical comparisons according to a rank-sum test, and mean and standard deviation of the wall-clock times for 50 independent runs on synthetic test functions $f_{1} - f_{3}$.}\label{tab:synth_var_1}
    \end{table}
    
    \begin{table}[H]
        \centering
        \resizebox{\textwidth}{!}{%
            \begin{tabular}{@{}|l|M{\subfigSizeL\textwidth}M{\subfigSizeM\textwidth}|M{\subfigSizeR\textwidth}|M{\subfigSizeL\textwidth}M{\subfigSizeM\textwidth}|M{\subfigSizeR\textwidth}|M{\subfigSizeL\textwidth}M{\subfigSizeM\textwidth}|M{\subfigSizeR\textwidth}|@{}}
                \toprule
                          & \multicolumn{3}{c}{Rosenbrock}    & \multicolumn{3}{c}{Branin-Hoo} & \multicolumn{3}{c}{Levy}                                                                                                                                                                             \\ \midrule
                Algorithm & $\mu\pm(\sigma)$                  & $+$/$\approx$/$-$              & Time                     & $\mu\pm(\sigma)$                  & $+$/$\approx$/$-$            & Time                    & $\mu\pm(\sigma)$                  & $+$/$\approx$/$-$             & Time                     \\ \midrule
                LABCAT    & \meanvarexp{1.08}{-10}{1.36}{-10} & N/A & \meanvar{0.056}{0.011}   & \meanvarexp{1.71}{-11}{3.02}{-11} & N/A & \meanvar{0.091}{0.046}  & \meanvarexp{1.26}{-01}{5.95}{-01} & N/A & \meanvar{0.061}{0.018}   \\ \midrule
                BADS      & \meanvarexp{6.86}{-04}{9.92}{-04} & $+$ & \meanvar{6.807}{0.647}   & \meanvarexp{4.57}{-07}{1.27}{-06} & $+$ & \meanvar{5.182}{0.795}  & \meanvarexp{4.25}{-07}{1.76}{-06} & $+$ & \meanvar{4.983}{0.861}   \\ \midrule
                BO        & \meanvarexp{2.76}{-01}{3.22}{-01} & $+$ & \meanvar{35.238}{1.950}  & \meanvarexp{3.08}{-05}{2.71}{-05} & $+$ & \meanvar{25.093}{1.075} & \meanvarexp{3.47}{-05}{3.67}{-05} & $+$ & \meanvar{24.546}{1.232}  \\ \midrule
                Random    & \meanvarexp{4.04}{+00}{4.85}{+00} & $+$ & \meanvar{0.017}{0.004}   & \meanvarexp{4.14}{-01}{4.12}{-01} & $+$ & \meanvar{0.010}{0.003}  & \meanvarexp{2.01}{-01}{2.02}{-01} & $+$ & \meanvar{0.010}{0.004}   \\ \midrule
                SRSM      & \meanvarexp{3.43}{+00}{4.12}{+00} & $+$ & \meanvar{32.168}{2.477}  & \meanvarexp{2.39}{-05}{3.93}{-05} & $+$ & \meanvar{36.125}{1.802} & \meanvarexp{2.40}{-05}{1.02}{-04} & $+$ & \meanvar{35.992}{2.253}  \\ \midrule
                TREGO     & \meanvarexp{1.25}{+00}{1.33}{+00} & $+$ & \meanvar{110.343}{2.684} & \meanvarexp{4.50}{-03}{1.29}{-02} & $+$ & \meanvar{110.813}{2.427}& \meanvarexp{2.83}{-03}{3.46}{-03} & $+$ & \meanvar{110.802}{2.280} \\ \midrule
                TRLBO     & \meanvarexp{1.68}{+00}{3.05}{+00} & $+$ & \meanvar{1.318}{0.062}   & \meanvarexp{8.52}{-03}{4.03}{-02} & $+$ & \meanvar{1.407}{0.077}  & \meanvarexp{2.13}{-01}{6.32}{-01} & $+$ & \meanvar{1.393}{0.057}   \\ \midrule
                TuRBO-1   & \meanvarexp{2.35}{+00}{3.27}{+00} & $+$ & \meanvar{1.417}{0.120}   & \meanvarexp{1.58}{-03}{2.23}{-03} & $+$ & \meanvar{1.445}{0.137}  & \meanvarexp{1.05}{-03}{3.97}{-03} & $+$ & \meanvar{1.499}{0.204}   \\ \midrule
                TuRBO-5   & \meanvarexp{1.65}{+00}{1.46}{+00} & $+$ & \meanvar{4.144}{0.254}   & \meanvarexp{9.21}{-03}{1.11}{-02} & $+$ & \meanvar{3.983}{0.353}  & \meanvarexp{9.13}{-03}{1.61}{-02} & $+$ & \meanvar{3.328}{0.536}   \\ \bottomrule
            \end{tabular}%
        }
        \caption{Average and standard deviation of the minimum global regret, their statistical comparisons according to a rank-sum test, and mean and standard deviation of the wall-clock times for 50 independent runs on synthetic test functions $f_{4} - f_{6}$.}\label{tab:synth_var_2}
    \end{table}

    \pagebreak    

\section{COCO benchmark results}\label{sec-app:coco}

\FloatBarrier{}

\subsection{BBOB 2009 objective functions}\label{subsec-app:bbob_2009}

\todo[inline]{New appendix subsection.}

This section contains the objective functions contained in the BBOB 2009 test suite as defined by Hansen et al.~\cite{COCO_2009_doc}, given in Tab.~\ref{app-tab:BBOB_funcs}.

%%%%%%%%% Standard

\begin{table}[H]
    \centering
    % \resizebox{\textwidth}{!}{%
    \begin{tabular}{@{}p{0.05\linewidth}p{0.40\linewidth}p{0.55\linewidth}@{}}
    \toprule
      & Function name                                        & Notes \\ \midrule
      & Separable functions                                  &       \\ \cmidrule(lr){2-2}
    $f_{1}$ & Sphere                                               &       \\
    $f_{2}$ & Ellipsoidal                                          & Conditioning $\approx{} 10^{6}$ \\
    $f_{3}$ & Rastrigin                                            & $10^{d}$ local minima with regular structure \\
    $f_{4}$ & Büche-Rastrigin                                      & Assymetric transform applied to $f_{3}$ \\
    $f_{5}$ & Linear Slope                                         & Linear function with solution on domain boundary \\ \cmidrule(lr){2-2}
      & Unimodal functions with low or moderate conditioning &       \\ \cmidrule(lr){2-2}
    $f_{6}$ & Attractive Sector                                    & Highly asymmetric \\
    $f_{7}$ & Step Ellipsoidal                                     & Similar to $f_2$, cosisting of many plateaus \\
    $f_{8}$ & Rosenbrock                                           & Curved $n - 1$ dimensional valley      \\
    $f_{9}$ & Rotated Rosenbrock                                   & Rotated $f_{8}$   \\ \cmidrule(lr){2-2}
      & Unimodal functions with high conditioning            &       \\ \cmidrule(lr){2-2}
    $f_{10}$ & Ellipsoidal                                          & Rotated $f_{2}$ \\
    $f_{11}$ & Discus                                               & Single search direction 1000 times more sensitive than all others \\
    $f_{12}$ & Bent Cigar                                           & Non-quadratic valley must be followed to optimum \\
    $f_{13}$ & Sharp Ridge                                          & Similar to $f_{12}$, non-differentiable valley floor \\
    $f_{14}$ & Different Powers                                     & Input variable sensitivities change approaching optimum \\ \cmidrule(lr){2-2}
      & Multimodal functions with adequate global structure  &       \\ \cmidrule(lr){2-2}
    $f_{15}$ & Rastrigin                                            & Non-separable $f_{3}$ \\
    $f_{16}$ & Weierstrass                                          & Non-unique optimum, highly rugged and moderately repetitive landscape \\
    $f_{17}$ & Schaffers F7                                         & Highly multimodal with varying frequency and amplitude of modulation\\
    $f_{18}$ & Schaffers F7, Moderately ill-conditioned             & Moderately ill-conditioned $f_{17}$ \\
    $f_{19}$ & Composite Griewank-Rosenbrock                        & Resembles $f_{8}$ in a highly multimodal way \\ \cmidrule(lr){2-2}
      & Multimodal functions with weak global structure      &       \\ \cmidrule(lr){2-2}
    $f_{20}$ & Schwefel                                             & $2^{d}$ prominent local minima in corners of unpenali\sz{}ed, rectangular search area \\
    $f_{21}$ & Gallagher's Gaussian 101-me Peaks                    & 101 optima with random heights and positions \\
    $f_{22}$ & Gallagher's Gaussian 21-hi Peaks                     & 21 optima with random heights and positions, higher conditioning ($1000$) vs. $f_{21}$ ($30$)\\
    $f_{23}$ & Katsuura                                             & $10^{d}$ global optima, highly rugged and repetitive \\
    $f_{24}$ & Lunacek bi-Rastrigin                                 & Highly multimodal with $2$ funnels, one leading to a local minimum that covers $70\%$ of the search space\\ \bottomrule
    \end{tabular}
    % }
    \caption{BBOB 2009 test suite objective functions. Notes regarding each objective function are given by the authors of the original test suite specification document by Hansen et al~\cite{COCO_2009_doc}.}
    \label{app-tab:BBOB_funcs}
    \end{table}

  \FloatBarrier{}
  \subsection{\ALGPA{} ablation study COCO results}\label{subsec-app:ertd_coco_pca_comp}
  
  The full results of the \ALGPA{} ablation study from Sec.~\ref{subsubsec:coco_ablation} using the COCO benchmark  are provided in this section.
  
  \captionsetup[table]{name=Figure}
  \setcounter{table}{0}
  \renewcommand{\subfigSize}{0.28}
  \newcolumntype{M}[1]{>{\centering\arraybackslash}m{#1}}
  
  \begin{table*}[h]
      \centering
      \begin{tabular}{M{\subfigSize\textwidth}M{\subfigSize\textwidth}M{\subfigSize\textwidth}}
         \toprule
          All Functions & \ref{li:separable} Separable & \ref{li:unimod_low_cond} Unimodal, Low~Conditioning \\
          \midrule
          \includesvg[width=\subfigSize\textwidth]{figures/Section4/coco_param_sweep/pprldmany_02D_noiselessall.svg} & \includesvg[width=\subfigSize\textwidth]{figures/Section4/coco_param_sweep/pprldmany_02D_separ.svg} & \includesvg[width=\subfigSize\textwidth]{figures/Section4/coco_param_sweep/pprldmany_02D_lcond.svg} \\
  
          \midrule
          \ref{li:unimod_high_cond} Unimodal, High~Conditioning & \ref{li:multimod_adeq_struct} Multimodal, Adequate~Structure & \ref{li:multimod_weak_struct} Multimodal, Weak~Structure \\
          \midrule
          
          \includesvg[width=\subfigSize\textwidth]{figures/Section4/coco_param_sweep/pprldmany_02D_hcond.svg} & \includesvg[width=\subfigSize\textwidth]{figures/Section4/coco_param_sweep/pprldmany_02D_multi.svg} & \includesvg[width=\subfigSize\textwidth]{figures/Section4/coco_param_sweep/pprldmany_02D_mult2.svg} \\
          
          \bottomrule
      \end{tabular}
      \caption{2-D runtime ablation ECDFs table from COCO dataset}
      \label{tab:ertd_2D_pca}
  \end{table*}
  
  \begin{table*}[h]
      \centering
      \begin{tabular}{M{\subfigSize\textwidth}M{\subfigSize\textwidth}M{\subfigSize\textwidth}}
         \toprule
          All Functions & \ref{li:separable} Separable & \ref{li:unimod_low_cond} Unimodal, Low~Conditioning \\
          \midrule
          \includesvg[width=\subfigSize\textwidth]{figures/Section4/coco_param_sweep/pprldmany_05D_noiselessall.svg} & \includesvg[width=\subfigSize\textwidth]{figures/Section4/coco_param_sweep/pprldmany_05D_separ.svg} & \includesvg[width=\subfigSize\textwidth]{figures/Section4/coco_param_sweep/pprldmany_05D_lcond.svg} \\
  
          \midrule
          \ref{li:unimod_high_cond} Unimodal, High~Conditioning & \ref{li:multimod_adeq_struct} Multimodal, Adequate~Structure & \ref{li:multimod_weak_struct} Multimodal, Weak~Structure \\
          \midrule
          
          \includesvg[width=\subfigSize\textwidth]{figures/Section4/coco_param_sweep/pprldmany_05D_hcond.svg} & \includesvg[width=\subfigSize\textwidth]{figures/Section4/coco_param_sweep/pprldmany_05D_multi.svg} & \includesvg[width=\subfigSize\textwidth]{figures/Section4/coco_param_sweep/pprldmany_05D_mult2.svg} \\
          
          \bottomrule
      \end{tabular}
      \caption{5-D runtime ablation ECDFs table from COCO dataset}
      \label{tab:ertd_5D_pca}
  \end{table*}
  
  \begin{table*}[h]
      \centering
      \begin{tabular}{M{\subfigSize\textwidth}M{\subfigSize\textwidth}M{\subfigSize\textwidth}}
         \toprule
          All Functions & \ref{li:separable} Separable & \ref{li:unimod_low_cond} Unimodal, Low~Conditioning \\
          \midrule
          \includesvg[width=\subfigSize\textwidth]{figures/Section4/coco_param_sweep/pprldmany_10D_noiselessall.svg} & \includesvg[width=\subfigSize\textwidth]{figures/Section4/coco_param_sweep/pprldmany_10D_separ.svg} & \includesvg[width=\subfigSize\textwidth]{figures/Section4/coco_param_sweep/pprldmany_10D_lcond.svg} \\
  
          \midrule
          \ref{li:unimod_high_cond} Unimodal, High~Conditioning & \ref{li:multimod_adeq_struct} Multimodal, Adequate~Structure & \ref{li:multimod_weak_struct} Multimodal, Weak~Structure \\
          \midrule
          
          \includesvg[width=\subfigSize\textwidth]{figures/Section4/coco_param_sweep/pprldmany_10D_hcond.svg} & \includesvg[width=\subfigSize\textwidth]{figures/Section4/coco_param_sweep/pprldmany_10D_multi.svg} & \includesvg[width=\subfigSize\textwidth]{figures/Section4/coco_param_sweep/pprldmany_10D_mult2.svg} \\
          
          \bottomrule
      \end{tabular}
      \caption{10-D runtime ablation ECDFs table from COCO dataset}
      \label{tab:ertd_10D_pca}
  \end{table*}
  
  \FloatBarrier

  \subsection{Full COCO results}\label{subsec-app:ertd_coco_full}

  The full results of the comparative algorithm study from Sec.~\ref{subsubsec:coco_full} using the COCO benchmark are provided in this section.
  
  \captionsetup[table]{name=Figure}
  \renewcommand{\subfigSize}{0.28}
  \newcolumntype{M}[1]{>{\centering\arraybackslash}m{#1}}
  
  \begin{table*}[hp]
      \centering
      \begin{tabular}{M{\subfigSize\textwidth}M{\subfigSize\textwidth}M{\subfigSize\textwidth}}
         \toprule
          All Functions & \ref{li:separable} Separable & \ref{li:unimod_low_cond} Unimodal, Low~Conditioning \\
          \midrule
          \includesvg[width=\subfigSize\textwidth]{figures/Section4/coco_results/pprldmany_02D_noiselessall.svg} & \includesvg[width=\subfigSize\textwidth]{figures/Section4/coco_results/pprldmany_02D_separ.svg} & \includesvg[width=\subfigSize\textwidth]{figures/Section4/coco_results/pprldmany_02D_lcond.svg} \\
  
          \midrule
          \ref{li:unimod_high_cond} Unimodal, High~Conditioning & \ref{li:multimod_adeq_struct} Multimodal, Adequate~Structure & \ref{li:multimod_weak_struct} Multimodal, Weak~Structure \\
          \midrule
          
          \includesvg[width=\subfigSize\textwidth]{figures/Section4/coco_results/pprldmany_02D_hcond.svg} & \includesvg[width=\subfigSize\textwidth]{figures/Section4/coco_results/pprldmany_02D_multi.svg} & \includesvg[width=\subfigSize\textwidth]{figures/Section4/coco_results/pprldmany_02D_mult2.svg} \\

          \bottomrule
      \end{tabular}
      \caption{2-D runtime ECDFs table from COCO dataset}
      \label{tab:ertd_2D}
  \end{table*}
  
  \begin{table*}[hp]
      \centering
      \begin{tabular}{M{\subfigSize\textwidth}M{\subfigSize\textwidth}M{\subfigSize\textwidth}}
         \toprule
          All Functions & \ref{li:separable} Separable & \ref{li:unimod_low_cond} Unimodal, Low~Conditioning \\
          \midrule
          \includesvg[width=\subfigSize\textwidth]{figures/Section4/coco_results/pprldmany_05D_noiselessall.svg} & \includesvg[width=\subfigSize\textwidth]{figures/Section4/coco_results/pprldmany_05D_separ.svg} & \includesvg[width=\subfigSize\textwidth]{figures/Section4/coco_results/pprldmany_05D_lcond.svg} \\
  
          \midrule
          \ref{li:unimod_high_cond} Unimodal, High~Conditioning & \ref{li:multimod_adeq_struct} Multimodal, Adequate~Structure & \ref{li:multimod_weak_struct} Multimodal, Weak~Structure \\
          \midrule
          
          \includesvg[width=\subfigSize\textwidth]{figures/Section4/coco_results/pprldmany_05D_hcond.svg} & \includesvg[width=\subfigSize\textwidth]{figures/Section4/coco_results/pprldmany_05D_multi.svg} & \includesvg[width=\subfigSize\textwidth]{figures/Section4/coco_results/pprldmany_05D_mult2.svg} \\

          \bottomrule
      \end{tabular}
      \caption{5-D runtime ECDFs table from COCO dataset}
      \label{tab:ertd_5D}
  \end{table*}
  
  \begin{table*}[ht]
      \centering
      \begin{tabular}{M{\subfigSize\textwidth}M{\subfigSize\textwidth}M{\subfigSize\textwidth}}
         \toprule
          All Functions & \ref{li:separable} Separable & \ref{li:unimod_low_cond} Unimodal, Low~Conditioning \\
          \midrule
          \includesvg[width=\subfigSize\textwidth]{figures/Section4/coco_results/pprldmany_10D_noiselessall.svg} & \includesvg[width=\subfigSize\textwidth]{figures/Section4/coco_results/pprldmany_10D_separ.svg} & \includesvg[width=\subfigSize\textwidth]{figures/Section4/coco_results/pprldmany_10D_lcond.svg} \\
  
          \midrule
          \ref{li:unimod_high_cond} Unimodal, High~Conditioning & \ref{li:multimod_adeq_struct} Multimodal, Adequate~Structure & \ref{li:multimod_weak_struct} Multimodal, Weak~Structure \\
          \midrule
          
          \includesvg[width=\subfigSize\textwidth]{figures/Section4/coco_results/pprldmany_10D_hcond.svg} & \includesvg[width=\subfigSize\textwidth]{figures/Section4/coco_results/pprldmany_10D_multi.svg} & \includesvg[width=\subfigSize\textwidth]{figures/Section4/coco_results/pprldmany_10D_mult2.svg} \\

          \bottomrule
      \end{tabular}
      \caption{10-D runtime ECDFs table from COCO dataset}
      \label{tab:ertd_10D}
  \end{table*}
  
  \FloatBarrier{}

%\subfile{backmatter/app_transform}
% \section*{References}
% \markright{}
% \addcontentsline{toc}{section}{References}

\bibliography{references}

\end{document}